\documentclass[journal]{IEEEtran}
\usepackage{amsmath,amsfonts}
\usepackage{algorithmic}
\usepackage{algorithm}
\usepackage{array}
\usepackage[caption=false,font=normalsize,labelfont=sf,textfont=sf]{subfig}
\usepackage{textcomp}
\usepackage{stfloats}
\usepackage{url}
\usepackage{verbatim}
\usepackage{graphicx}
\usepackage{cite}
\usepackage{svg}
\usepackage{authblk}
\usepackage{booktabs}
\begin{document}

\title{Medical Multimodal Foundation Models in Clinical Diagnosis and Treatment: Applications, Challenges, and Future Directions}

\author[1]{Kai Sun\thanks{$^{\ast}$ These authors contributed equally.}$^{\ast}$}
\author[1]{Siyan Xue$^{\ast}$}
\author[2]{Fuchun Sun\thanks{$^{\dag}$ Corresponding authors: Fuchun Sun, Jun Yan, and Jiahong Dong.}$^{\dag}$}
\author[1]{Haoran Sun}
\author[2]{Yu Luo}
\author[3]{Ling Wang}
\author[4]{Siyuan Wang}
\author[5]{Na Guo}
\author[1]{Lei Liu}
\author[1]{Tian Zhao}
\author[6]{Xinzhou Wang}
\author[4]{Lei Yang}
\author[4]{Shuo Jin}
\author[4]{Jun Yan$^{\dag}$}
\author[1,4,7]{Jiahong Dong$^{\dag}$}

\affil[1]{School of Biomedical Engineering, Tsinghua University, Beijing, China}
\affil[2]{Department of Computer Science and Technology, Tsinghua University, Beijing, China}
\affil[3]{Xi’an Research Institute of Hi-Tech, Shaanxi, China}
\affil[4]{Hepato-Pancreato-Biliary Center, Beijing Tsinghua Changgung Hospital, School of Clinical Medicine, Tsinghua University, Beijing, China}
\affil[5]{School of Computer and Communication Engineering, University of Science and Technology Beijing}
\affil[6]{College of Electronic Information Engineering, Tongji University, Shanghai, China}
\affil[7]{Research Unit of Precision hepatobiliary Surgery Paradigm, Chinese Academy of Medical Sciences, Beijing, China}

\maketitle

\begin{abstract}
Recent advancements in deep learning have significantly revolutionized the field of clinical diagnosis and treatment, offering novel approaches to improve diagnostic precision and treatment efficacy across diverse clinical domains, thus driving the pursuit of precision medicine. The growing availability of multi-organ and multimodal datasets has accelerated the development of large-scale Medical Multimodal Foundation Models (MMFMs). These models, known for their strong generalization capabilities and rich representational power, are increasingly being adapted to address a wide range of clinical tasks, from early diagnosis to personalized treatment strategies. This review offers a comprehensive analysis of recent developments in MMFMs, focusing on three key aspects: datasets, model architectures, and clinical applications. We also explore the challenges and opportunities in optimizing multimodal representations and discuss how these advancements are shaping the future of healthcare by enabling improved patient outcomes and more efficient clinical workflows. 
\end{abstract}


\section{Introduction}
\IEEEPARstart{T}{he} rapid advancements in artificial intelligence (AI) have revolutionized numerous fields, with medical data analysis being one of the most profoundly impacted\cite{rajpurkar2022ai}. AI has enabled significant improvements in both diagnostic precision and therapeutic planning, making it a cornerstone of precision medicine\cite{tu2024towards}. Using the power of large-scale data and deep learning algorithms, AI systems have been increasingly capable of extracting valuable insights from complex medical data, thus enhancing clinical decision-making processes\cite{acosta2022multimodal}.

Within this broader AI landscape, foundation models (FMs) have emerged as a critical technological leap\cite{bommasani2021opportunities}. Unlike traditional deep learning models that are often tailored for specific tasks, foundation models are pre-trained on vast and diverse datasets, enabling them to generalize across a wide array of downstream tasks without the need for task-specific retraining. These models leverage the scale of their training data to exhibit strong zero-shot learning capabilities, which means they can effectively handle new tasks with minimal or no additional labeled data. Notable examples of foundation models, such as BERT\cite{bert}, CLIP\cite{CLIP}, and DALL-E\cite{dall_e}, have demonstrated their ability to generalize across modalities and tasks, becoming benchmarks in the AI community.

Building on the success of foundation models, the concept of large multimodal models (LMMs) has gained traction in multiple domains, including medical imaging\cite{yang2023dawn,area_1,area_2,area_3,area_4,llava_med}. LMMs are designed to integrate and analyze data from multiple modalities-such as images, text, and clinical reports-enabling a more comprehensive understanding of complex data. This integrative capability is of paramount importance in healthcare, where precise diagnosis and effective treatment planning often require the comprehensive fusion of data from diverse imaging modalities, such as magnetic resonance imaging (MRI), computed tomography (CT), and X-ray, along with non-imaging sources like clinical reports and laboratory findings. LMMs have shown the potential to unify multimodal data, providing more holistic insights and improving the performance of various downstream tasks such as segmentation, classification, and disease detection\cite{moor2023foundation}.

\begin{figure*}[!t]
\begin{center}
\includegraphics[width=1.0\linewidth]{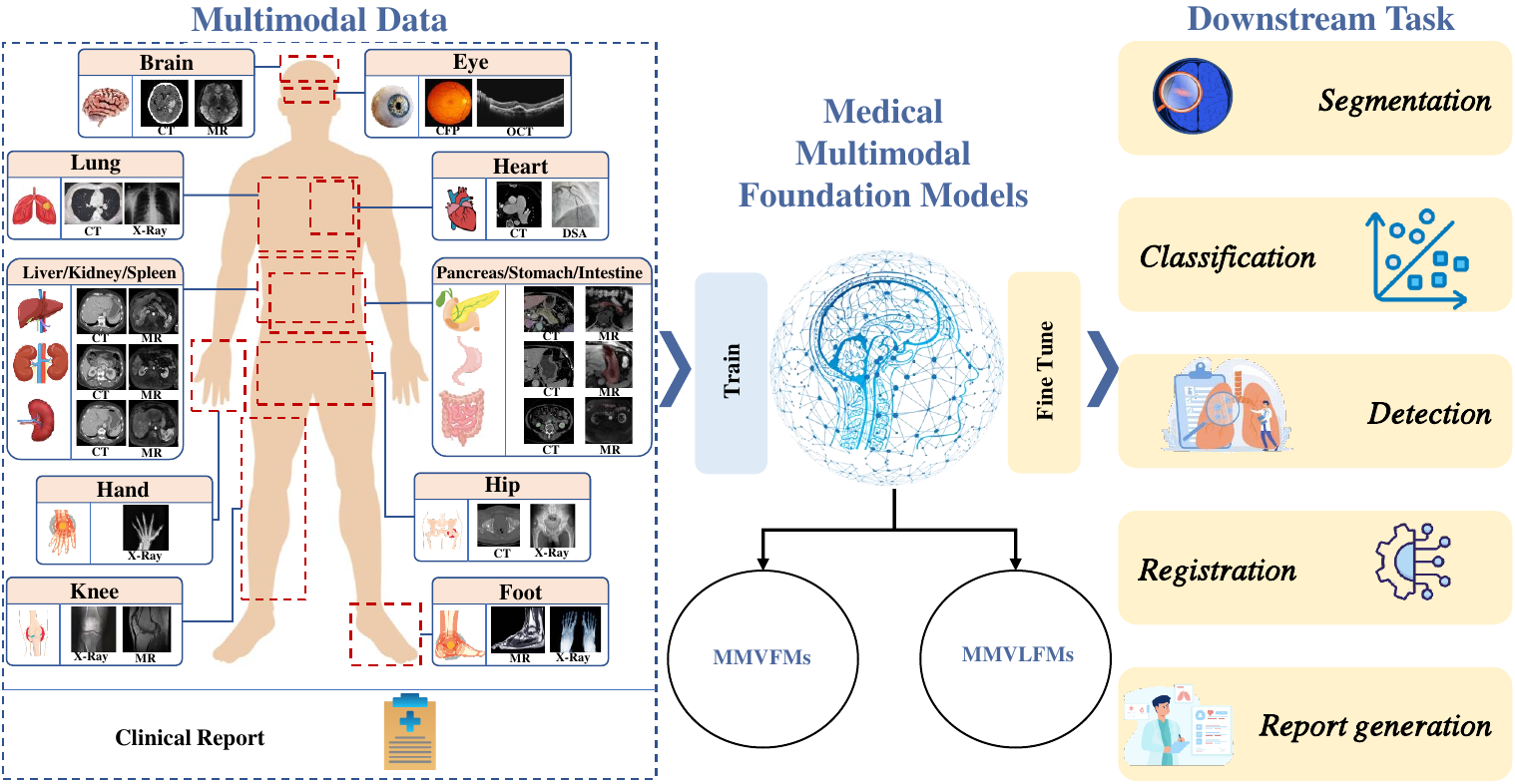}
\end{center}
\vspace{-1em}
\caption{Overview of the MMFMs workflow, illustrating how diverse multimodal medical imaging data from various organs are used for training and fine-tuning MMFMs. The two main model categories, MMVFMs and MMVLFMs, are trained and then fine-tuned for downstream tasks such as segmentation, classification, detection, registration, and clinical report generation.}
\label{overall}
\end{figure*}

As LMMs continue to evolve, the field of medical image analysis has seen the emergence of MMFMs, which are specialized models designed to address the unique challenges posed by medical data\cite{zhang2024challenges}. MMFMs take advantage of the multimodal nature of medical imaging and are trained in large, diverse datasets that cover multiple organs and modalities. These models offer robust generalization capabilities, making them well suited for a wide range of medical tasks, from diagnostic segmentation and classification to more complex tasks such as image registration and clinical report generation. The development of MMFMs represents a significant step forward in AI-driven healthcare, as these models can seamlessly integrate and adapt to various medical contexts.

Within the MMFM landscape, two prominent categories have emerged: Medical Multimodal Vision Foundation Models (MMVFMs) and Medical Multimodal Vision-Language Foundation Models (MMVLFMs). MMVFMs focus on multimodal vision tasks, such as integrating and processing different types of medical images, while MMVLFMs extend the multimodal approach by incorporating both visual and textual data, thus enabling more comprehensive analyzes that bridge the gap between images and clinical documentation. These two approaches form the foundation of MMFMs and are critical to advancing the field of medical artificial general intelligence.

The workflow in Fig. \ref{overall} illustrates the structure and training process of MMFMs. Diverse multimodal medical data are used for training these models, which are subsequently fine-tuned for various downstream medical tasks. The figure highlights the integration of multiple organs and modalities, emphasizing how MMVFMs and MMVLFMs contribute to the segmentation, classification, detection, registration and clinical report generation tasks.

This review provides a comprehensive exploration of the latest advancements in MMFMs, focusing on three key dimensions: datasets, model architectures, and clinical applications. First, we examine the large-scale datasets used to train these models, analyzing the impact of dataset diversity and scale on model performance. Next, we explore the architectural innovations and technical approaches employed in MMVFMs and MMVLFMs, with a focus on their multimodal capabilities. Finally, we discuss the practical performance of these models in clinical settings, highlighting both their successes and the challenges encountered in real-world medical applications.

Through this detailed analysis, we aim to elucidate the transformative potential of MMFMs in medical imaging and provide a roadmap for future research. As these models continue to evolve, we anticipate that MMFMs will play a pivotal role in the advancement of medical artificial general intelligence, thus driving the next wave of innovation in precision medicine.


\section{Background}
\subsection{Foundation Representation Model: A brief history and milestones}
Foundation models represent a novel paradigm in the era of general artificial intelligence, referring to models pre-trained on large-scale datasets and capable of transferring to a wide range of downstream tasks. The core idea of foundation models originates from the long-standing development of transfer learning\cite{bommasani2021opportunities}. The rapid emergence of foundation models is driven by advancements in three key factors: computational hardware, the introduction of the Transformer architecture, and the availability of large-scale training data.

\begin{itemize}
\item \textbf{Computational Hardware:} With advancements in graphics processing units (GPUs) and specialized integrated circuits such as tensor processing units (TPUs), large-scale parallel computing has become markedly more efficient for deep learning training. In particular, NVIDIA’s CUDA architecture and Google’s TPUs have provided unparalleled computational power to accelerate the training of deep learning models. CUDA supports parallel computing, significantly enhancing data processing speed during neural network training, while TPUs optimize core operations such as matrix multiplication, reducing training times, and offering superior energy efficiency. These advancements in hardware provide the computational infrastructure necessary for the success of large-scale foundation models and enable AI models to scale from specialized tasks to more general multimodal tasks.
\end{itemize}

\begin{itemize}
\item \textbf{Transformer Architecture:} The introduction of the Transformer architecture proposed by \cite{vaswani2017attention}, the Transformer architecture leverages the self-attention mechanism to capture dependencies within sequences, overcoming the limitations of RNN and LSTM. The self-attention mechanism enables efficient parallel processing, making the Transformer highly effective in handling long-range dependencies. Specifically, the multi-head self-attention mechanism enables the model to learn different representations from various subspaces, thereby enhancing feature learning.
\end{itemize}

\begin{itemize}
\item \textbf{Large-scale Training Data:} In the data-driven era of AI, the proliferation of the internet and digitalization has made large-scale multimodal data (text, images, audio, video, etc.) available for model pretraining. The availability of large-scale data, coupled with the rise of self-supervised learning, has reduced the reliance on manually labeled data. Self-supervised learning, through tasks such as masked language modeling (MLM), enables models to train on unlabeled data by predicting missing parts of the input. Models like BERT and GPT have been pre-trained on large-scale text datasets via self-supervised learning, significantly improving their generalization capabilities and performance in various application scenes.
\end{itemize}

Moreover, the representation learning of foundation models typically follows two main paradigms: (1) Supervised pretraining, where models are trained on large-scale labeled datasets (e.g., ImageNet\cite{deng2009imagenet}) and then applied to downstream tasks through transfer learning; (2) Self-supervised learning, where models learn from unlabeled data by performing predictive tasks. BERT and the GPT series have achieved widespread success in NLP through self-supervised learning, and this paradigm is gradually extending to computer vision.

Together, these factors have firmly established foundation models as a core component of current and future AI technological developments. By gaining a deeper understanding of foundation models, their potential can be harnessed more effectively in solving complex problems.

\subsubsection{Transformer in foundation model}
The Vanilla Transformer\cite{vaswani2017attention} is a novel neural network architecture specifically designed to handle sequential data. The Transformer architecture relies entirely on the self-attention mechanism to capture dependencies in sequential data, making it highly effective in addressing long-range dependency problems. The Transformer consists of two main components: an encoder and a decoder, both of which are composed of multiple identical layers. Each layer contains two key sublayers: a multihead self-attention mechanism and a feed-forward neural network. In the encoder, the input sequence first passes through an embedding layer and then through multiple encoder layers to generate contextual representations. The decoder, in turn, takes the encoder's output and combines it with the target sequence embeddings to generate the final output through a step-by-step decoding process.

\paragraph{Self-Attention (SA) Mechanism }
The core of the Transformer success lies in the self-attention mechanism, which calculates the relevance between different positions in the input sequence to generate weighted representations. This operation is commonly referred to as scaled dot-product attention. The goal of self-attention is to capture the interdependencies between different positions in the sequence. Given an input sequence $X$ of length $N$, position encoding is first applied to obtain the position-aware input sequence:
\begin{equation}
Z=X \oplus P
\end{equation}
where $\oplus$ denotes element-wise summation, and $P$ is position embedding of input $X$.

The embedded sequence $Z$ is then projected in queries $Q$, keys $K$, and values $V$ using projection matrices ${W}_{Q}$, ${W}_{K}$, and ${W}_{V}$.
\begin{equation}
Q=Z W_{Q}, \quad K=Z W_{K}, \quad V=Z W_{V}
\end{equation}
The attention scores are computed by taking the dot product of $Q$ and $K$, followed by a softmax operation to normalize the scores:
\begin{equation}
\operatorname{Attention}(Q, K, V)=\operatorname{Softmax}\left(\frac{Q K^{T}}{\sqrt{d_{k}}}\right) V
\end{equation}
where $\sqrt{d_{k}}$ is the dimension of the key vector, used to scale the dot product to prevent excessively large values.

The self-attention mechanism allows each element in the input sequence to interact with every other element, including itself, enabling the Transformer to capture long-range dependencies from a global perspective, thereby improving its performance on long sequential data.

\paragraph{Multi-Head Self-Attention Mechanism (MHSA)}
Building on the self-attention mechanism, the Transformer introduces multi-head self-attention, which is key to its ability to capture diverse feature representations. Unlike single-head attention, multi-head self-attention enables the model to compute multiple attention mechanisms in parallel, with each focusing on a different aspect of the input sequence. Each attention head is computed independently:
\begin{equation}
{head}_{i}=\operatorname{Attention}\left(Q_{i}, K_{i}, V_{i}\right)
\end{equation}
where each head ${head}_{i}$ is an independent attention mechanism, and ${Q}_{i}$, ${K}_{i}$ and ${V}_{i}$ are the query, key, and value vectors after independent linear transformations.

The outputs of all heads are concatenated and passed through a linear projection:
\begin{equation}
\operatorname{MHSA}=\operatorname{Concat}\left(head_{1}, ..., head_{h}\right)W
\end{equation}
where $W$ is the projection matrix, and h denotes the number of attention heads.

Multi-head self-attention enables the model to capture diverse feature representations from multiple subspaces of the input sequence, significantly enhancing the model's representation capability. Each attention head can focus on different parts of the sequence, thereby improving the model’s ability to generalize to complex tasks.

\paragraph{Feed-Forward Neural Network (FFN)}
Each layer of the transformer encoder and decoder comprises not only MHSA but also FFN. The FFN operates independently on each position, applying a nonlinear transformation to each input position. The FFN consists of two linear transformations followed by a ReLU activation function. The design of the FFN, though simple, plays a crucial role in adding non-linearity to the model's overall structure. When combined with the multi-head attention mechanism, it gives the Transformer a powerful ability to handle complex sequential data.

\subsubsection{Vision Transformer (ViT) in vision foundation model}
Unlike foundation models in natural language processing (NLP), which have established a unified learning paradigm based on contextual prediction and have already captured a certain level of common sense knowledge, the development of vision foundation models is still in its early stages. There is no standardized paradigm for learning common sense in the visual world, which presents challenges in achieving generalization and robustness in feature representations. Traditional supervised learning approaches for vision foundation models rely heavily on high-quality labeled data, which limits the adaptability of the model to various tasks. In contrast, the self-supervised learning paradigm does not depend on explicit labels; it leverages methods such as contrastive learning to train models more efficiently on diverse large-scale datasets. Self-supervised learning has shown significant potential to build common sense visual knowledge. For example, contrastive learning methods such as MoCo\cite{he2020momentum} and SimCLR\cite{chen2020simple} use raw, unlabeled data by forming positive and negative pairs to learn general visual representations. This learning strategy, with its broad applicability in the visual domain, offers a promising path toward learning common sense knowledge in vision. The introduction of the Vision Transformer architecture has further unlocked the potential of vision foundation models, enhancing their ability to capture and model complex visual relationships.

The Vision Transformer, proposed by \cite{dosovitskiy2020image}, represents a novel approach to computer vision by using the transformer architecture originally designed for sequence modeling tasks. The core innovation of ViT lies in converting images into sequences of patches, treating each patch as a token. These patches are then linearly embedded and fed into the Transformer model, where the self-attention mechanism processes the relationships among the patches. This architecture enables ViT to capture long-range dependencies in images more efficiently than convolutional neural networks (CNNs), particularly when handling large-scale, high-resolution visual data.

A key feature of ViT is its partitioning of input images into fixed-size, nonoverlapping patches (e.g., $16\times16$ pixels), each of which is embedded into a vector. These embedded vectors, combined with positional encodings, serve as the input sequence for the transformer. By converting images into sequences, ViT can directly apply the multi-head self-attention mechanism to image data, eliminating the need for convolutional layers. This process enables ViT to capture global context across the image, which is particularly advantageous for complex image tasks.

The multi-head self-attention mechanism in ViT enables the model to compute representations in parallel across different subspaces for each image patch. Unlike convolution operations that rely on static filters, the self-attention mechanism dynamically computes filters based on the data, enabling ViT to adjust the receptive field flexibly. This adaptability empowers ViT to model both global and local features effectively, making it highly efficient in various visual tasks such as image classification, object detection, and segmentation. Similarly to the classical Transformer, each ViT layer also includes FFN in addition to multi-head self-attention. The FFN performs independent linear transformations and non-linear activations for each patch representation, thereby enhancing the model's ability to capture complex non-linear interactions. This design significantly improves the model's ability to generalize to complex image recognition tasks.

ViT excels at capturing long-range dependencies in images through its global self-attention mechanism, making it particularly suited for high-resolution image tasks. However, the quadratic growth in computational complexity with image resolution poses challenges for processing ultrahigh-resolution images. To address this, window-based attention mechanisms have been introduced, partitioning images into smaller windows and applying attention within each window, thereby reducing computational demands.

\subsection{Multimodal medical image analysis}

\begin{figure*}[!t]
  \begin{center}
  \includegraphics[width=1.0\linewidth]{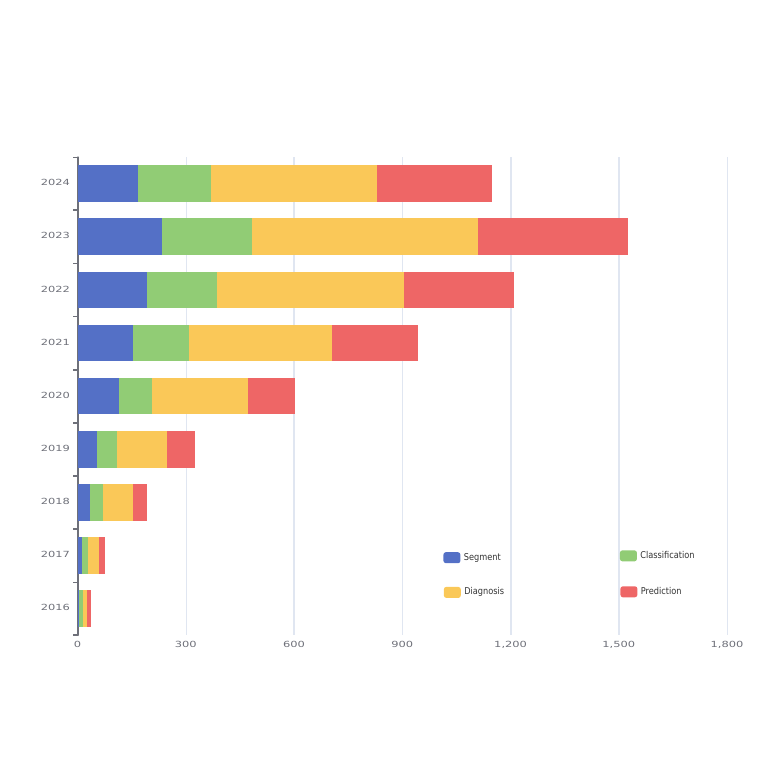}
  \end{center}
  \vspace{-1em}
  \caption{Result of analysis of literature with the keywords 'deep learning', 'multi modality' and following four types of tasks which are most widely studied: 'Segment', 'Classification', 'Diagnosis' and 'Prediction' on PubMed search engine, which is queried on August 9, 2024.}
  \label{statistician}
\end{figure*}

{\color{black}The integration of multimodal medical imaging has significantly advanced the technical methodologies in image analysis, improving diagnostic precision and treatment planning, which is particularly evident in the cardiovascular field, where the combination of different imaging modalities provides a more comprehensive understanding of cardiac structures and functions.
}

Advances in multimodal medical imaging have transformed clinical practice by improving diagnostic precision and treatment planning, particularly in cardiovascular medicine. Medical imaging techniques such as X-ray, computed tomography (CT) and magnetic resonance imaging (MRI) provide crucial complementary insights to the management of conditions such as coronary artery disease, heart failure, and aortic stenosis. For instance, while CT excels in identifying calcified plaques in coronary arteries, MRI is indispensable for evaluating myocardial tissue characteristics, such as fibrosis or inflammation. These complementary modalities illustrate the limitations of single-modality imaging, which often provides insufficient data for a comprehensive cardiovascular assessment. Studies have shown that the integration of multimodal imaging significantly improves outcomes in tasks such as cardiac tumor segmentation, myocardial infarction classification, and coronary artery disease diagnosis.

With the rapid advancement of medical image acquisition systems, numerous studies have emerged in recent years on multimodal medical image analysis. Medical imaging techniques, such as X-ray, CT, MRI, and positron emission tomography (PET), play an important role in the whole healthcare process, including diagnosis, staging, and planning of treatment, and offer different information in the process. For example, as noted in \cite{model_GB}, a CT image can be used in the diagnosis of dense structures such as bones, while an MR image provides soft tissue information. Thus, compared to multimodal images, images from a single modality tend to provide limited information when assisting in medical processes. Several studies \cite{ts_ZG,cpn_XL, YM} have shown that the use of multimodal images can lead to better performance compared to the use of images from a single modality in many medical tasks, including the detection of cervical cancer \cite{YM}, the classification of parotid neoplasms \cite{cpn_XL} and tumor segmentation \cite{ts_ZG}.
 \begin{figure}[ht]
  	\begin{center}
  		\includegraphics[width=\linewidth]{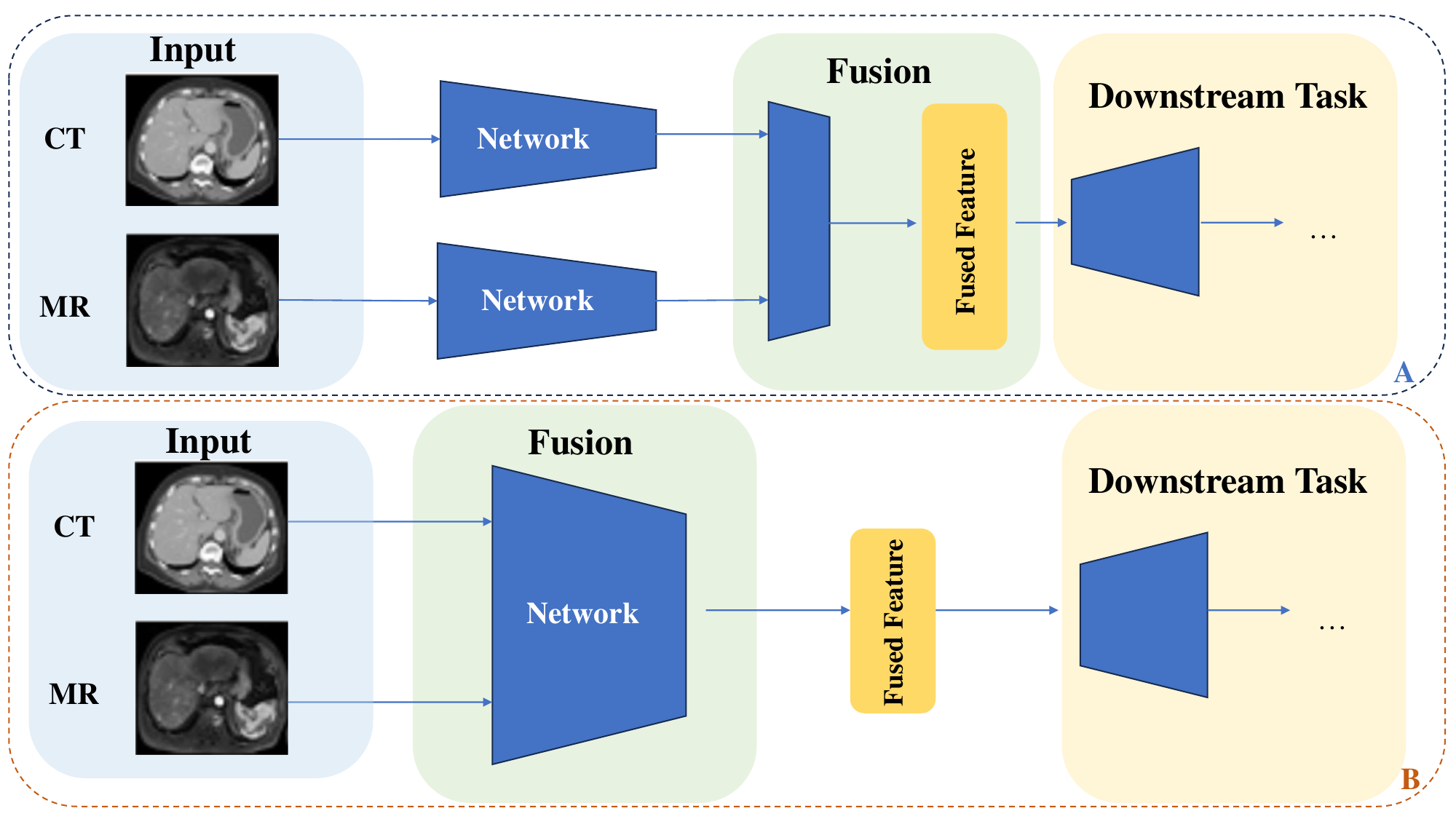}
  	\end{center}
  	\vspace{-1em}
  	\caption{Two different architectures to fuse images under various modalities. A. Architecture in which images are fused during the input process and transferred into the fused feature space; B. Architecture in which features are first extracted into different latent spaces and then fused together.}
  	\label{fuse_type}
  \end{figure}

To address the challenges of medical image fusion, various strategies have been developed, such as the use of wavelet transform \cite{wavelet}, curvelet \cite{curvelet}, non-subsampled contourlet transform \cite{nsct,model_GB}, fuzzy approaches \cite{fuzzy_SD,fuzzy_PB}, hidden Markov fields \cite{hmf}, belief functions \cite{believe_m,believe_f}, and machine learning techniques \cite{ml_NS,ml_HC,ml_AV}. However, with the rapid development of deep learning, compared to traditional methods, deep learning-based approaches have demonstrated outstanding performance in various tasks, including segmentation \cite{seg_bt,seg_ga,seg_non_paired_organ,seg_sl}, classification \cite{class_az,class_bc,class_md}, cross-modality generation \cite{gen_mri,gen_video}, diagnosis \cite{diag_az,diag_koa,diag_ras}, prediction \cite{pred_cir,pred_mvi,pred_nac}, and other tasks \cite{eval_doc}. We analyzed the literature using keywords such as 'deep learning' and'multimodality', focusing on the following four most widely studied tasks: segmentation, classification, diagnosis, and prediction. The analysis was carried out using the PubMed \footnote{https://pubmed.ncbi.nlm.nih.gov/} search engine on 9 August 2024, with the results shown in Fig. \ref{statistician}. Our findings indicate that the number of articles in these fields has increased every year from 2016 to 2024, suggesting that methods utilizing multimodal medical images are gaining increasing attention. Among these, the diagnosis task is the most widely studied.

Based on how images from different modalities are fused, we categorize these methods into two types, as illustrated in Fig. \ref{fuse_type}. In Fig. \ref{fuse_type}-B, images are fused at the input stage and then transformed into a fused feature space \cite{if_bt,if_bg,if_bt_brats_2017,if_bt_miccai,matr,m4oe,mm_fusion_seg}, while in Fig. \ref{fuse_type}-A, features are first extracted into separate latent spaces and then fused together \cite{of_bi,of_eb,of_rbts}. The latter approach leverages more information from each individual modality but lacks the ability to capture the corresponding relationships between the modalities, compared to the method shown in Fig. \ref{fuse_type}-B.

\subsection{Large scale medical datasets}
{\color{black}The advancement of foundational models in medical imaging critically depends on the availability of large, diverse, and multimodal datasets, which improve diagnostic precision and treatment planning accuracy of trained clinical deep learning models.

In clinical practice, particularly within the cardiovascular discipline, accurate diagnosis and effective treatment planning are highly dependent on detailed and comprehensive medical imaging.}
Another key factor that limits the representation capacity of foundational models in medical imaging is the size and diversity of available data sets. In the era of general artificial intelligence, large-scale datasets are crucial to developing robust foundational models in medicine. The quality, scale and diversity of the data directly influence the generalizability of these models, making pretraining on comprehensive datasets particularly important\cite{scale_law}. In the medical field, the accumulation of large-scale data serves as a rich source of knowledge for training models that can perform diverse tasks in multiple domains.

Although unimodal data can provide valuable insight, the integration of multimodal data has been shown to offer significant advantages. By combining different types of information, multimodal data provide a more comprehensive perspective, leading to more accurate diagnoses, better treatment planning, and improved patient outcomes. However, the challenge lies in the acquisition and integration of multimodal data. Compared to unimodal data, multimodal data collection is generally more complex, and standardization is difficult due to the heterogeneous and intricate nature of its sources, each with unique structures and formats. This complexity increases the difficulty of data harmonization and model training. Furthermore, the scarcity of annotations in multimodal datasets exacerbates these challenges, making it crucial to optimize the use of existing data.

Given these challenges, it is imperative to develop methods that effectively utilize the available multimodal data to enhance the performance of foundational models in medicine, allowing better generalization between tasks and datasets. By maximizing the use of multimodal data, researchers can push the limits of foundational models, paving the way for more precise and personalized healthcare solutions.

In the following, we will introduce three types of dataset that we believe can be used for the training and testing of MMFMs in order of data modality: plain text datasets, medical image datasets (including segmentation masks or labels), and image-text pair datasets.

\begin{table*}
\centering
\caption{Text Modality Datasets}
\label{tab:text_datasets}
\begin{tabular}{p{2cm}@{\hskip 10pt}p{3cm}@{\hskip 10pt}p{1.5cm}@{\hskip 10pt}p{2.5cm}@{\hskip 10pt}p{2.5cm}@{\hskip 10pt}p{4cm}}
\toprule
\textbf{Dataset Name} & \textbf{Type} & \textbf{Modality} & \textbf{Purpose} & \textbf{Size} & \textbf{Key Features} \\
\midrule 
\textbf{MedNLI}\cite{MedNLI} & Natural language inference & Text only & Reasoning in medical texts & 14K sentence pairs & Useful for developing NLU models in the medical domain \\
\textbf{SEER}\cite{SEER} & Epidemiological cancer data & Epidemiological Data & Cancer epidemiology, public health research & 23K records & Extensive data on cancer incidence, survival, and treatment outcomes \\ 
\textbf{MIMIC-III}\cite{MIMIC3} & Electronic health records (EHR) & Text, Time-Series & Clinical data analysis, NLP, patient care & 730K records & Rich clinical data, widely used in medical research \\ 
\textbf{MeQSum}\cite{MeQSum} & Medical question summarization & Text only & NLP research in healthcare & 1K questions & Focuses on summarizing complex medical queries \\ 
\textbf{HealthCareMagic}\cite{hcm} & Medical question-answer pairs & Text only & Healthcare NLP & 226K question-answer pairs & Derived from online medical consultation, useful for question-answering models \\
\bottomrule
\end{tabular}
\end{table*}

\subsubsection{plain text datasets}

Datasets are shown in Table. \ref{tab:text_datasets} provide diverse resources for various tasks in medical natural language processing and clinical research, which can improve understanding in the medical domain for the multimodal foundation representation model. MedNLI\cite{MedNLI} is specifically designed for natural language inference in clinical settings, helping bridge the gap between general NLP models and specialized medical applications using transfer learning and domain knowledge. SEER\cite{SEER}, on the other hand, focuses on epidemiological cancer data, providing extensive records on cancer incidence and outcomes, which are essential for public health and cancer research. MIMIC-III\cite{MIMIC3} stands out for its large-scale electronic health records, widely used for clinical data analysis and NLP tasks, covering structured and unstructured data. MeQSum\cite{MeQSum} facilitates the summarization of medical queries, addressing the need for concise responses to complex medical questions, thus contributing to advancements in summarization models. Finally, HealthCareMagic\cite{hcm} provides a rich dataset of medical question-answer pairs derived from online consultations, which is valuable for training healthcare-specific question-answering models.

\begin{table*}
\centering
\caption{Image Modality Datasets}
\label{tab:image_datasets}
\begin{tabular}{p{2cm}@{\hskip 10pt}p{3cm}@{\hskip 10pt}p{2cm}@{\hskip 10pt}p{3cm}@{\hskip 10pt}p{2cm}@{\hskip 10pt}p{4cm}}
\toprule
\textbf{Dataset Name} & \textbf{Type} & \textbf{Modality} & \textbf{Purpose} & \textbf{Size} & \textbf{Key Features} \\
\midrule
\textbf{MC-CXR \& SZ-CXR}\cite{cxr} & Chest X-rays & X-ray & Disease detection and classification & 138(MC)/662(SZ) images & Focus on diagnostic challenges in chest X-ray interpretation \\ 
\textbf{CBIS-DDSM-CALC \& MASS}\cite{cdcm} & Mammography images & Mammography & Breast cancer detection research & 1K images & Detailed annotations for calcifications and masses \\ 
\textbf{MMR Datasets}\cite{mmr} & Various medical image types & Multimodal & Multimodal medical research & 166K images & Multimodal dataset covering multiple specialties \\ 
\textbf{MMR-Colon Pathology}\cite{mmr} & Pathology images & Pathology & Disease diagnosis and treatment planning & 107K images & Focuses on colon pathology \\ 
\textbf{MMR-Dermatoscopy}\cite{mmr} & Dermatoscopy images & Dermatoscopy & Skin disease diagnosis & 10K images & Contains dermatoscopy images for skin disease diagnosis \\ 
\textbf{MMR-Retinal OCT}\cite{mmr} & Retinal OCT images & Retinal OCT & Eye disease diagnosis & 1K images & Retinal OCT scans for diagnosing retinal conditions \\ 
\textbf{MMR-Chest X-ray}\cite{mmr} & Chest X-ray images & X-ray & Lung disease diagnosis & 5K images & Chest X-rays with focus on lung-related diseases \\ 
\textbf{MMR-Breast Ultrasound}\cite{mmr} & Ultrasound images & Ultrasound & Breast cancer detection & 780 images & Ultrasound images for breast cancer diagnosis \\ 
\textbf{MMR-Blood Cell Microscope}\cite{mmr} & Microscope images & Microscopy & Blood cell analysis & 17K images & Blood cell microscope images for various hematological conditions \\ 
\textbf{MMR-Coronal Abdominal CT}\cite{mmr} & Abdominal CT images & CT Imaging & Abdominal disease diagnosis & 23K images & Coronal Abdominal CT scans for abdominal pathology \\ 
\textbf{CLIP-Driven Universal Model}\cite{cdum} & CT images & CT Imaging & Tumor segmentation and detection & 3K training CT scans, 6K external scans & Model developed from 14 datasets targeting 25 organs and 6 tumor types \\ 
\textbf{AbdomenAtlas-8K}\cite{aba8k} & Annotated CT Volumes & CT Imaging & Organ segmentation & 8K CT volumes & Focus on 8 key abdominal organs including liver, spleen, kidneys \\
\textbf{AbdomenAtlas 1.1} & Annotated CT Volumes & CT Imaging & Multi-organ and tumor segmentation & 9K CT volumes & Detailed voxel-wise annotations of 25 anatomical structures and 7 tumor types \\ 
\textbf{MedSAM}\cite{medsam} & Medical image mask pairs & Multimodal (10 imaging modes) & Segmentation and cancer detection & 1.5M image-mask pairs & Covers 10 imaging modalities and 30+ cancer types \\ 
\textbf{SAM-Med3D}\cite{sammed3d} & 3D medical image segmentation & CT, MRI & 3D image segmentation for organs and lesions & 21K images, 131K masks & Covers 27 modalities and 240+ target categories \\ 
\textbf{EyeFound}\cite{EyeFound} & Ophthalmology images & 11 clinical imaging modalities & Eye disease research & 3M images & Collected from 227 hospitals, includes CFP, FFA, ICGA, OCT, ultrasound, etc. \\
\textbf{RETFound}\cite{RETFound} & Retinal images (CFP, OCT) & Fundus Imaging, OCT & Eye disease detection & 904K CFP, 736k OCT & Focus on fundus and OCT images for eye condition diagnosis \\ 
\textbf{Swin UNETR}\cite{swinunetr} & CT images & CT Imaging & Pretraining for 3D segmentation & 5K CT volumes & Includes chest, abdomen, and head/neck volumes \\
\textbf{Disruptive Autoencoders}\cite{da} & CT and MRI images & CT, MRI (T1, T2, FLAIR, T1ce) & Pretraining for 3D segmentation and autoencoders & 10K 3D volumes & Combines 5 modalities for diverse segmentation tasks \\
\bottomrule
\end{tabular}
\end{table*}

\subsubsection{medical image datasets}

The datasets listed in Table. \ref{tab:image_datasets} provide diverse imaging modalities, which are crucial for building medical foundation models that can generalize across various domains. MC-CXR \& SZ-CXR\cite{cxr} and MMR-Chest X-ray\cite{mmr} focus on chest X-rays, facilitating disease classification and detection, which is fundamental for learning X-ray-based diagnostics. CBIS-DDSM-CALC \& MASS\cite{cdcm} and MMR-Breast Ultrasound\cite{mmr} offer high-quality annotated imaging datasets for breast cancer detection, advancing model accuracy in mammography and ultrasound interpretation. MMR Datasets\cite{mmr}  and its subcategories span multiple modalities, including pathology, retinal OCT, and dermatoscopy, contributing to multimodal model training that generalizes between specialties. Large data sets such as EyeFound\cite{EyeFound} and RETFound\cite{RETFound}, with millions of ophthalmology images, support model training for the detection of retinal diseases, while specialized datasets such as MedSAM\cite{medsam} and SAM-Med3D\cite{sammed3d} provide vast image mask pairs across modalities, key to segmentation tasks. CLIP-Driven Universal Model\cite{cdum} and Disruptive Autoencoders (DAE)\cite{da} emphasize tumor segmentation, using multi-organ and multimodal data, improving 3D image segmentation models for oncology applications. 

Segmentation tasks, as exemplified by datasets such as MedSAM, SAM-Med3D, and AbdomenAtlas, focus on delineating structures within medical images, such as organs or tumors. These tasks help models develop a detailed understanding of spatial patterns, enabling them to recognize fine-grained anatomical features, boundaries, and spatial relationships between tissues. For example, MedSAM's multimodal image-mask pairs offer essential data to refine the recognition of spatial patterns of models in diverse types of imaging, which is critical for surgical planning or oncology treatments.

However, detection and classification tasks presented in datasets like MC-CXR, CBIS-DDSM, and MMR-Breast Ultrasound emphasize the recognition of high-level semantic patterns, such as the presence of disease or the classification of medical conditions. These datasets aid models in extracting meaningful features from images, which enhances their ability to identify abnormalities and classify diseases accurately. For instance, MMR-Breast Ultrasound provides data on breast cancer detection, sharpening models' semantic understanding by learning disease-related features in ultrasound images.

\begin{table*}
\centering
\caption{Image-Text Pair Modality Datasets}
\label{tab:image_text_datasets}
\begin{tabular}{p{2cm}@{\hskip 10pt}p{3cm}@{\hskip 10pt}p{2.5cm}@{\hskip 10pt}p{3cm}@{\hskip 10pt}p{2.5cm}@{\hskip 10pt}p{3cm}}
\toprule
\textbf{Dataset Name} & \textbf{Type} & \textbf{Modality} & \textbf{Purpose} & \textbf{Size} & \textbf{Key Features} \\ 
\midrule
\textbf{ROCO}\cite{roco} & Radiology images with captions & X-ray, CT, MRI, Ultrasound & Multimodal learning, image-captioning & 87K image-text pairs & Variety of radiology images with associated captions \\ 
\textbf{MedICaT}\cite{medicat} & Medical images with captions \& documents & X-ray, CT, MRI & Vision-language tasks, VQA, report generation & 217K image-text pairs & Image-caption pairs with full-text articles \\ 
\textbf{PMC-OA}\cite{pmcoa} & Biomedical image-text pairs & CT scans, MRIs, X-rays, ultrasounds, microscopy, and other diagnostic images & Image-text retrieval, medical image classification, and VQA & 1.6M image-text pairs & The figures and captions from PubMed Central \\ 
\textbf{ChiMed-VL}\cite{chimedvl} & Chinese medical text and images & X-ray, Ultrasound & Multimodal research in Chinese medical tasks & 580K image-text pairs & Focus on Chinese medical data \\
\textbf{FFA-IR}\cite{ffair} & Fundus fluorescein angiography images, medical reports and lesion annotations & Fundus fluorescein angiography & Ophthalmology research & 10K image-text pairs & Detailed annotations for eye-related conditions \\
\textbf{PadChest}\cite{FNL_Bustos} & Chest X-rays with annotations & X-ray & Medical image analysis, pulmonary disease & 109K image-text pairs & Multi-label annotations per image \\ 
\textbf{MIMIC-CXR}\cite{MIMICCXR} & Chest X-ray images with reports & X-ray & Medical imaging, NLP & 377K images 227K text & Extensive dataset with linked radiology reports \\ 
\textbf{CT-RATE}\cite{CT-RATE} & CT images with radiomics features & CT Imaging & Image analysis, radiomics research & 50K image-text pairs & Focus on radiomics features for oncology \\
\textbf{OpenPath}\cite{OpenPath} & Whole-slide pathology images paired with natural language descriptions & Whole-slide Imaging & Digital pathology, histopathological analysis & 208K image-text pairs & Largest pathology image dataset for cancer diagnosis research \\
\textbf{Quilt1M}\cite{Quilt1M} & Paired Pathology-image-text & Histopathology & Medical image classification & 1M image-text pairs & High-resolution images for detailed analysis \\ 
\textbf{PEIR GROSS \& IU X-RAY}\cite{pgiu} & Gross pathology and radiology images & X-ray(IU X-RAY), Pathology(PEIR GROSS) & Educational and research purposes & 7K(PEIR GROSS) image-text pairs /6K 7,470 images and 3,955 reports & Used for medical image classification and educational tools \\
\textbf{SLAKE}\cite{SLAKE} & Multimodal question-answer pairs & X-ray, CT, MRI, Ultrasound & Medical VQA & 7K image-text pairs & Combines image data with text for VQA tasks \\
\textbf{PathVQA}\cite{pathvqa} & Pathology visual question answering & Pathology & VQA tasks in digital pathology & 32K image-text pairs & Focuses on understanding and analyzing pathology images \\
\textbf{VQA-RAD}\cite{VQARAD} & Radiology visual question answering & X-ray & Medical VQA in radiology & 2K image-text pairs & Emphasizes radiology interpretation in VQA format \\
\textbf{RET-CLIP}\cite{retclip} & Retinal images and text triplets & Fundus Imaging & Retinal disease diagnosis & 193K image-text pairs & Patient-level retinal image and clinical diagnosis triplets \\
\bottomrule
\end{tabular}
\end{table*}

\subsubsection{image-text pair datasets}

The image-text pair datasets are listed in Table. \ref{tab:image_text_datasets} are pivotal to advance the development of MMFMs. ROCO\cite{roco} and MedICaT\cite{medicat} provide radiology images paired with captions, which facilitate tasks such as image-captioning and cross-modal learning in radiology and general medical fields. PMC-OA\cite{pmcoa} stands out with its scale, providing more than 1.6 million biomedical image-text pairs, which support image-text retrieval and medical image classification. Datasets like ChiMed-VL\cite{chimedvl} and FFA-IR\cite{ffair} expand the capability of the model into specialized medical fields, such as Chinese medical imaging and ophthalmology, respectively, offering substantial amounts of annotated data for multimodal research. CT-RATE\cite{CT-RATE} and RET-CLIP\cite{retclip} further enhance the training of models in radiology and retinal disease diagnosis, by combining patient-level data with detailed reports. Furthermore, PathVQA\cite{pathvqa} and SLAKE\cite{SLAKE} focus on visual question answering (VQA) in medical contexts, which is key for models that aim to integrate and interpret medical imagery along with question-based reasoning. These datasets collectively provide the diverse multimodal inputs necessary for training models that can be generalized across different medical domains and tasks.

\section{Medical Multimodal Vision Foundation models}

MMVFMs have emerged as a pivotal framework for advancing medical image analysis by leveraging multimodal data from various imaging modalities. Central to the development of these models is the use of proxy tasks during pretraining, which serve as surrogate objectives designed to expose models to diverse data representations and latent structures inherent in medical images. These proxy tasks not only enable models to generalize across different downstream tasks, but also enhance their capacity to extract clinically relevant features from complex datasets. The design of proxy tasks is critical, as it directly influences the model's ability to capture fine-grained, modality-specific nuances and cross-modal correlations, which are essential in medical imaging.

\begin{figure}[t]
  	\begin{center}
  		\includegraphics[width=1.0\linewidth]{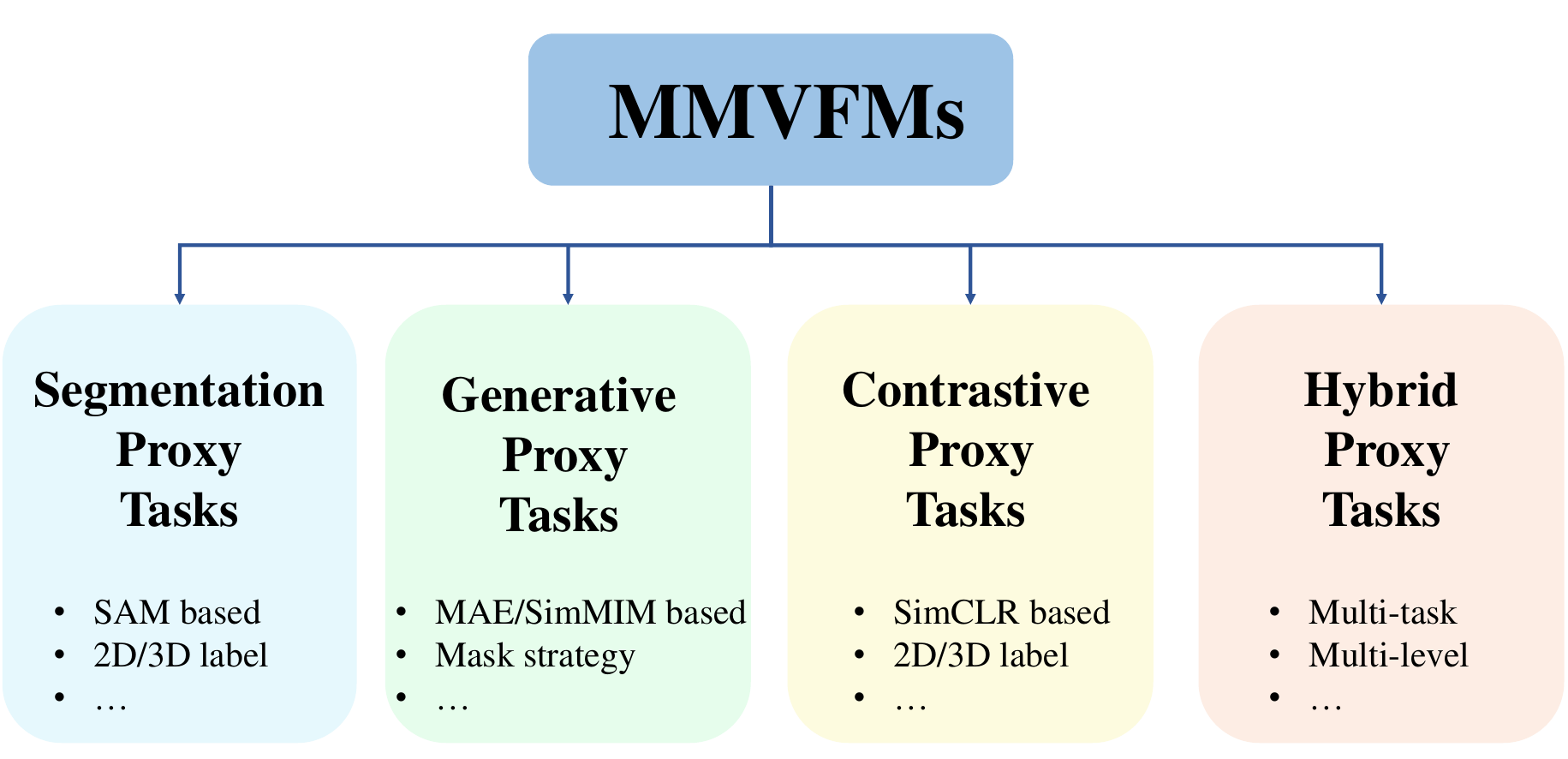}
  	\end{center}
  	\vspace{-1em}
  	\caption{A comprehensive overview of the four primary proxy task categories within MMVFMs. The categories include Segmentation Proxy Tasks, Generative Proxy Tasks, Contrastive Proxy Tasks, and Hybrid Proxy Tasks.}
  	\label{mmvlfm}
  \end{figure}

As illustrated in Fig. \ref{mmvlfm}, proxy tasks in MMVFMs are broadly categorized into four types: Segmentation Proxy Tasks, Generative Proxy Tasks, Contrastive Proxy Tasks, and Hybrid Proxy Tasks. Segmentation proxy tasks, for example, focus on delineating anatomical structures and pathological regions, providing a foundational understanding of spatial relationships within 2D and 3D medical images. Generative proxy tasks, on the other hand, emphasize reconstructive capabilities, utilizing a mask-reconstruction strategy to challenge models with the task of recovering occluded or masked parts of an image. Contrastive proxy tasks focus on distinguishing between different modalities or anatomical regions, ensuring the model can learn robust feature representations across varied imaging types. Finally, hybrid proxy tasks combine multiple learning paradigms, optimizing model performance through a multifaceted approach that incorporates multitask and multilevel learning strategies. These proxy tasks collectively drive the success of MMVFMs, equipping them with the versatility required to excel in complex clinical scenarios, including segmentation, classification, detection, and beyond.

\subsection{Segmentation Proxy Task}
Segmentation tasks capture and represent critical anatomical and pathological structures in multimodal medical images that often have high clinical relevance. The primary goal of segmentation is to divide the image into regions that correspond to key structures, such as organs, lesions, or tumors. Through segmentation tasks, models learn to differentiate these structures, providing foundational visual representations for other pixel-level downstream tasks, including classification and registration. One of the primary reasons for using segmentation as a proxy task for foundation models is its strong representation capability. Segmentation tasks require models to capture both local details and global context, enabling them to extract cross-modal consistent features from multimodal data. For example, although CT and MRI images may differ in contrast and texture, the underlying anatomical structures are often consistent, and segmentation tasks help models capture this consistency across different modalities.

With the introduction of the SAM\cite{kirillov2023segment}, foundation models have seen significant progress in the field of image segmentation. SAM is a general, promptable segmentation model capable of generating high-quality segmentation masks based on predefined prompts such as points, boxes, or text. Its primary advantage lies in its zero-shot learning capability, which allows it to generate accurate segmentation masks even on unseen data. The core architecture of SAM includes an image encoder, a prompt encoder, and a mask decoder. The image encoder uses a large Vision Transformer model to extract features from the input image, while the prompt encoder encodes user-provided prompts, and the mask decoder combines the image features with the prompt information to generate the corresponding segmentation mask.
  
However, the direct application of SAM to medical image segmentation has not yielded the desired results, primarily due to significant differences in data characteristics between natural and medical images. Medical images typically require higher resolution and more precise edge detection, particularly in identifying critical anatomical structures. Furthermore, medical image datasets are relatively limited, and the cost of annotation is substantial. To better meet the requirements of medical image segmentation, researchers often need to fine-tune SAM or incorporate specific preprocessing steps tailored to the characteristics of medical images, thereby enhancing the foundation model's ability to represent and segment complex multimodal medical images.

To address existing limitations in medical image segmentation, this article introduces MedSAM\cite{medsam}, a foundation model specifically designed for universal medical image segmentation. The MedSAM model was trained on a large-scale medical image dataset comprising 1,570,263 image-mask pairs, which encompassed 10 imaging modalities and more than 30 types of cancer. This diverse dataset enables MedSAM to learn rich representations of medical images, capable of managing various anatomical structures and lesions across different modalities, thus improving its adaptability and performance across multiple tasks. MedSAM employs a prompt-adjustable segmentation approach that allows users to specify regions of interest by providing bounding boxes or point prompts. Compared to fully automated models, this method offers enhanced flexibility and adaptability, enabling it to accommodate a variety of task requirements and imaging modalities. In both internal and external validation tasks, MedSAM significantly outperforms existing state-of-the-art (SOTA) segmentation foundation models, such as SAM, and in many instances, it performs comparably to or even better than specialized models like U-Net and DeepLabV3+. These results underscore MedSAM's significant potential as a foundational model for multimodal, multitask segmentation. Likewise, SAM-Med2D\cite{cheng2023sam} is a fine-tuned version of the original SAM model specifically adapted for 2D medical image segmentation. The model was trained on approximately 4.6 million images and 19.7 million masks, covering a wide range of medical imaging modalities and anatomical structures. By integrating various prompt modes (including points, bounding boxes, and masks) and fine-tuning the original SAM encoder and decoder, SAM-Med2D exhibited significant performance enhancements across multiple datasets. In particular, its extensive evaluation across nine MICCAI 2023 challenge datasets demonstrated robust generalization capabilities beyond the training data.

Despite the strong performance of MedSAM and SAM-Med2D in most tasks, there are still areas for improvement, especially in handling complex anatomical structures or regions with unclear boundaries. Moreover, these models cannot directly handle 3D medical images, as the original SAM architecture was primarily designed for 2D natural images and is less effective in capturing spatial information in 3D medical data. These challenges highlight the need for further optimization of these models to better address the complexities of medical imaging in the future.

The limitations of SAM stem mainly from its design for 2D images. The original 2D structure neglects the spatial information inherent in 3D modalities, leading to a lack of deep understanding of medical images. This limitation restricts its application in fields that require three-dimensional comprehension of medical images. To overcome this limitation, SAM2\cite{dong2024segment} extends the original SAM by introducing video input support, enabling the segmentation of any object in a video or image, even if the object or visual domain is previously unseen, without the need for custom adaptation. This makes SAM2 a more versatile tool, offering greater flexibility for various tasks. 
SAM2 introduces the potential to treat 3D medical images as videos, providing a solution for 3D image segmentation. In a comprehensive evaluation of SAM2 applied to medical images, researchers demonstrated both the potential and limitations of SAM2 in 2D and 3D medical image segmentation. Using a bidirectional propagation strategy, SAM2 shows significant advantages in 3D image segmentation. This strategy improves segmentation accuracy by propagating information both forward and backward from an annotated slice. MedSAM-2\cite{zhu2024medical} also follows the philosophy of treating 3D medical images as videos, developing a unique memory bank and a weighted selection strategy that allows the model to automatically segment similar objects in all subsequent images after only a single prompt. This capability was difficult to achieve with previous models, especially for image sequences without temporal relationships. MedSAM-2 can accurately segment all subsequent images with just one prompt. This method not only applies to 3D medical images, but also enables the "one-prompt segmentation" capability across a series of independent 2D medical images, significantly reducing user interaction. In evaluations on 26 tasks, MedSAM-2 not only performed exceptionally well on training data, but also demonstrated superior generalization across a wide range of medical image segmentation tasks.

However, despite the impressive performance of SAM2 in multi-frame 3D segmentation, there are cases where over-segmentation errors in the initial slice can accumulate during propagation, leading to suboptimal final segmentation results. This highlights that while SAM2 has made strides in adapting to 3D segmentation, there are still challenges that need to be addressed.

Another typical approach to segmenting 3D multimodal medical images is the 3DSAM-adapter\cite{gong20243dsam}, a method that holistically adapts SAM from 2D to 3D for medical image segmentation tasks. The 3DSAM-adapter modifies the original 2D SAM architecture to support volumetric inputs by introducing a visual sampler instead of positional encoding, enabling the prompt and image embeddings to share the same semantic features. This strategy effectively addresses the over-smoothing problem that can arise when elevating dimensionality and improves the model's robustness to noisy prompts. To maintain the model's lightweight nature, the 3DSAM-adapter adopts a multilayer aggregation mechanism in the mask decoder, replacing all 2D convolutions with 3D convolutions, significantly enhancing segmentation accuracy when generating 3D masks. Although the model adds 7.79\% more parameters, most of the pre-trained weights are retained, allowing it to efficiently capture 3D spatial patterns in medical images. In experiments carried out on several public tumor segmentation datasets, the 3DSAM-adapter demonstrated superior segmentation performance. For example, the segmentation accuracy of the model improved by 8.25\%, 29.87\%, and 10.11\% in the segmentation tasks of kidney, pancreas, and colon cancer, respectively. However, challenges remain when dealing with unseen modalities or tumors with complex shapes. The method also increases computational complexity, and its practical efficiency in clinical applications requires further improvement.

MA-SAM\cite{chen2024ma} achieves significant performance improvements in 3D multimodal medical image segmentation through innovative parameter-efficient fine-tuning strategies and 3D adapter design. It utilizes the FacT technique, where only a small portion of weight increments is updated during fine-tuning, while most of SAM's pre-trained weights are retained. This approach not only reduces computational cost, but also preserves the general knowledge of natural image tasks, ensuring robust performance across various tasks. Similarly to the 3DSAM-adapter, MA-SAM incorporates 3D adapters within the transformer blocks of the image encoder, enabling it to effectively extract 3D spatial information from input data. These adapters allow the original 2D backbone to handle 3D medical image data, improving the model's performance in both volumetric and temporal data processing. This design enhances the modality-agnostic nature of the model, allowing it to share common features between different imaging modalities without depending on specific modality characteristics. MA-SAM has been comprehensively evaluated in multiple medical image segmentation tasks in CT, MRI, and surgical video data. The results show that MA-SAM significantly outperforms many state-of-the-art 3D segmentation methods, such as nnU-Net, with outstanding performance, particularly in tumor segmentation tasks. In addition, MA-SAM exhibits strong generalization capabilities and adaptability to complex medical scenarios.

Despite the strong performance of MA-SAM in most tasks, the adapter module's ability to extract 3D spatial features is still not as effective as a purely 3D architecture. In specific complex 3D medical image segmentation tasks, the accuracy of the model still has room for improvement. To address these challenges, SAM-Med3D\cite{wang2023sammed3d} makes significant advancements in 3D medical image segmentation through a fully redesigned 3D architecture and large-scale data training. SAM-Med3D completely redesigns the SAM architecture into a fully learnable 3D structure, enabling the direct capture of 3D spatial information. Unlike previous methods that process 3D data slice by slice, SAM-Med3D employs full 3D convolutions, positional encoding, and self-attention mechanisms, excelling in capturing intricate 3D spatial details. Compared to 2D methods, SAM-Med3D requires significantly fewer prompt points during inference while still achieving higher dice scores. This efficiency improvement means that, in clinical settings, doctors and professionals can complete segmentation tasks faster using SAM-Med3D. Trained in a large-scale dataset with 247 categories, SAM-Med3D demonstrates strong generalization capabilities, handling various anatomical structures and lesions. This not only makes it highly effective in processing 3D medical images of different anatomical structures, but also shows its remarkable adaptability when faced with complex, unseen lesions.

Similarly to SAM-Med3D, SegVol\cite{du2023segvol} is a foundational model that supports universal and interactive 3D medical image segmentation. It combines large-scale data training, multiple prompt fusion techniques, and computational efficiency optimizations to achieve breakthrough advancements in 3D medical image segmentation. SegVol is trained in 90,000 unlabeled CT volumes and 6,000 labeled CT volumes, supporting the segmentation of over 200 anatomical categories. By integrating both semantic and spatial prompts, SegVol enables high-precision segmentation across various anatomical structures. One of the key innovations of SegVol is the introduction of a "Zoom-out-Zoom-in" strategy, which effectively reduces computational costs while maintaining segmentation accuracy. This mechanism initially reduces the input volume size to produce a coarse segmentation, followed by a high-precision inference on the identified region of interest (ROI). This approach allows SegVol to efficiently handle large-volume 3D medical images with both speed and precision. Additionally, SegVol supports various types of prompts, including spatial prompts (e.g., points and bounding boxes) and semantic prompts (e.g., textual descriptions). The combination of these prompts significantly enhances the model's performance, especially in generalizing across datasets. In a comprehensive evaluation of 22 anatomical segmentation tasks, SegVol outperformed other competing models in 19 tasks, with improvements up to 37.24\%. Despite SegVol's strong generalization and efficiency, there is still room for improvement when dealing with unseen modalities or complex tumor shapes.

Despite the strong performance of SAM-Med3D and SegVol in general organ and tumor segmentation tasks, they still face limitations in specific clinical tasks. Supervised Finetuning (SFT), while effective at adapting to task-specific needs, often leads to the degradation of the general knowledge stored in the original foundation model. To address this issue, \cite{wang2024sam} introduces the SAM-Med3D-MoE model, which integrates the Mixture of Experts (MoE) technology with the foundational model to successfully implement non-forgetting learning in foundational representation tasks, significantly improving the model's task adaptability and performance. SAM-Med3D-MoE innovatively incorporates MoE technology, combining task-specific fine-tuned models with the foundational model. By training a lightweight gating network to process images and prompt embeddings, the model generates confidence scores for each expert model. In addition, a selection strategy is introduced to adaptively combine the output of the foundational model and the expertise of the specific tasks, allowing the model to maintain excellent performance in specific tasks without compromising its general capabilities. Although SAM-Med3D-MoE shows excellent performance in handling complex tasks, the increase in the number of expert models may affect the training and inference speed. Future research could explore dynamically adjusting the selection strategy parameters based on specific scenarios to improve the model's adaptability.

Currently, segmentation models based on convolutional neural networks and Transformers are typically small in scale, with parameter sizes in the tens of millions, making it difficult to scale them up to handle larger datasets or tasks. To address this issue, \cite{huang2023stu} introduces STU-Net, a scalable and transferable medical image segmentation model. STU-Net is based on the nnU-Net framework and incorporates several key improvements. First, residual connections were introduced to alleviate the problem of gradient vanishing, which enhances the model's ability to scale in depth. Second, the upsampling strategy was optimized to improve the transferability of the model's weights, allowing it to adapt to various task requirements. These improvements not only enhance the model's adaptability to multimodal tasks but also make it easier to scale the model to larger datasets. Experimental results show that as the model size increases, STU-Net's performance improves significantly on large-scale datasets. For example, in experiments on the TotalSegmentator dataset, the performance of STU-Net consistently improves as the number of parameters increases. In practical applications, STU-Net has demonstrated excellent generalization capabilities in multiple downstream tasks, particularly multimodal medical image segmentation tasks such as CT and MRI. By pretraining on large-scale datasets, STU-Net achieves strong results in both direct inference and fine-tuning tasks, demonstrating the potential of large-scale pretrained models in medical image segmentation and offering new insights for future research in medical image analysis.

Segmentation tasks play a crucial role in medical image analysis by directly capturing structural information from images and generating clear boundaries and region delimitations. These results not only assist in the current task but also serve as the foundation for other downstream tasks, such as multimodal fusion and contrastive learning. However, segmentation faces challenges when applied across different modalities of medical images, as significant variations in contrast, resolution, and noise may cause the model to perform well in one modality but poorly in another. Ensuring cross-modal consistency and avoiding bias is a key challenge. Furthermore, noise and artifacts in medical images can impact segmentation accuracy, requiring models to handle these issues carefully to avoid incorrect segmentation. These challenges underscore the need for segmentation models with robust generalization capabilities and adaptability across modalities in foundational model tasks.

\subsection{Generative Proxy Task }
Generative tasks play a crucial role in multimodal image analysis, typically involving the reconstruction of complete data from partial or missing information. These tasks are foundational to the construction of models, as they train models to generate high-fidelity images or features, enabling them to learn the inherent structure and distribution of data. This capability makes generative tasks an ideal proxy task for multimodal foundational models, where consistency and diversity between different modalities are paramount. Through generative tasks, models can capture similarities and differences between modalities, offering deeper insights in medical image analysis. Additionally, generative tasks are particularly effective in data-scarce scenarios, leveraging self-supervised learning to train models on large amounts of unlabeled data, thereby reducing the dependency on extensively annotated datasets, which is critically important in medical image analysis.

Masked Image Modeling (MIM) is a rapidly evolving self-supervised learning technique in computer vision that trains models to reconstruct masked portions of images, helping them learn rich visual representations. Two widely used MIM methods are Masked Autoencoders (MAE)\cite{he2022masked} and Simple Masked Image Modeling (SimMIM)\cite{xie2022simmim}. MAE works by masking a large part of the input image, allowing the encoder to process only the visible parts, and the decoder reconstructs the full image. This design reduces computational overhead while excelling in learning global visual features. SimMIM, on the other hand, uses a more straightforward approach by randomly masking image patches and directly predicting the missing pixels, optimizing training efficiency. Both methods share the advantage of enabling powerful self-supervised pretraining without the need for explicit supervision. After training on large-scale datasets, MAE and SimMIM demonstrate strong transferability across various downstream tasks, including classification and segmentation. This makes MIM an effective proxy task for foundational models in multimodal medical image analysis, particularly in data-scarce environments, where it leverages unlabeled data for efficient training and reduces the dependence on annotated datasets.

Traditional MIM methods are often applied to natural images, but due to the sparse and concentrated nature of pathological features in medical images, random masking strategies can be less effective in capturing critical regions. 

autoSMIM\cite{wang2023autosmim} is a superpixel-based automatic masking image modeling method designed to address the limitations of traditional MIM methods in handling crucial anatomical information in medical images. Traditional random masking strategies often miss key pathological regions, whereas autoSMIM introduces a more targeted approach by leveraging superpixel masking strategies combined with Bayesian optimization to form a more precise self-supervised learning framework. The core of autoSMIM involves using the SLIC algorithm to generate superpixels, which are then masked, and the model undergoes proxy tasks (such as jigsaw, rotation, or colorization) to evaluate its ability to reconstruct critical information. Bayesian optimization strategy enables the model to automatically determine the optimal superpixel generation and masking strategy during training, thereby minimizing unnecessary information loss. The goal of optimization is to minimize the loss of task evaluation or maximize the robustness to loss of information, allowing the model to learn valuable semantic and structural features. Experimental results show that autoSMIM achieves impressive performance in multiple medical imaging datasets, including the ISIC 2016, 2017, and 2018 datasets, with Jaccard indices of 0.8705, 0.7787, and 0.8396, respectively, significantly outperforming other state-of-the-art methods. However,  the processes of generating superpixels, masking strategies, and Bayesian optimization contribute to the complexity. Particularly, the choice of compactness parameters in superpixel generation significantly impacts the final segmentation result, requiring careful tuning across different datasets or tasks.

AnatoMask\cite{li2024anatomask} introduces a reconstruction-guided dynamic masking strategy that identifies and masks crucial anatomical areas, thereby enhancing pretraining efficiency. The core innovation of AnatoMask lies in its use of reconstruction loss to dynamically generate mask regions, thereby ensuring that the model focuses on more challenging and discriminative areas. The self-distillation mechanism of AnatoMask ensures that the model continuously adjusts task difficulty during training, facilitating gradual improvement in anatomical understanding throughout the learning process. This method effectively addresses the limitations of random masking strategies in medical imaging. The results of datasets such as TotalSegmentator, FLARE22, and AMOS22 demonstrate the superiority of AnatoMask. Compared to other state-of-the-art methods, AnatoMask demonstrates notable improvements in segmentation performance and training efficiency, particularly in handling complex anatomical regions and ensuring cross-dataset generalization.

Grid Mask Image Modeling (GMIM)\cite{wang2023masked} guides the network to learn correlations between organs and tissues by reconstructing the original image from partial observations. GMIM’s architecture consists of an online branch and a target branch, both sharing an encoder and utilizing adaptive masking strategies to process input images. The online branch uses a convolutional decoder to reconstruct the masked images, while two nonlinear projectors support contrastive learning to reduce representation redundancy. GMIM’s adaptive masking strategy enhances the learning of boundary regions or small semantic changes, especially improving performance in tasks involving small organs or tumors. GMIM optimizes the model by combining reconstruction and contrastive losses. Reconstruction loss is calculated by rebuilding the masked image blocks, while contrast loss minimizes the cosine similarity of adjacent 3D volumes, thus improving the consistency of feature representation. Evaluated on BraTS 2021 (brain tumor segmentation) and Amos 2022 (abdominal multiorgan segmentation), GMIM outperforms other self-supervised methods, particularly in segmentation tasks involving small organs, significantly improving performance.

Uni4Eye++\cite{cai2024uni4eye++} integrates a MIM task utilizing Vision Transformers and introduces an entropy-guided masking strategy, along with a dynamic head generation module, to enhance the efficiency of multimodal pretraining. The entropy-guided masking strategy selects patches with high information content to mask based on entropy values, allowing the model to prioritize learning from the most informative regions of the image. Furthermore, the dynamic head generation module dynamically generates reconstruction heads according to the input modality, thus mitigating the catastrophic forgetting problem often encountered in multimodal training. Evaluations in multiple downstream tasks demonstrate that Uni4Eye++ outperforms existing self-supervised learning methods, particularly in the management of data in different modalities. The strengths of Uni4Eye++ lie in its innovative masking strategy, which directs the model to focus on more informative regions of the image, thereby addressing the limitations of random masking. Moreover, dynamic generation of reconstruction heads enables the model to maintain consistent performance across diverse modalities, thereby mitigating conflicts among modalities. Although the framework demonstrates excellent performance on large-scale datasets, an imbalance between 2D images and 3D volumetric data in the pretraining dataset may restrict performance in certain tasks. Future research could address these imbalances and further enhance the framework's efficiency and generalization capabilities.

MedIM\cite{xie2023medim} introduces two novel masking strategies: Knowledge-Driven Masking (KDM) and Sentence-Driven Masking (SDM) to enhance medical image representation. KDM utilizes Medical Subject Headings (MeSH) to identify key regions of interest in the images corresponding to radiological reports, allowing the model to focus on critical anatomical information and effectively learn discriminative features. SDM, on the other hand, generates attention maps by analyzing sentences in reports and masking the image regions relevant to specific sentences. This combination of KDM and SDM allows MedIM to reconstruct key regions during training, resulting in significant improvements in downstream tasks such as multi-label classification and segmentation, especially when labeled data are scarce. Experimental results demonstrate that, compared to random masking, KDM and SDM provide stronger semantic and feature representations. However, the performance of MedIM is highly dependent on the quality of radiological reports. Experiments have shown that noise or errors in these reports can significantly degrade the performance of the model, particularly when generating attention maps related to image regions. Future work will focus on optimizing natural language processing techniques to filter noisy reports and integrating expert feedback to improve the model's robustness.

Masked Relation Modeling (MRM)\cite{yang2023mrm} is an innovative approach for medical image pretraining that focuses on capturing the relationships between self- and cross-modality features rather than directly masking input data. This strategy allows the model to retain essential disease-related information while learning global dependencies. The two key components, self-modality relation masking and cross-modality relation masking, aim to mask token-wise feature relations within and between modalities (e.g., images and genetic data). By masking strong correlations, MRM forces the model to reconstruct from weaker connections, improving its ability to capture nuanced disease-related patterns across different modalities. To enhance the model's semantic understanding, MRM introduces a relation-matching strategy, which aligns intact and masked features both within a single modality and across multiple modalities. This encourages the model to learn consistent and semantically rich representations that can be effectively transferred to downstream tasks such as classification or segmentation. In extensive experiments on various datasets, MRM demonstrated superior performance over traditional self-supervised learning methods, particularly in tasks such as fine-grained classification and segmentation of medical images. The model achieved significant improvements in disease detection tasks, including an increase in metrics like Dice and Kappa scores, underscoring its ability to better capture medical knowledge through relation modeling.

In modern radiograph representation learning, models typically rely on self-supervision to encode invariant semantic features or on the supervision of radiology reports to incorporate expert domain knowledge. However, these two approaches often neglect the complementarity between image data and textual descriptions. To address this, the Masked Record Modeling (MReM)\cite{zhou2023advancing} framework introduces a multi-task approach that integrates both image and text data through MIM and masked language modeling. MReM pre-trains models by masking parts of both the radiograph and the associated radiology report, encouraging the model to reconstruct the missing data. This multitask learning strategy helps the model capture richer and more generalizable features. In particular, MReM enhances representation learning in scenarios with limited labeled data by leveraging both visual and textual information. For example, MReM achieves superior performance in the CheXpert dataset with only 1\% of labeled data, surpassing other models that rely on full supervision. The robustness of the model is further demonstrated through its performance in tasks such as pneumonia detection and pneumothorax segmentation, where it significantly outperforms existing self-supervised and report-supervised methods. However, MReM’s success is closely tied to the quality of the input data, particularly the accuracy and consistency of the radiology reports. If the reports contain noise or errors, the model's performance may be adversely affected.

Self-distillation Augmented Masked Autoencoders (SD-MAE)\cite{luo2022self} optimizes the random masking process by applying self-distillation loss to visible image patches. In this process, the visible patch features generated by the encoder act as the "student," while the features from the decoder serve as the "teacher." This design transfers knowledge from the deeper layers of the decoder to the shallow layers of the encoder, enhancing the model's feature learning ability. The SD-MAE framework combines the generative capability of MAE with a self-distillation process, with a particular emphasis on shallow feature learning. While the task of reconstructing masked image patches proceeds as usual, the introduction of self-distillation loss allows the model to make better use of visible patches, thus improving the overall encoding process. This additional layer of learning enhances the representational capacity of the encoder, which is crucial for downstream tasks. Experimental results demonstrate that SD-MAE significantly outperforms standard MAE in six public benchmark datasets, including PatchCamelyon and NCT-CRC-HE. In both classification accuracy and segmentation performance, SD-MAE shows marked improvements. The self-distillation mechanism also excels in domain transfer tasks, enabling the model to perform more robustly across datasets with different feature distributions.

Global-Local Masked Autoencoder (GL-MAE)\cite{zhuang2023advancing} addresses key challenges in applying MAE to 3D volumetric medical imaging. Traditional MAE approaches face limitations due to the absence of global context and unstable representation learning caused by random masking of input data. To overcome these challenges, GL-MAE integrates global and local views to create a more robust pretraining framework for 3D medical image segmentation tasks. The core innovation of GL-MAE lies in its global-local reconstruction mechanism. Global views provide low-resolution, wide-context information, which is crucial for understanding the entire clinical context of a patient, while local views offer high-resolution details for specific regions. By reconstructing both global and local masked views, GL-MAE enables the learning of rich features that encompass both broad and fine-grained information. Furthermore, the introduction of a global-guided consistency learning strategy enhances the model’s ability to maintain consistent representations across these views, which stabilizes learning and accelerates convergence. GL-MAE demonstrates its effectiveness in multiple datasets. The results show that GL-MAE significantly outperforms state-of-the-art self-supervised methods, especially when using limited annotated data, underscoring its strong generalization capabilities. Furthermore, GL-MAE showcases superior performance in cross-modality tasks, such as transferring from CT to MRI data, indicating its versatility and adaptability across different medical imaging modalities.

Traditional MAE exhibits limitations when capturing 3D complex anatomical structures. To address this challenge, the MiM (Mask in Mask)\cite{liu2023m3ae} framework introduces a multilevel reconstruction approach. By implementing a multilevel masked volume generation strategy, MiM effectively captures the hierarchical details of 3D medical images. The model performs hierarchical masking and progressively reconstructs invisible tokens, enhancing its feature representation. Cross-level alignment using contrastive learning ensures consistency across different hierarchical levels. Experiments demonstrate that MiM achieves superior performance across multiple datasets, including BTCV and MSD, excelling in both classification and segmentation tasks. MiM not only improves feature learning through multilevel reconstruction, but also ensures that the model captures critical global and local information in medical images by incorporating cross-level alignment. Compared to traditional single-level reconstruction, the MiM design enhances generalization and robustness when handling large-scale 3D volumetric images. 

To address the problem of missing modalities, the M3AE\cite{liu2023m3ae} framework combines multimodal masked autoencoders with modality completion techniques. By masking parts of the modalities and image patches, the model is forced to learn global inter-modal relationships and local intra-modal structures. Additionally, model inversion helps generate substitute modalities, enabling accurate reconstruction even when data are incomplete, thus improving overall performance. The self-distillation strategy further enhances the model's robustness by randomly dropping modalities during training and applying consistency loss to ensure semantic alignment across different modality combinations. This design enables M3AE to handle incomplete modalities while maintaining consistent segmentation performance. Additionally, M3AE leverages the Swin Transformer as its encoder and a lightweight decoder to efficiently reconstruct masked regions and extract latent 3D medical image features. Experiments on the BraTS 2018 and BraTS 2020 datasets demonstrate that M3AE excels in missing modality scenarios, particularly outperforming state-of-the-art methods under multimodality absence. The strong performance of M3AE in low-label data and cross-modality tasks highlights its practical potential in medical image analysis, demonstrating its wide applicability in real-world settings.

 In medical image analysis, the scarcity and high cost of annotated data pose significant challenges. Generative tasks, through self-supervised or unsupervised learning, can be trained on large amounts of unlabeled data, reducing the dependence on annotated data and thus making generative models advantageous in data-scarce environments. These tasks require the model not only to reconstruct the data, but also to understand the underlying structure and distribution, allowing it to learn richer and more detailed feature representations. Such representations generalize well across modalities, particularly in cross-modality generation, where the model can capture complex relationships between modalities, enhancing the performance of multimodal foundational models. Despite the potential of generative tasks to improve model performance, the generated images may not always be stable, sometimes showing blurring, artifacts, or inconsistencies, which can negatively impact downstream tasks. Furthermore, generative model training often requires high computational resources, which may limit their application in practice. Future research could focus on improving the stability and efficiency of generation. For example, optimizing model architectures and training algorithms could reduce mode collapse and output distortion, enhancing the quality of generated images. Moreover, exploring lightweight generative models, which reduce the number of trainable parameters, could maintain high performance while reducing computational costs.

\subsection{Contrastive proxy task}
Self-supervised learning (SSL) has emerged as a crucial methodology to advance multimodal foundational models, especially in the realm of 3D medical image analysis, where it has shown great potential. However, generative SSL methods tend to focus on reconstructing low-level image details, often neglecting the extraction of high-level semantic information, which can limit their effectiveness in downstream tasks. Contrastive learning offers a solution to this limitation by enabling models to learn discriminative representations by comparing similarities between positive pairs (e.g., different augmentations of the same image or related images) and contrasting them with negative pairs (e.g., samples from different images). In the context of multimodal medical image analysis, contrastive learning is particularly valuable when aligning features across different modalities, such as CT and MRI scans, or when comparing images with other types of clinical data, such as textual reports. Methods like SimCLR, which rely on contrasting different augmented views of the same image, have been adapted for multimodal contrastive learning, enabling models to learn powerful visual representations. The key strength of contrastive tasks lies in their ability to align information across modalities, facilitating the discovery of a shared latent space that captures inter-modal relationships. This leads to better feature alignment across modalities, resulting in stronger multimodal fusion and enhanced performance on downstream tasks.

SimCLR\cite{chen2020simple} is a contrastive self-supervised learning method designed to simplify the representation learning process without requiring complex proxy tasks or specialized architectures. The core idea behind SimCLR is to maximize the similarity between different augmented views of the same data sample (positive pairs) and minimize the similarity between different data samples (negative pairs). This approach allows the model to learn more robust and generalized feature representations without the need for labeled data, making it highly effective in unsupervised and semi-supervised learning tasks. A key feature of SimCLR is its use of data augmentation. The framework applies three primary augmentation techniques, random cropping and resizing, color distortion, and Gaussian blur, to generate different views of the same image. This variation enables the model to learn more generalized representations by treating these augmented views as positive pairs. Random cropping and color distortion have been identified as critical components for enhancing model performance. SimCLR also incorporates a simple yet effective multilayer perceptron (MLP) projection head that maps the encoded features to a space where contrastive learning is applied. This projection head is essential for improving the quality of learned representations.SimCLR employs the NT-Xent (Normalized Temperature-scaled Cross Entropy) contrastive loss function. This loss encourages the model to separate representations from different images, while bringing together different augmentations of the same image. A temperature parameter in the loss function modulates the contrast between positive and negative pairs, helping to refine the learned representations. In experiments on the ImageNet dataset, SimCLR achieved an impressive 76.5\% top-1 accuracy under the linear evaluation protocol, outperforming previous state-of-the-art methods by 7\%. Furthermore, when fine-tuned with only 1\% labeled data, it achieved 85.8\% top-5 accuracy, demonstrating the framework's effectiveness in semi-supervised learning. SimCLR's strength lies in its simplicity and adaptability. The framework is compatible with various neural network architectures, including convolutional networks and transformers. Its combination of data augmentation, MLP projection heads, and contrast loss allows it to excel in different tasks and datasets. 

Momentum Contrast (MoCo)\cite{he2020momentum} is a contrastive learning framework designed for unsupervised visual representation learning, which addresses the key challenges in building large, consistent dictionaries necessary for effective learning from high-dimensional visual data. Unlike conventional methods that rely on memory banks or end-to-end mechanisms for dictionary management, MoCo introduces a dynamic dictionary built as a queue, where keys are progressively updated using a momentum-based encoder. This momentum mechanism ensures that the dictionary remains consistent over time as the key encoder evolves more smoothly compared to the rapidly updated query encoder. The core idea is to match encoded queries to keys using contrastive loss, where positive pairs (the same image in different views) are pulled together in the representation space, while negative pairs (different images) are pushed apart. MoCo's framework decouples the dictionary size from the mini-batch size, allowing for larger, more representative dictionaries that span numerous mini-batches, effectively sampling the visual space. The model updates the dictionary with each mini-batch and replaces the oldest batch, ensuring that the most outdated samples, which may lack consistency with newer ones, are removed. The momentum update prevents the key encoder from changing too quickly, preserving consistency in the representations across mini-batches. This is in contrast to memory bank approaches, where encoded keys from earlier training steps may become inconsistent with current representations. Evaluated on several downstream tasks, such as object detection and segmentation, MoCo demonstrates competitive performance, often exceeding supervised learning counterparts. It achieves strong results in linear classification protocols on ImageNet and exhibits significant transferability to downstream tasks, where it outperforms supervised pretraining methods in object detection and segmentation tasks. The momentum-based dictionary and its flexible size contribute to the robustness of MoCo across different visual tasks and datasets.

VoCo\cite{wu2024voco} is a simple yet effective volume contrastive learning framework designed for 3D medical image analysis, using anatomical consistency to enhance semantic representation. The framework consists of two main branches: one for contextual position prediction and the other for regularization. Initially, the input 3D image is divided into nonoverlapping base volumes, and a randomly cropped volume is used to predict its position in the image by contrasting its similarity to these base volumes. VoCo learns contextual position information between volumes by calculating the similarity between random crops and base volumes. Additionally, VoCo introduces a regularization loss to amplify feature discrepancies between different base volumes, ensuring that features are distinctive. This process allows VoCo to encode high-level semantic information effectively, improving downstream tasks such as segmentation and classification. VoCo was evaluated in several public 3D medical image datasets, including BTCV, LiTs, and BraTS, outperforming existing SSL methods, especially in tasks requiring rich semantics. The experimental results demonstrated a significant improvement in the Dice scores, and VoCo also showed strong cross-modality performance. The context-aware volume contrast learning strategy significantly enhances the representation of 3D medical images and aids in extracting valuable semantic information across complex anatomical structures. 

In the domain of medical image analysis, position contrast learning (PCL)\cite{zeng2021positional} offers an innovative self-supervised learning framework that addresses the common challenge of false negative pairs. In traditional contrastive learning, similar anatomical structures in different medical images can lead to incorrect negative pairs that can degrade the performance of the representation. PCL leverages the positional information of slices within volumetric images to generate contrastive data pairs, reducing the incidence of false negatives and improving the quality of the representation. Each 2D slice is associated with its relative position within the 3D image, and by calculating the positional differences between slices, PCL effectively distinguishes positive from negative pairs, making better use of spatial features. The PCL framework employs a 2D U-Net architecture along with a shallow MLP projection head to embed features. During pretraining, it optimizes the model using contrastive loss. Once pretraining is complete, the projection head is discarded and the encoder initializes a segmentation network that is fine-tuned on a small labeled dataset. PCL excels at generating high-quality feature representations, particularly effective in scenarios with limited labeled data. Using positional information to construct contrastive pairs, PCL effectively captures anatomical differences in medical images while minimizing false negative interference. Experimental results on four public medical image datasets show that PCL outperforms existing contrastive learning methods in semi-supervised and transfer learning tasks, especially with limited labeled data. This underscores its wide applicability and advantages in medical image segmentation. Future research could investigate the extension of PCL to 3D contrastive learning tasks and its integration with other self-supervised learning methods to enhance multimodal representation capabilities.

The COMET\cite{wang2024contrast} framework is designed to address the limitations of existing self-supervised contrastive learning approaches in medical time series analysis. Current methods often focus on individual data levels, failing to exploit the inherent hierarchical structure in medical time series data. COMET leverages a hierarchical contrastive approach, capturing data consistency across multiple levels, including observation, sample, trial, and patient levels. By doing so, the framework systematically captures fine-grained to coarse-grained information, significantly enhancing the learning capacity of representation in the context of medical time series. At the observation level, COMET captures consistency by contrasting original and augmented views of individual observations, assuming that minor perturbations should preserve the core information. At the sample level, this contrastive approach is extended by contrasting whole samples against their augmented counterparts. Moving to the trial level, COMET assumes that the samples within the same trial are more similar than those from different trials. Finally, at the patient level, COMET captures the overall patient-specific consistency by contrasting samples between different trials of the same patient. COMET outperforms existing state-of-the-art methods such as TS2Vec and CLOCS in multiple data sets, including EEG-based Alzheimer's detection, ECG-based myocardial infarction detection, and EEG-based Parkinson's disease diagnosis. The results demonstrate its effectiveness, particularly when labeled data are scarce, as COMET shows superior performance even with only 1\% or 10\% labeled data. 

In \cite{azizi2021big}, the authors propose a novel multi-instance contrast learning (MICLe) method that significantly improves medical image classification performance by leveraging both natural and domain-specific unlabeled medical images for pretraining. MICLe builds upon the SimCLR framework by utilizing self-supervised learning to maximize the similarity between different augmentations of the same image. However, MICLe extends SimCLR by creating positive pairs from various images of the same medical case (e.g., images from different views of the same pathology), thereby enhancing the model's ability to learn representations that are robust to changes in viewpoint, lighting, and other factors. This approach proves particularly effective in medical contexts where multiple images of the same condition are frequently available (e.g., different angles in X-rays or dermatological photographs). The MICLe method constructs more informative positive pairs for contrastive learning, which makes it particularly well suited for medical datasets. Pretraining with MICLe on dermatology and chest X-ray datasets yielded a 6.7\% improvement in top-1 accuracy for dermatology classification and a 1.1\% improvement in AUC for chest X-ray classification, outperforming models pre-trained on supervised ImageNet data. The combination of self-supervised learning on both natural and medical images not only bridges the domain gap but also ensures that the model acquires domain-specific features crucial for medical tasks. The results demonstrate that self-supervised pretraining on medical images represents a viable alternative to traditional supervised pretraining, particularly in scenarios where labeled data are scarce. 

Contrastive learning, a pivotal self-supervised approach, significantly improves model generalization by learning robust feature representations by comparing positive and negative sample pairs. This method is particularly beneficial in multimodal medical image analysis, where it captures cross-modality shared features, enhancing the performance of fusion tasks. Additionally, it reduces reliance on large labeled datasets by training on unlabeled data, which is critical in medical fields with limited annotations. However, challenges remain, such as the complexity of negative sample selection and high computational demands, as improper sample selection can hinder model performance. Future research should focus on optimizing negative sample selection and computational efficiency to further strengthen contrastive learning’s application in medical imaging tasks.

\subsection{Hybrid proxy task}
Hybrid tasks, when used to build a foundational model, present considerable advantages over proxies with a single task, especially in the management of multimodal data. In contrast to single tasks, which may be limited in scope, composite tasks concurrently leverage multiple learning objectives, enabling models to capture more diverse and comprehensive feature representations. This multitask learning paradigm not only enhances the model's capability to extract multidimensional features but also promotes superior generalization by facilitating cross-task interactions, thereby improving performance on complex data distributions and diverse task requirements. Multimodal data typically comprises various types of inputs. Using composite tasks as proxies, models can learn shared representations across modalities, thereby facilitating the integration of information from diverse sources. For example, the combination of discriminative and restorative learning tasks improves both global information discrimination and preservation of local details. This multitask learning mechanism enables models to integrate data from multiple modalities more effectively, ultimately enhancing their performance in multimodal fusion tasks while bolstering robustness and adaptability in downstream applications.

The DIRA\cite{haghighi2024self} framework presents an innovative self-supervised learning approach by combining three distinct strategies: discriminative, restorative, and adversarial learning. This combination aims to overcome the limitations of traditional self-supervised learning methods that typically focus on one or two strategies in isolation. By uniting these approaches, DIRA facilitates a collaborative learning mechanism that enhances the model's performance across various downstream tasks, including medical image classification and segmentation. Discriminative learning helps DIRA to learn global image representations by maximizing the similarity between similar (pseudo) class instances. Restorative learning enables the model to capture fine-grained details by reconstructing distorted images, while adversarial learning increases the model's capacity to retain intricate image details by employing a generative adversarial network\cite{goodfellow2020generative} to challenge the restoration process. Extensive experiments on datasets such as ChestX-ray14 and CheXpert show that DIRA significantly outperforms traditional supervised and unsupervised models, particularly in low-data regimes where labeled data are scarce. The model demonstrates a strong ability to generalize across different organs, diseases, and modalities, making it highly effective for medical imaging tasks. Moreover, DIRA's joint optimization framework allows for a fine balance between global and local feature learning, further improving its robustness in medical image analysis. However, the integration of multiple learning strategies also increases computational complexity, requiring careful tuning of loss function weights to ensure optimal performance. However, DIRA sets a new standard for self-supervised learning in medical imaging by combining discriminative, restorative, and adversarial learning into a cohesive framework.

DAE\cite{da} is designed to address the unique challenges presented by 3D medical imaging, which requires the capture of fine-grained anatomical details such as tissues, lesions, and organs. Unlike natural images, medical images, especially 3D scans, such as CT and MRI, contain more intricate and localized features. The DAE framework introduces a novel approach to pretraining by focusing on local masking and low-level perturbations, such as downsampling and noise addition, to improve the model's ability to learn low-level features. Traditional masking techniques, such as Masked Autoencoders, rely on global masking strategies that mask entire channel dimensions. This often results in the loss of critical local information. DAE, on the other hand, employs a local masking strategy that selectively masks parts of channel embeddings, preserving enough local information to accurately reconstruct key medical structures, particularly fine anatomical features. Additionally, the framework incorporates classic tasks such as denoising and down-sampling. By adding Gaussian noise to the original images, the model is trained to restore these noisy images, thus improving its capacity to capture subtle details. Furthermore, the down-sampling task generates low-resolution images and requires the model to reconstruct high-resolution versions, a particularly useful skill for medical imaging where long scan times can lead to low-resolution problems. In terms of multimodal learning, DAE introduces cross-modal contrast learning (CMCL) to maintain consistent representations across different modalities, such as CT and MRI. By contrasting feature representations across modalities, the model is encouraged to cluster similar features within the same modality while separating features across different modalities. This improves the model’s ability to differentiate between data from different imaging techniques. The effectiveness of DAE has been validated through self-supervised pretraining in several public 3D medical imaging datasets, such as BTCV and FeTA, followed by fine-tuning of downstream tasks such as segmentation. DAE offers a powerful solution for pretraining in medical imaging, with particular strengths in fine-grained feature capture and multimodal adaptation.

Preservational Contrastive Representation Learning (PCRL)\cite{zhou2021preservational} aims to improve the self-supervised learning process in medical image analysis by integrating contrastive learning with context reconstruction to preserve more comprehensive and meaningful information in the learned representations. PCRL addresses the limitations of traditional contrastive learning methods, which primarily focus on contrasting image pairs without explicitly preserving detailed image information. By introducing additional context reconstruction, PCRL captures richer details, making it highly suitable for medical imaging tasks where high precision and detailed feature representation are essential. The PCRL framework incorporates three different encoders (an ordinary encoder, a momentum encoder, and a hybrid encoder) along with a shared decoder. This architecture enables the model to perform multiple reconstructions of diverse contexts, thereby enriching the information captured by the learned representations. The TransAtt (Transformation-conditioned Attention) module dynamically adjusts the reconstruction tasks based on transformation indicators, allowing the model to reconstruct diverse image contexts from the same input image. Additionally, the cross-model mixing module generates a hybrid encoder by combining feature maps from the ordinary and momentum encoders, further enhancing the robustness of the learned representations. Through extensive evaluations of several datasets, including LUNA and LiTS, PCRL demonstrated significant improvements in both classification and segmentation tasks. It outperformed traditional contrastive learning approaches and even supervised methods, particularly when trained on limited labeled data. The introduction of multi-context reconstruction proved to be highly effective in improving the model's performance across various downstream medical imaging tasks.

PCRLv2\cite{zhou2023unified} builds on the original PCRL framework by introducing a more unified and comprehensive approach to preserve pixel-level, semantic, and scale information in self-supervised medical image pretraining. It incorporates multiscale pixel restoration and feature comparison to ensure that the model can capture both local details and high-level semantics across various scales. Unlike typical U-Net architectures, PCRLv2 leverages a non-skip U-Net (nsUNet) to construct a feature pyramid without relying on skip connections. This design allows deeper integration of pixel-level and semantic features, enhancing the model's representational power without shortcut solutions. Additionally, PCRLv2 introduces a "sub-crop" strategy to better link local and global views in 3D medical images, improving the consistency and effectiveness of training. Extensive experiments on datasets demonstrate the superiority of PCRLv2 over other self-supervised learning frameworks, especially in semi-supervised and transfer learning scenarios. By incorporating pixel-level restoration, multiscale representations, and feature comparison, PCRLv2 achieves state-of-the-art performance in both classification and segmentation tasks, with notable improvements in small-scale labeled data settings.

The UniMiSS\cite{xie2022unimiss} framework (Universal Medical Self-Supervised Learning through Breaking Dimensionality Barrier) addresses a critical challenge in medical image analysis: the scarcity of 3D data due to the high cost of collection and privacy concerns. UniMiSS proposes a novel method to break the dimensionality barrier by leveraging abundant 2D medical images, such as X-rays, to enhance the SSL process for 3D medical images like CT scans. UniMiSS utilizes a MiT backbone (Pyramid U-like Medical Transformer) with a Switchable Patch Embedding (SPE) module, which can handle both 2D and 3D image inputs by automatically switching between 2D and 3D patch embeddings. This enables UniMiSS to process a mixed dataset of 2D and 3D images in a self-distillation training process. The self-distillation strategy employs a student-teacher paradigm, where the teacher network guides the student by enforcing view consistency across different perspectives of the input data. One of the key innovations in UniMiSS is the volume-slice consistency constraint, which ensures that 3D volumes and their corresponding 2D slices are treated consistently, thereby improving the quality of 3D feature representation. By incorporating 2D images into the SSL training for 3D tasks, UniMiSS achieves significant improvements in both 3D segmentation and classification tasks, outperforming competing SSL methods and even models pre-trained on ImageNet. This framework was evaluated on multiple datasets, including more than 5000 3D CT scans and 100,000 2D X-ray images, and demonstrated superior performance in a range of 3D and 2D medical image tasks. In downstream evaluations, UniMiSS excelled in segmentation tasks on various datasets, with clear performance advantages over existing models.

The SwinMM\cite{wang2023swinmm} framework is a novel approach designed to enhance 3D medical image segmentation through the integration of self-supervised multiview learning. By combining multiple perspectives from 3D medical data, such as axial, sagittal, and coronal views, SwinMM processes the diverse information in an optimized pipeline. This is achieved through a masked multiview encoder in the pretraining phase and a cross-view decoder in the fine-tuning phase. The encoder learns high-level representations using four distinct proxy tasks: image reconstruction, rotation prediction, contrastive learning, and mutual learning. The mutual learning task, a novel contribution, ensures that the predictions of different views remain consistent, thus maximizing the multiview information extraction. During the fine-tuning phase, the cross-view decoder utilizes cross-attention mechanisms to aggregate the multiview data into a coherent and accurate prediction, improving segmentation performance. A multiview consistency loss further enhances the robustness of the segmentation by enforcing consistency between different view outputs. This multiview ensemble approach effectively reduces prediction uncertainties and enhances the model's ability to handle difficult segmentation tasks, particularly with smaller or less distinct targets. SwinMM has shown significant improvements over previous state-of-the-art models, such as Swin UNETR, especially in terms of segmentation accuracy and data efficiency. It achieved superior performance on multiple benchmark datasets with higher Dice scores and reduced Hausdorff distances, even when trained with fewer labeled data. The framework’s reliance on multiview information and its adaptability to semi-supervised learning make it an innovative and effective solution for 3D medical image analysis.

In conclusion, the integration of various advanced self-supervised learning frameworks, such as SwinMM, UniMiSS, and PCRL, marks a significant advancement in multimodal medical image analysis. These frameworks, founded on multiview, multiscale, and cross-modal learning, have effectively tackled the inherent challenges of medical imaging, including data scarcity, multimodality, and fine-grained feature extraction. By integrating innovative proxy tasks, contrastive learning, and cross-view consistency mechanisms, these models not only enhance feature representation but also improve the overall robustness and generalization capabilities of the foundational models. The diverse applications, ranging from segmentation to classification, showcase the versatility and efficacy of these unified approaches. Future research should focus on refining these methods to reduce computational complexity, ensuring scalability across larger datasets, and exploring their applicability in real-world clinical settings. This path forward highlights a promising avenue for more efficient, accurate, and generalizable medical image analysis models that could significantly improve diagnostic and therapeutic workflows.

\section{Medical Multimodal Vision-Language Foundation models}

\subsection{Vision-Language Representation}

{\color{black}Advancements in machine learning, particularly vision-language representation, are revolutionizing clinical practice by enhancing the interpretation of medical images through natural language understanding. This integration is especially impactful in cardiovascular medicine, where precise analysis of imaging data is crucial for diagnosis and treatment planning. Contrastive Language-Image Pretraining (CLIP), a cutting-edge machine learning approach developed by OpenAI\footnote{https://openai.com/}, offers a novel approach to align medical images with textual descriptions.} The core idea behind CLIP is to create a model that can learn powerful generalized representations by aligning images and text in a shared feature space, enabling the model to perform a wide range of tasks without task-specific training.

 \begin{figure}[ht]
  	\begin{center}
  		\includegraphics[width=\linewidth]{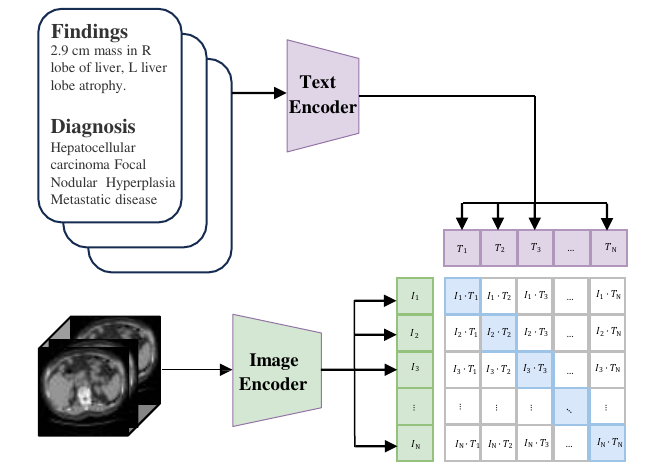}
  	\end{center}
  	\vspace{-1em}
  	\caption{ Overall architecture of CLIP.}
  	\label{clip}
  \end{figure}

\subsubsection{Architecture of CLIP}
The Contrastive Language-Image Pretraining (CLIP) architecture is designed to effectively align images and text in a shared feature space, enabling the model to understand and relate these two modalities. As shown in Fig. \ref{clip} The architecture is composed of two main components: the Text Encoder and the Image Encoder. Each component is responsible for processing one type of input, text, or image and transforming them into a high-dimensional vector representation that can be compared within the same space.

\subsubsection{Contrastive pretraining}
Contrastive pretraining in CLIP is a core component of its learning strategy, allowing the model to learn visual representations directly from language supervision. In practical terms, given a batch of $N$ (image, text) pairs, the model considers all
$N^2$ possible pairings of images and texts, but only $N$ of these pairings are correct. The contrastive objective pushes the correct pairs closer together in the embedding space, while pushing the incorrect pairs further apart.
For a batch of size $N$, where $\mathbf{v}_i$ is the image embedding and $\mathbf{t}_i$ is the text embedding corresponding to the $i$-th image-text pair:

\begin{small} 
 	\begin{equation}
 		\begin{aligned}
 			\mathcal{L}=\frac{1}{2 N} \sum_{i=1}^N\Bigg[-\log \frac{\exp \left({sim}\left(\mathbf{v}_i, \mathbf{t}_i\right) / \tau\right)}{\sum_{j=1}^N \exp \left({sim}\left(\mathbf{v}_i, \mathbf{t}_j\right) / \tau\right)}- \\ \log \frac{\exp \left({sim}\left(\mathbf{t}_i, \mathbf{v}_i\right) / \tau\right)}{\sum_{j=1}^N \exp \left({sim}\left(\mathbf{t}_i, \mathbf{v}_j\right) / \tau\right)}\Bigg]
 		\end{aligned}
 		\label{eq_s}
 	\end{equation}
 \end{small}

Where ${sim}\left(\mathbf{v}_i, \mathbf{t}_i\right)$ is the measure of similarity (often cosine similarity) between the image embedding $\mathbf{v}_i$ and the text embedding $\mathbf{t}_i$, and $\tau$ is a temperature parameter that controls the sharpness of the distribution.

This loss is averaged over all the positive pairs in the batch, considering both directions (image-to-text and text-to-image). The first term in the summation corresponds to matching the image with its corresponding text, and the second term corresponds to matching the text with its corresponding image.


 \begin{figure}[ht]
  	\begin{center}
  		\includegraphics[width=1.0\linewidth]{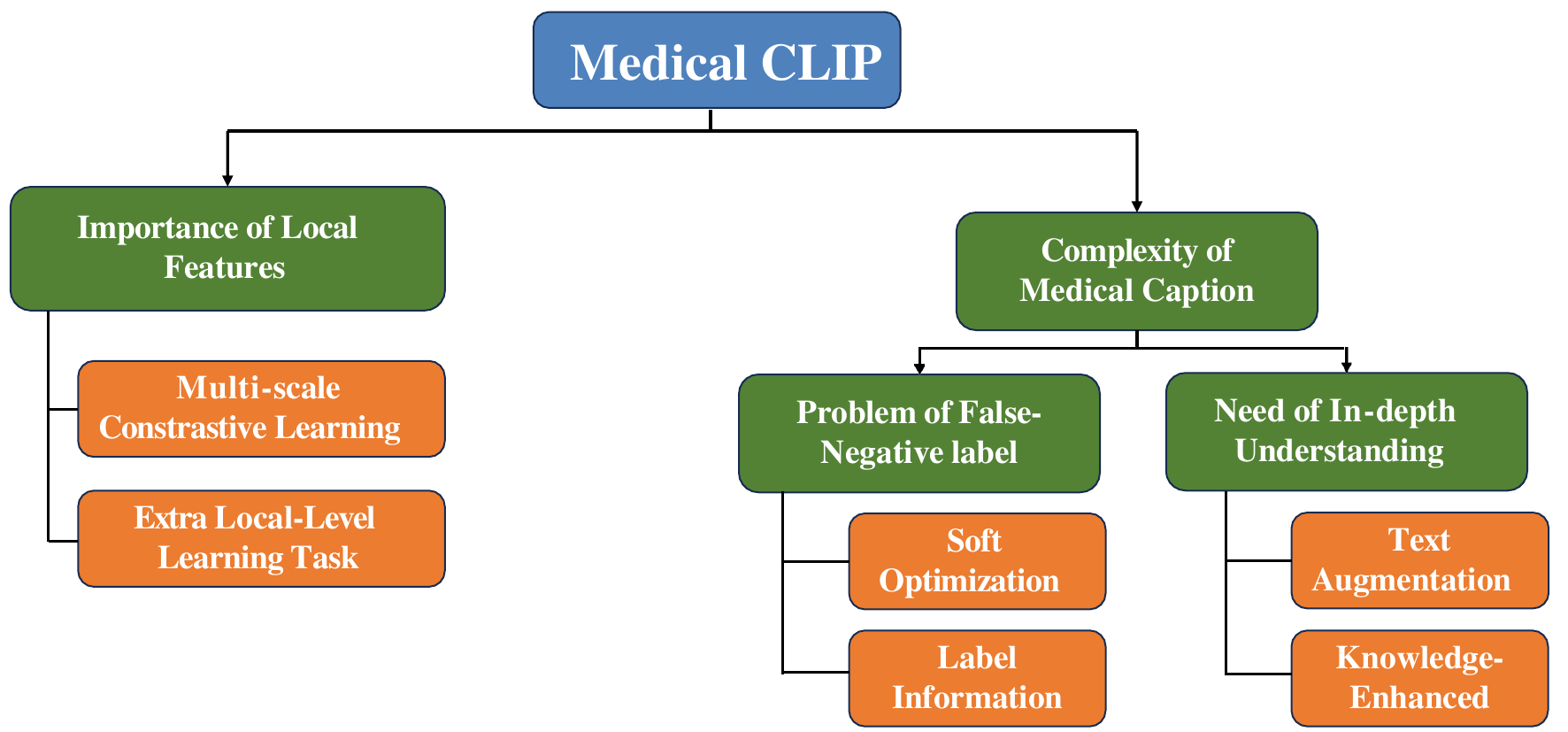}
  	\end{center}
  	\vspace{-1em}
  	\caption{ Taxonomy of works of CLIP within medical image domain.}
  	\label{content}
  \end{figure}

\subsection{Medical CLIP}

{\color{black}Although the CLIP model has shown remarkable success in natural image processing, directly using CLIP on medical images presents significant challenges due to fundamental differences in image characteristics and textual descriptions between the two domains. In clinical practice, the precise interpretation of medical images is crucial for accurate diagnosis and effective treatment planning, particularly when detecting subtle abnormalities that may indicate serious conditions. For example, in cardiovascular imaging, the identification of minute lesions or variations in vascular structures is vital for early intervention and patient outcomes, which, however, may not be detected by CLIP trained with natural images because training data for natural images do not characterize features of such precision. }

Thus, directly using CLIP within the medical domain may cause a performance not as great as within the natural image domain for challenges both from image and text data. The structure of this section is shown in Fig. \ref{content}. 

One key distinction between medical and natural image representation is the importance of local-level features. Unlike masked image modeling, CLIP is trained with semantic-level supervision and exhibits limitations in extracting local-level features. The lesion regions typically occupy a small region of the entire medical image, while the medical images themselves are often structurally standardized and share substantial structural similarities at the general level. In addition, the captions of medical images typically focus on specific small areas, such as lesions, vessels, or organs. Medical reports, a common form of medical image captioning, are generally divided into sections, each focusing on individual structures occupying relatively small regions. In contrast, captions for natural images tend to offer general-level descriptions. Due to the model's structure, image and text features in CLIP are aligned at a general level, leading to a focus on general-level features in natural images while limiting its capacity to capture local-level details, which are crucial in medical image representation.

Unlike the natural image domain, the caption of medical images is often complex and highly specialized. This has resulted in two main challenges for medical CLIP: \textbf{problem of false-negative label} and \textbf{need of in-depth understanding}. (1) For pretraining in the natural image domain, unpaired images and text are often treated as negative pairs. However, in medical captions such as medical reports, a small number of diseases and findings account for the majority of cases\cite{FNL_Bustos}, and thus there is often significant semantic overlap between each other. Thus, directly considering unpaired medical images and text as negative pairs can result in poor performance during pretraining. (2) The relationships inside natural image captions are usually simple, like "Pepper the Aussie pup", while the relationship between each organ and structure described in the medical caption is intricate. Directly using medical image captions for pretraining can cause a lack of in-depth understanding and result in a degradation of the pretraining model effect.

The challenges mentioned above highlight the deficiencies of directly using CLIP for pretraining within the medical domain and emphasize the necessity of improving the pretraining process based on the characteristics of the medical image domain.

\begin{figure*}[!t]
  	\begin{center}
  		\includegraphics[width=0.9\linewidth]{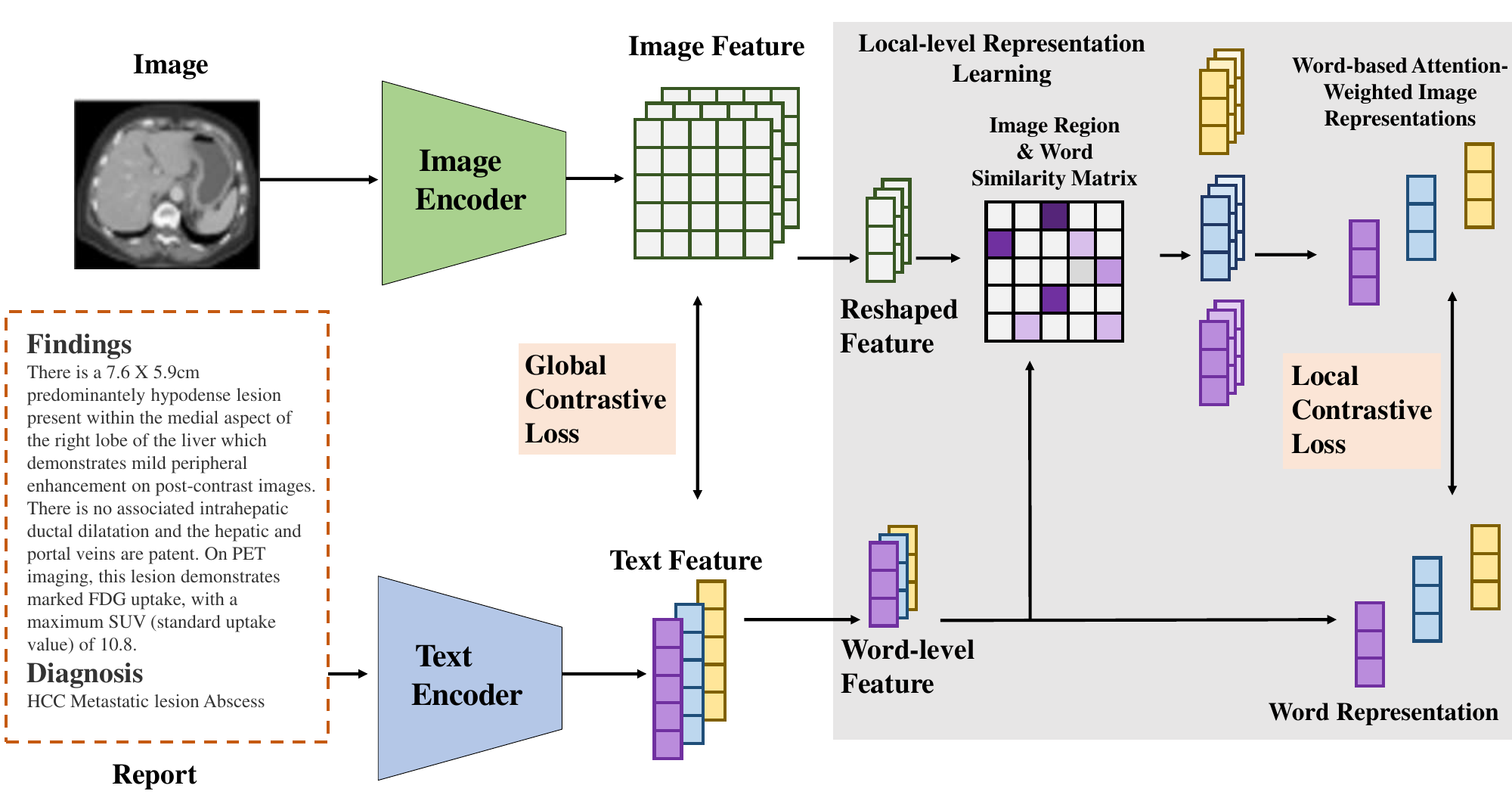}
  	\end{center}
  	\vspace{-1em}
  	\caption{ Brief structure of the mechanism which enhances the representation of local features, using the attention-based mechanism.}
  	\label{gloria}
  \end{figure*}

\subsubsection{Importance of Local Features}
Some studies have attempted to train CLIP in the medical image domain\cite{clip_global_1,clip_global_2}, following the CLIP structure raised by \cite{CLIP}, which aligns image and text features at a global level. However, when dealing with more detailed downstream tasks, such as segmentation, it fails to achieve the desired performance. Current research has mainly tried to address this issue in two ways: (1) implementing contrastive learning on both global and local features; (2) incorporating additional learning tasks to emphasize local features.

\paragraph{Multi-scale Contrastive Learning}

To enhance the representation of local features, \cite{gloria} uses an attention-based mechanism, called GLoRIA, to learn both global and local representations by contrasting image subregions with the corresponding words in the paired report. Specifically, as shown in Fig. \ref{gloria}, the local and global features corresponding to the image and the report are first extracted by the image encoder and the text encoder, respectively. Then, global-level representation is learned via contrastive loss between the global features of the text and the image. At the same time, to learn local-level representation, attention-weighted image representations are calculated based on a similarity matrix between image subregion features and word-level features. And local-level representation is learned from contrastive loss between attention-weighted image representations and corresponding word representations. 

Unlike \cite{gloria}, LoVT\cite{LoVT} uses image subregions and corresponding sentences, not words, to learn the representation at the local level. Specifically, similar to GLoRIA, based on query-key-value attention, LoVT has calculated a cross-modal representation $Z_{m}^{I \rightarrow R}$ for each sentence $m$ by letting the sentence feature attend to all image region representations and aligned local sentence representation with $Z_{m}^{I \rightarrow R}$. Similarly, for each image $k$, LoVT has also calculated $Z_{k}^{R \rightarrow I}$ and aligned the local image representation with $Z_{k}^{R \rightarrow I}$. The model learns local-level representation through the above two losses. Based on LoVT, \cite{local_pm} further breaks down global and local contrastive losses into alignment components (which pull similar representations closer) and distribution priors (which push dissimilar representations apart) and suggests that the alignment effects of local and global losses are similar, and thus, the local alignment component can be simplified. Following this, this approach has removed the traditional local alignment components and replaced them with a novel local uniformity regularizer (Gaussian uniformity loss).

\cite{localmi} has tried to improve local-level representations by maximizing mutual information (MI) between local features in images and the corresponding descriptions in texts. \cite{local_seibold} uses local-level features extracted from the images to learn local-level representations by aligning them with the embedded features of each sentence in the report separately. \cite{local_tier} introduces an entropy-based regularization scheme which aims to ensure that each text token correlates strongly with only a few image patches and each image patch correlates strongly with only a few text tokens, penalizing the entropy of the similarity scores between text token and image patch embeddings.

\paragraph{Extra Local-Level Learning Task}

During the pre-training process, \cite{PRIOR} has introduced a cross-modality conditional reconstruction (CCR) module to further enhance local-level and fine-grained information learning, in addition to using the alignment of images and the corresponding reports at both global and local levels for learning. This module has reconstructed masked images and generated report sentence prototypes based on cross-modality representations, ensuring that both visual and linguistic features are effectively learned at the local level.

\cite{zhou2023advancing} does not use contrastive loss but instead has proposed a unified framework called Masked Record Modeling that combines two tasks, allowing for the reconstruction of masked image patches and masked report tokens. For global-level representation, by combining the global feature with the report tokens, MReM integrates image-level information into the report restoration process. This facilitates the incorporation of domain-specific knowledge into the representation and reinforces the global understanding of the radiograph. For local-level representation, MReM uses a high masking ratio to hide random patches of the radiograph image, which are then reconstructed by the model. The task of reconstructing these masked patches requires the model to focus on fine-grained details and local patterns within the image, which encourages the model to learn local-level features. Additionally, MReM enhances local-level representation learning by reconstructing high-resolution image patches from low-resolution input, forcing the model to encode detailed and local-level information.


\subsubsection{Complexity of Medical Caption}
Caption of medical images is often complex and highly specialized, resulting in two main challenges for medical CLIP: the problem of false negative labels and the need for a detailed understanding.

\paragraph{Problem of False-Negative Label}
In medical reports, due to the clear purpose of diagnostics, the diagnosis of diseases and the corresponding findings are often written, and the majority cases are of a small number of diseases and findings\cite{FNL_Bustos}. Thus, there is often significant semantic overlap between each other, and it is not supposed to directly consider unpaired medical images and text to be a negative pair. MedCLIP\cite{medclip} solves this by decoupling images and texts and using medical knowledge to build a semantic similarity matrix. This matrix guides the training process, ensuring that semantically similar images and texts are not incorrectly classified as negatives. The soft semantic matching loss replaces the traditional InfoNCE loss, allowing MedCLIP to capture the subtle but crucial medical meanings more effectively, thereby reducing the impact of false negatives during training. \cite{FNL_LiuBo} addresses the issue of false-negative labels by introducing report pairs with high semantic similarities as additional positive samples while treating those with moderate similarities as neutral samples, which are excluded from the optimization process. \cite{umcl} has leveraged specific representational information from the image-label data and combined it with the generic representation information from image-text pairs. The key to this approach is the inclusion of disease labels within the image data, which helps the model better understand the semantic consistency between images and their corresponding prompts.

\paragraph{Need of In-depth Understanding}
\cite{biovil} incorporates a text enhancement technique in which sentences within sections of the radiology report are randomly shuffled during pretraining, enhancing the robustness and generalization of the language model, particularly in the context of radiology reports, which typically have a specific structure but may vary in sentence order. Furthermore, the study has introduced a novel pretraining task, called Radiology Section Matching (RSM), which involves training the model to match the Impression section to the Findings section of the same report, thereby improving the model's ability to comprehend the relationship between different parts of the report and enhancing its overall text comprehension.

 \begin{figure*}[!t]
  	\begin{center}
  		\includegraphics[width=\linewidth]{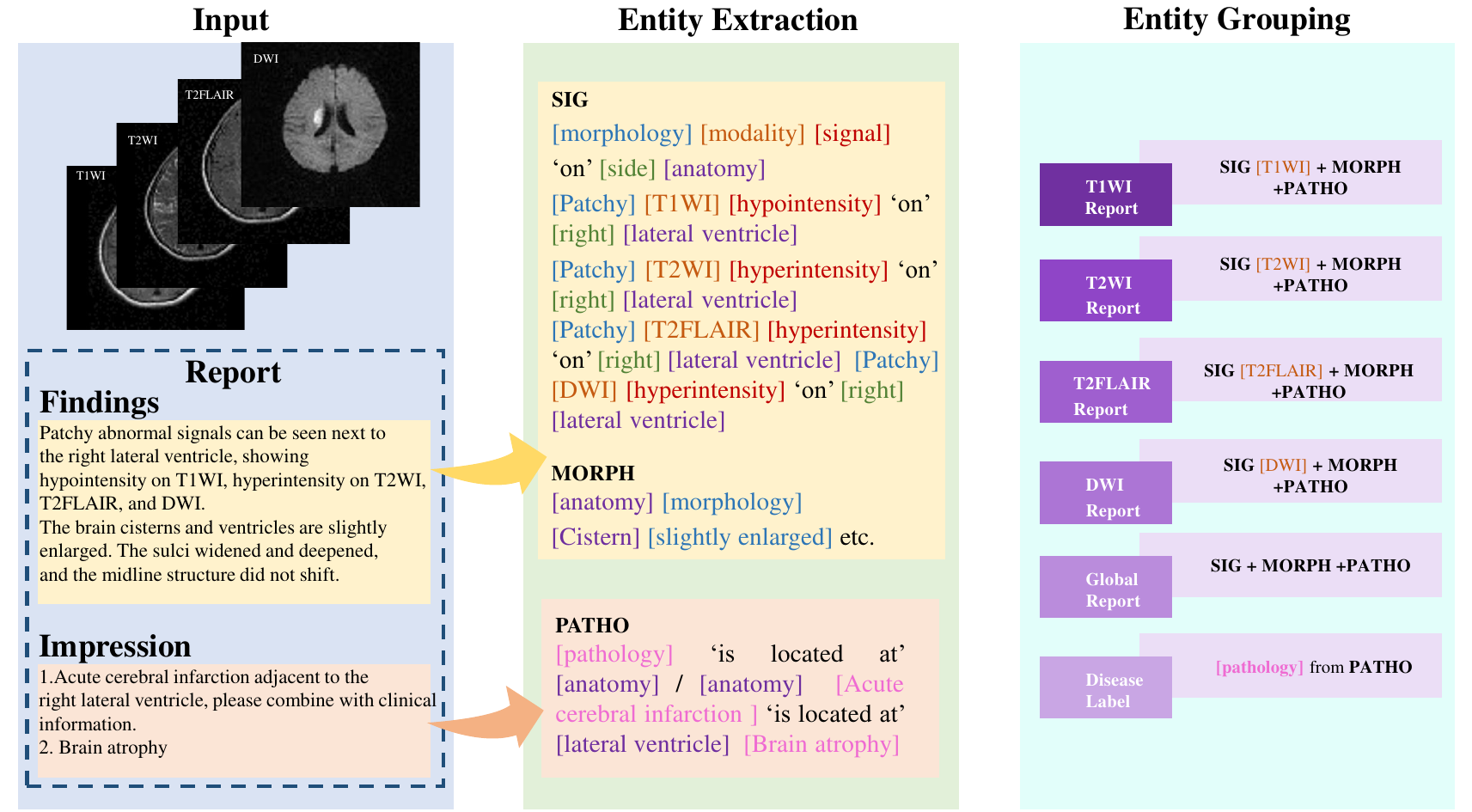}
  	\end{center}
  	\vspace{-1em}
  	\caption{Architecture of Automatic Report Decomposition process, an automatic report decomposition pipeline to extract structured modality-wise and global MRI report from the vanilla MRI report findings and report impression.}  
  	\label{ka}
  \end{figure*}

\cite{ka_1} has introduced the Automatic Report Decomposition (ARD) process, which is a crucial component of the hierarchical knowledge-enhanced pretraining framework. It is designed to transform often complex and unstructured MRI reports into a structured format that can be used effectively in the pretraining of the model. The ARD process involves two main steps: Entity Extraction and Entity Grouping. As shown in Fig. \ref{ka}, six types of entities—anatomy, side, modulality, signal, morphology, and pathology—are extracted and restructured into one of the following three predefined formats. (1) SIG (signal): This format captures information on signal intensities related to specific modalities. (2) MORPH (morphology): This format captures morphological changes. (3) PATHO (pathology): This format captures the pathology and its location. Using a priori knowledge to structure the report to bootstrap the model for more in-depth learning of the hierarchical information patterns in the report. MedKEBERT\cite{medkebert}, a specialized version of the BERT that has been pre-trained on medical text data, is introduced to process structured medical text data and generate embeddings that represent textual information in a format that can be aligned with visual features from the medical image. \cite{medklip} has used SciSpacy, a named entity recognition tool, to identify medical entities in the text and link them to the Unified Medical Language System (UMLS) knowledge base. After identifying relevant entities, this study extracted knowledge graph triples from the UMLS knowledge base and aligned the embedding of knowledge graph triple with image embedding and text embedding.

\section{Application}

MMFMs have emerged as pivotal tools in clinical practice, addressing complex medical challenges by integrating and analyzing multimodal data. By synthesizing diverse data types, MMFMs deliver holistic insights that surpass the capabilities of traditional unimodal approaches. Their transformative impact spans critical clinical tasks, such as radiology report generation, diagnosis, and treatment, enabling precision medicine and improving clinical workflows. In this section, we discuss these applications in three key domains, illustrating how MMFMs serve as powerful tools for analysis, decision making, and interaction in complex medical scenarios.

\subsection{MMFMs for Radiology Report}

The primary applications of multimodal foundational models for radiology reports lie in two tasks: radiology report generation and comprehension. For the generation task, these models can focus on abnormal details within each modality's images, while also taking into account the patient's textual medical history and clinical examination results. By combining text and images, the models can describe different regions to produce more accurate radiology reports, significantly reducing radiologists' workload\cite{CT-RATE}. Additionally, for the comprehension task, doctors can utilize multimodal models for a more targeted understanding of case details through multi-round text-based dialogues focused on specific image sections, allowing for detailed descriptions of particular image areas.\cite{llava_med,med_flamingo,pathchat}

{\color{black}
\subsubsection{MMFMs for Radiology Report Generation}
The field of Radiology Report Generation (RRG) has advanced significantly with the integration of deep learning and multimodal data processing techniques, particularly in the cardiovascular domain. Early foundational work by \cite{Vinyals} introduced a neural image caption generator that combined convolutional neural networks (CNNs) for image feature extraction with recurrent neural networks (RNNs) for sequential text generation, setting the stage for automated medical report generation. Building on this, \cite{tienet} developed TIE-Net, a text-image embedding network designed for thoracic disease classification and reporting in chest X-rays, effectively integrating visual and textual data to enhance diagnostic accuracy.

Similarly, \cite{Wehbe} provided a comprehensive review of deep learning applications in cardiovascular imaging, emphasizing their clinical relevance. They explored how deep learning models have improved tasks such as left ventricular function assessment, myocardial segmentation, and plaque characterization. Moreover, the integration of advanced NLP methods in generating cardiovascular imaging reports was highlighted as a pathway to improving efficiency and consistency in clinical workflows.

Further advancements for usage of MMFMs in RRG include \cite{liknowledge}, who proposed a knowledge-driven approach emphasizing domain-specific knowledge in report generation, and \cite{chen2020generating}, who introduced a memory-driven transformer model to enhance the coherence and clinical relevance of generated reports. \cite{liu2021exploring} expanded on these ideas, focusing on cross-modal information distillation for radiology report generation.}

{\color{black}
\subsubsection{MMFMs for Medical Image Comprehension}
The integration of MMFMs into medical image comprehension represents a significant leap forward, enhancing diagnostic precision and enabling efficient clinical decision-making. These models leverage vast datasets and multimodal learning techniques to interpret complex medical images, providing actionable insights to healthcare professionals. For instance, a study on pre-trained MMFMs highlighted their adaptability to medical imaging tasks through the use of specialized prompts, enabling the models to generalize effectively to unseen datasets and improve comprehension of medical visuals \cite{qin2022medical}.

Further advancements have been made in training multimodal assistants such as LLaVA-Med, a model designed specifically for biomedicine. By training on a large-scale dataset of biomedical figure captions, LLaVA-Med demonstrated superior performance on visual question answering tasks, showcasing its ability to provide nuanced and contextual insights into biomedical images \cite{llava_med}. Additionally, models like MedBLIP have introduced innovative paradigms by integrating 3D medical images with textual data, utilizing modules such as MedQFormer to bridge the gap between pre-trained language and vision models. This approach enhances the model's ability to answer complex medical queries and offers a scalable solution for adapting to the demands of medical image analysis \cite{chen2023medblip}.
}

\subsection{MMFMs for Diagnosis}
The integration of various data modalities - from imaging and pathology to clinical text - marks the transformative potential of MMFM in diagnostics. These models excel in tasks such as abnormality identification and disease classification, improving diagnostic accuracy and supporting more informed clinical decision making. 

\subsubsection{MMFMs for Abnormality identification}
Abnormality identification forms the basis of clinical diagnostics, which includes tasks such as detecting abnormal regions and segmenting lesions. MMFMs leverage their multimodal processing capabilities and advanced representation learning to overcome challenges such as limited labeled datasets and variability in human annotations. This section explores their contributions in two subdomains: abnormal region detection and lesion segmentation.

MMFMs have shown remarkable efficacy in identifying abnormal regions in various imaging modalities, including X-rays, CT scans, and MRI. For example, the CXRBase\cite{xu2024foundation} model uses SSL on more than 1.04 million unlabeled chest radiographs, followed by fine-tuning with labeled data. It achieves robust performance in multicenter datasets and excels in disease localization tasks, making it an indispensable tool for screening and early detection. Similarly, CT-CLIP\cite{CT-RATE}, a contrastive language-image pretraining framework, utilizes paired chest CT scans and radiology reports to identify abnormalities in a zero-shot setting. Trained on the CT-RATE dataset, CT-CLIP achieves AUROC values of up to 90\%, demonstrating adaptability to diverse clinical environments without task-specific fine-tuning.

Lesion segmentation, crucial for assessing disease burden and planning interventions, benefits immensely from MMFMs' cross-modal integration capabilities. The CLIP-Driven Universal Model\cite{cdum} uses CLIP embeddings to establish semantic relationships across anatomical structures, enabling the precise segmentation of 25 organs and 6 tumor types. The model's ability to generalize across datasets underscores its robustness in diverse scenarios. In cardiovascular applications, an AI model automates epicardial adipose tissue (EAT) volume and attenuation measurements from low-dose, unrefractory CT scans, completing these tasks in less than two seconds. This efficiency significantly reduces manual annotation workloads while providing actionable insights to predict cardiovascular risk\cite{miller2024ai}. Furthermore, RETFound\cite{RETFound}, a foundation model trained in more than 1.6 million unlabeled retinal images, excels in detecting abnormal regions and segmenting lesions in organ-specific and systemic diseases. Its adaptability across imaging modalities such as fundus photographs and OCT highlights its versatility and diagnostic precision.

In summary, MMFMs represent a paradigm shift in abnormality identification, seamlessly integrating imaging and textual data to achieve superior performance in abnormal region detection and lesion segmentation. These advancements not only improve diagnostic accuracy, but also streamline clinical workflows,

\subsubsection{MMFMs for disease classification}
MMFMs are transforming disease classification by integrating multimodal data for comprehensive diagnostic insights. HeartBEiT\cite{vaid2023foundational}, a foundational vision transformer, demonstrates exceptional accuracy in classifying cardiovascular conditions, such as atrial fibrillation and myocardial infarction, by integrating echocardiograms and patient histories. Its ability to handle noisy and sparse data highlights its practical relevance in real-world clinical settings. Similarly, MMFMs trained on multimodal imaging data have achieved high precision in diagnosing respiratory and cardiovascular conditions by using cross-modal relationships between imaging and textual reports\cite{quer2024potential}.

RET-CLIP\cite{retclip} achieves state-of-the-art performance in multiple datasets for retinal diseases, including diabetic retinopathy and glaucoma. Using tripartite optimization on the monocular, binocular, and patient levels, textual diagnostic reports are used to improve classification precision and generalization. Furthermore, the FLAIR\cite{silva2025foundation} model uses expert-annotated textual descriptions to classify retinal diseases such as macular degeneration and hypertensive retinopathy, effectively addressing complex multi-label classification tasks.

MMFMs such as the Virchow\cite{vorontsov2024foundation} model integrate whole-slide images with textual annotations to classify multiple cancer types, including rare subtypes, achieving specimen-level AUROC scores of 0.95. This enables the identification of histological subtypes and prognostic biomarkers. CHIEF\cite{wang2024pathology} employs weakly supervised learning to align attention with nuclear atypia and tissue morphology, achieving high accuracy in cancer subtype classification and patient outcome prediction. Mammo-CLIP\cite{chen2024mammo}, a CLIP-based vision-language framework, integrates multiview mammograms with text prompts to predict malignancy. Its efficient fine-tuning strategy requires only 1\% of the model parameters, offering a scalable solution to reduce false positives and variability between readers in breast cancer diagnostics. RadFM\cite{wu2023towards} uses 2D and 3D imaging data to classify rare diseases such as sarcoidosis and Langerhans cell histiocytosis. Its use of pre-trained multimodal embeddings ensures high diagnostic accuracy even in data-scarce scenarios.

In diagnostic workflows, MMFMs are instrumental in synthesis of diverse data streams, allowing more accurate and holistic diagnostic decision-making. By analyzing these complex multimodal datasets, MMFMs can uncover hidden correlations between imaging features and clinical outcomes, significantly improving early disease detection and personalized treatment planning, especially for few-shot and zero-shot tasks. These diagnostic models are crucial for accelerating diagnostic times and improving diagnostic consistency in different healthcare settings.

\subsection{MMFMs for treatment}

MMFMs can enhance treatment procedures through visualization, quantification, and controllability. MMFMs enable real-time annotation of key anatomical regions and surgical instruments in video streams during operations, performing both visualization and quantification tasks. They can send alerts when treatment steps deviate from the planned protocol or assist in retrieving relevant literature when surgeons encounter rare anatomical phenomena. Furthermore, MMFMs can use language models and knowledge graphs to incrementally reason through surgical tasks and dynamically update treatment plans\cite{deepbid,qagnn,surgicalgpt}, a crucial capability for handling rare cases\cite{longtail}. 

The controllability aspect is particularly significant in surgical robotics. Endo-FM\cite{Endo_FM} exemplifies a high precision foundational model designed for endoscopic video classification, segmentation, and detection, thus enhancing the visual capabilities of robotic surgery systems. Beyond visual enhancements, MMFMs, with their robust generative capabilities, can simulate and model surgical procedures, enabling surgeons to refine their skills before performing actual robotic surgeries on patients. This facilitates the integration of collaborative multirobot operations tailored to specific surgical goals\cite{palme,vima}. These advancements could accelerate the development and iteration of autonomous medical robots.

The incorporation of MMFMs into medical robotics presents a significant advancement. Robotic systems, especially those used in surgery and rehabilitation, derive significant advantages from the capability of MMFMs to process multimodal data streams, including visual, tactile, and auditory information, in real time. Using foundational models, these systems can improve decision making during robotic surgeries, optimize assisted movements, and tailor rehabilitation protocols based on individual patient data. The predictive capabilities of MMFMs enable robots to anticipate patient needs and responses, supporting more responsive and adaptive treatments. This capacity for data-driven decision making in real time improves the precision and safety of complex medical procedures, such as minimally invasive surgeries.

{\color{black}
\subsubsection{MMFMs for Treatment Decision-Making and Surgical Planning}

MMFM methods have shown transformative potential in medical treatment decision-making and surgical planning, particularly in the cardiovascular domain. \cite{schlesinger2020deep} emphasized the role of DL in cardiovascular risk stratification, proposing frameworks to evaluate the clinical utility of these models, highlighting their predictive power for patient-specific risks. \cite{Wehbe} reviewed DL applications in cardiovascular imaging, highlighting their ability to enhance diagnostic accuracy and efficiency in various modalities. \cite{zhou2024comprehensive} provided a comprehensive review of DL-based heart disease prediction models, comparing traditional and integrated DL algorithms, and focusing on real-time predictive capabilities for early intervention. Furthermore, \cite{williams2024artificial} explored the integration of AI and DL in cardiovascular CT, discussing their impact on image acquisition, processing, and analysis, as well as their potential to improve the entire clinical pathway. 

The integration of medical imaging models into treatment decision making and surgical planning has significantly advanced, particularly within the cardiovascular domain. \cite{moshawrab2023reviewing} highlighted the role of multimodal ML in improving diagnostic accuracy by integrating data from imaging, clinical records, and other sources, addressing the inherent heterogeneity of medical datasets. This integration facilitates more comprehensive analyses and enables precise patient-specific care strategies.

In cardiovascular applications, \cite{alaa2019demystifying} emphasized the predictive capabilities of ML in assessing the risk of coronary artery disease, using clinical and imaging data to stratify patient risks more effectively. Furthermore, \cite{esteva2017dermatologist} demonstrated the application of CNNs in medical imaging tasks, providing foundational models for integrating imaging modalities into broader diagnostic pipelines. These advancements have informed surgical planning by improving the precision of preoperative evaluations and enhancing intraoperative guidance.
}

{\color{black}
\subsubsection{MMFMs for Surgical Robotics}

The integration of medical imaging models into surgical robotics has significantly improved the precision and efficacy of surgical interventions. This synergy enables real-time visualization and navigation, improving the surgeon's ability to perform complex procedures with greater accuracy.

A pivotal development in this domain is the creation of realistic surgical image datasets through 3D gaussian splatting. This technique involves training Gaussian models to represent surgical scenes and instruments separately, allowing for the generation of high-fidelity synthetic surgical scenarios. Such datasets are instrumental in training and validating surgical robots, ensuring that they can operate effectively in diverse clinical environments. \cite{zeng2024realistic}

\cite{ahmed2024deep} highlighted the role of advanced CNNs in improving instrument segmentation accuracy, citing the use of models such as U-Net and Mask R-CNN to delineate surgical tools in complex scenes \cite{ronneberger2015u,he2017mask}. These models enable real-time guidance, enhancing the precision and safety of robotic-assisted surgeries.

\cite{hussain2022deep} emphasized the integration of multimodal datasets, combining video streams and kinematic data to improve performance in gesture segmentation tasks \cite{dipietro2016recognizing}. The review also highlighted advancements in leveraging the JIGSAWS dataset for the evaluation of surgical skills, where DL-based methods outperformed traditional techniques in trajectory prediction and surgical gesture analysis \cite{ahmidi2017dataset}. In addition, attention mechanisms, such as those proposed by \cite{vaswani2017attention}, have been applied to improve temporal consistency in video-based models for the analysis of surgical processes.}

\section{Future Direction}

As MMFMs continue to evolve, several key areas require focused research and development to enhance their performance, scalability, and applicability in healthcare. The future of MMFMs will be shaped by advancements in data and computation, capability and sustainability, reliability and interpretability, regulation and privacy. Addressing these challenges is critical for developing models that are adaptable, efficient, and trustworthy in various medical contexts.

\subsection{Data and Computation}
MMFMs depend heavily on the availability and quality of diverse large-scale multimodal datasets, which integrate information from various sources such as medical imaging, genomics, clinical records, and real-time patient monitoring. However, aggregating and standardizing such data pose significant challenges due to the fragmentation of healthcare data between institutions, regions, and countries, where differing data formats and privacy regulations may restrict data-sharing practices. As we move forward, a key area of focus will be the development of internationally recognized standards and frameworks that support the seamless integration of medical data from disparate sources. These datasets must be representative, encompassing diverse patient populations and medical conditions, to avoid biases that could lead to poor generalization in MMFMs.

In addition, MMFMs research must also address the immense computational power required to process and learn from these multimodal datasets. The current training of such models is resource-intensive, necessitating advanced GPUs and cloud computing infrastructure. Techniques like federated learning and edge computing, which allow distributed model training without centralizing sensitive patient data, will be pivotal. In addition, reducing the environmental footprint of training large-scale models is crucial. Optimizing models through techniques such as model pruning, knowledge distillation, and the development of efficient hardware such as AI accelerators can reduce computational costs. These innovations could pave the way for more sustainable and scalable MMFMs that can be trained and deployed in real-world healthcare settings.

\subsection{Capability and Sustainability}
MMFMs are expected not only to deliver high performance in diverse medical applications but also to demonstrate sustainable learning capabilities over time. This implies that MMFMs should be designed to continuously evolve and improve as they encounter new data and tasks, thus improving their adaptability to evolving medical needs without having to undergo complete retraining. Lifelong learning, as a concept, is crucial for MMFMs, as it allows them to retain prior knowledge while learning new information, a capability that is particularly significant in the medical field, where new medical knowledge and technologies emerge regularly. A major challenge in this area is the issue of catastrophic forgetting, in which the model's performance on previously learned tasks deteriorates as new information is integrated. To overcome this, future research should focus on advanced lifelong learning strategies that help MMFMs integrate new information while preserving the knowledge they have already acquired.

In addition to continuous learning, sustainability for MMFMs also refers to their ability to generalize across a wide range of medical environments and tasks. These models should be highly adaptable and function efficiently in different clinical settings, whether in resource-rich hospitals or smaller, less equipped facilities. This requires the development of flexible architectures that can be fine-tuned for specific tasks with minimal additional data, ensuring their practical relevance in varied scenarios. A significant future direction is improving the capacity of MMFMs to leverage unannotated or sparsely labeled datasets, thus enabling the models to learn sustainably from real-world clinical data while reducing their dependence on fully annotated datasets. Such advancements will not only enhance the scalability of the models but also ensure their longevity and usability across diverse medical domains.

\subsection{Reliability and Interpretability}
As MMFMs are deployed in clinical settings, the demand for high levels of reliability and interpretability will be paramount. These models must function reliably across different patient populations, healthcare environments, and imaging modalities, minimizing the risks of bias or misinterpretation. The development of MMFMs that can adapt to local healthcare settings while maintaining a high level of accuracy in various conditions is vital. One potential approach is the inclusion of domain-specific calibration processes that allow models to self-adjust based on the demographic or environmental data of the population they are serving.

For healthcare professionals to trust MMFM, they must provide clear, understandable insights into how they arrive at their conclusions. This means moving beyond “black box” AI models to systems that provide visual or textual explanations for their decisions. Methods like attention mechanisms, which highlight key areas of an image or report that led to a specific decision, or saliency maps that visually represent which parts of input were most influential, will be critical for improving interpretability. In addition, ensuring the ethical use of MMFMs will require ongoing validation and performance monitoring, particularly when used in high-stakes scenarios such as surgery or critical diagnostics.

\subsection{Regulation and Privacy}
Given the sensitive nature of medical data, regulation, and privacy considerations are crucial for MMFMs. Current privacy laws impose strict restrictions on how patient data can be used, especially for training AI models. As MMFMs evolve, compliance with these regulations must remain a top priority, which means that innovative privacy-preserving techniques, such as differential privacy, secure multiparty computation, and homomorphic encryption, will become more prominent. These methods allow for data sharing and model training without compromising patient privacy.

Moreover, as AI-specific regulatory frameworks for healthcare emerge, the certification and auditing of MMFMs will become an integral part of their deployment. Regulatory bodies will need to establish clear guidelines for AI in healthcare care, covering aspects such as transparency, fairness, and accountability of models. In addition, ownership and consent for data usage will be critical issues to address. Patients must have control over their personal data, including how they are used in model training and deployment. Ensuring that MMFMs align with evolving regulatory standards while respecting individual privacy rights will be the key to fostering public trust and ensuring ethical deployment in clinical environments.

\section{Conclusion}
In this review, we have thoroughly examined the latest advancements and challenges in the development and application of MMFMs in the field of medical artificial intelligence. MMFMs have introduced significant breakthroughs in medical image analysis, diagnosis, and treatment planning by integrating data from various modalities. This integration has substantially improved the precision and efficiency of healthcare delivery.

Across various domains, MMFMs demonstrate remarkable generalizability and robustness.
In medical imaging, MMFMs leverage proxy tasks such as multiview learning and contrastive learning to capture fine anatomical details, leading to improved performance in segmentation and classification tasks. In medical diagnosis, MMFMs excel at integrating multimodal information, reducing the reliance on labeled data and enabling unsupervised and semi-supervised learning models to perform well even in data-scarce environments.

Despite their notable achievements, MMFMs still encounter several significant challenges, including limitations in generalization, substantial data, and computational requirements, among others. Future research should focus on improving data standardization and integration, improving sustainable learning mechanisms, and improving model reliability in diverse clinical settings. Addressing these challenges will be crucial for the widespread adoption of MMFMs in real-world healthcare applications, ultimately leading to more efficient, accurate, and equitable medical care.

Looking ahead, the ongoing evolution of MMFMs will undoubtedly shape the future of medical AI, advancing us towards the realization of artificial general intelligence in healthcare. As research continues to refine the models’ data and computational efficiency, sustainability, reliability, and regulatory compliance, MMFMs will play a crucial role in transforming healthcare delivery and empowering clinicians with intelligent tools that optimize diagnostics, treatment planning, and overall patient care. By addressing the challenges ahead, MMFMs will ultimately pave the way for more efficient, accurate, and equitable healthcare solutions that are adaptable to a wide range of medical contexts.

\section*{Acknowledgments}
This study was supported by the National Natural Science Foundation of China (No. 82090050, No. 82090053) and Tsinghua University Initiative Scientific Research Program of Precision Medicine (No.2022ZLA006).

\newpage

\vspace{11pt}

\vfill


\begin{thebibliography}{999}
\bibliographystyle{IEEEtran}

\bibitem{rajpurkar2022ai} P.~Rajpurkar, E.~Chen, O.~Banerjee, and E.~J. Topol, ``Ai in health and
medicine,'' \emph{Nature Medicine}, vol.~28, no.~1, pp. 31--38, 2022.

\bibitem{tu2024towards} T.~Tu, S.~Azizi, D.~Driess, M.~Schaekermann, M.~Amin, P.-C. Chang, A.~Carroll,
C.~Lau, R.~Tanno, I.~Ktena \emph{et~al.}, ``Towards generalist biomedical
ai,'' \emph{NEJM AI}, vol.~1, no.~3, p. AIoa2300138, 2024.

\bibitem{acosta2022multimodal} J.~N. Acosta, G.~J. Falcone, P.~Rajpurkar, and E.~J. Topol, ``Multimodal
biomedical ai,'' \emph{Nature Medicine}, vol.~28, no.~9, pp. 1773--1784,
2022.

\bibitem{bommasani2021opportunities} R.~Bommasani, D.~A. Hudson, E.~Adeli, R.~Altman, S.~Arora, S.~von Arx, M.~S.
Bernstein, J.~Bohg, A.~Bosselut, E.~Brunskill \emph{et~al.}, ``On the
opportunities and risks of foundation models,'' \emph{arXiv preprint
	arXiv:2108.07258}, 2021.

\bibitem{bert} J.~Devlin, M.~Chang, K.~Lee, and K.~Toutanova, ``{BERT:} pretraining of deep
bidirectional transformers for language understanding,'' \emph{arXiv preprint
	arXiv:abs/1810.04805}, 2018.

\bibitem{CLIP} A.~Radford, J.~W. Kim, C.~Hallacy, A.~Ramesh, G.~Goh, S.~Agarwal, G.~Sastry,
A.~Askell, P.~Mishkin, J.~Clark \emph{et~al.}, ``Learning transferable visual
models from natural language supervision,'' in \emph{International Conference
	on Machine Learning}.\hskip 1em plus 0.5em minus 0.4em\relax PMLR, 2021, pp.
8748--8763.

\bibitem{dall_e} A.~Ramesh, M.~Pavlov, G.~Goh, S.~Gray, C.~Voss, A.~Radford, M.~Chen, and
I.~Sutskever, ``Zero-shot text-to-image generation,'' \emph{arXiv preprint
	arXiv:2102.12092}, 2021.

\bibitem{yang2023dawn} Z.~Yang, L.~Li, K.~Lin, J.~Wang, C.-C. Lin, Z.~Liu, and L.~Wang, ``The dawn of
lmms: Preliminary explorations with gpt-4v (ision),'' \emph{arXiv preprint
	arXiv:2309.17421}, vol.~9, no.~1, p.~1, 2023.

 \bibitem{area_1}
B.~Jiang, X.~Chen, W.~Liu, J.~Yu, G.~Yu, and T.~Chen, ``Motiongpt: Human motion
  as a foreign language,'' \emph{Advances in Neural Information Processing
  Systems}, vol.~36, pp. 20\,067--20\,079, 2023.

\bibitem{area_2}
J.~Ye, A.~Hu, H.~Xu, Q.~Ye, M.~Yan, Y.~Dan, C.~Zhao, G.~Xu, C.~Li, J.~Tian
  \emph{et~al.}, ``mplug-docowl: Modularized multimodal large language model
  for document understanding,'' \emph{arXiv preprint arXiv:2307.02499}, 2023.

\bibitem{area_3}
H.~Zhang, X.~Li, and L.~Bing, ``Video-llama: An instruction-tuned audio-visual
  language model for video understanding,'' \emph{arXiv preprint
  arXiv:2306.02858}, 2023.

\bibitem{area_4}
J.~Roberts, T.~L{\"u}ddecke, R.~Sheikh, K.~Han, and S.~Albanie, ``Charting new
  territories: Exploring the geographic and geospatial capabilities of
  multimodal llms,'' in \emph{Proceedings of the IEEE/CVF Conference on
  Computer Vision and Pattern Recognition}, 2024, pp. 554--563.

\bibitem{llava_med} C.~Li, C.~Wong, S.~Zhang, N.~Usuyama, H.~Liu, J.~Yang, T.~Naumann, H.~Poon, and
J.~Gao, ``Llava-med: Training a large language-and-vision assistant for
biomedicine in one day,'' \emph{arXiv preprint
	arXiv:2306.00890}, 2023.

\bibitem{moor2023foundation} M.~Moor, O.~Banerjee, Z.~S.~H. Abad, H.~M. Krumholz, J.~Leskovec, E.~J. Topol,
and P.~Rajpurkar, ``Foundation models for generalist medical artificial
intelligence,'' \emph{Nature}, vol. 616, no. 7956, pp. 259--265, 2023.

\bibitem{zhang2024challenges} S.~Zhang and D.~Metaxas, ``On the challenges and perspectives of foundation
models for medical image analysis,'' \emph{Medical Image Analysis}, vol.~91,
p. 102996, 2024.

\bibitem{vaswani2017attention} A.~Vaswani, ``Attention is all you need,'' \emph{Advances in Neural Information
	 Processing Systems}, 2017.

\bibitem{deng2009imagenet} J.~Deng, W.~Dong, R.~Socher, L.-J. Li, K.~Li, and L.~Fei-Fei, ``Imagenet: A
large-scale hierarchical image database,'' in \emph{Proceedings of the
	IEEE/CVF Conference on Computer Vision and Pattern Recognition}, 2009, pp. 248--255.

\bibitem{he2020momentum} K.~He, H.~Fan, Y.~Wu, S.~Xie, and R.~Girshick, ``Momentum contrast for
unsupervised visual representation learning,'' in \emph{Proceedings of the
	IEEE/CVF Conference on Computer Vision and Pattern Recognition}, 2020, pp.
9729--9738.

\bibitem{chen2020simple} T.~Chen, S.~Kornblith, M.~Norouzi, and G.~Hinton, ``A simple framework for
contrastive learning of visual representations,'' in \emph{International
	Conference on Machine Learning}.\hskip 1em plus 0.5em minus 0.4em\relax PMLR,
2020, pp. 1597--1607.

\bibitem{dosovitskiy2020image} A.~Dosovitskiy, ``An image is worth 16x16 words: Transformers for image
recognition at scale,'' \emph{arXiv preprint arXiv:2010.11929}, 2020.


\bibitem{model_GB} G.~Bhatnagar, Q.~J. Wu, and Z.~Liu, ``A new contrast based multimodal medical
image fusion framework,'' \emph{Neurocomputing}, vol. 157, pp. 143--152,
2015. [Online]. Available:
\url{https://www.sciencedirect.com/science/article/pii/S0925231215000466}

\bibitem{ts_ZG} Z.~Guo, X.~Li, H.~Huang, N.~Guo, and Q.~Li, ``Deep learning-based image
segmentation on multimodal medical imaging,'' \emph{IEEE Transactions on
	Radiation and Plasma Medical Sciences}, vol.~3, no.~2, pp. 162--169, 2019.

\bibitem{cpn_XL} X.~Liu, Y.~Pan, X.~Zhang, Y.~Sha, S.~Wang, H.~Li, and J.~Liu, ``A deep learning
model for classification of parotid neoplasms based on multimodal magnetic
resonance image sequences,'' \emph{The Laryngoscope}, vol. 133, no.~2, pp.
327--335, 2023. [Online]. Available:
\url{https://onlinelibrary.wiley.com/doi/abs/10.1002/lary.30154}

\bibitem{YM} Y.~Ming, X.~Dong, J.~Zhao, Z.~Chen, H.~Wang, and N.~Wu, ``Deep learning-based
multimodal image analysis for cervical cancer detection,'' \emph{Methods},
vol. 205, pp. 46--52, 2022.

\bibitem{wavelet} G.~Qu, D.~Zhang, and P.~Yan, ``Medical image fusion by wavelet transform
modulus maxima,'' \emph{Optics Express}, vol.~9, no.~4, pp. 184--190, 2001.

\bibitem{curvelet} F.~Ali, I.~El-Dokany, A.~Saad, and F.~Abd El-Samie, ``Curvelet fusion of mr and
ct images,'' \emph{Progress In Electromagnetics Research C}, vol.~3, pp.
215--224, 2008.

\bibitem{nsct} T.~Li and Y.~Wang, ``Biological image fusion using a nsct based variable-weight
method,'' \emph{Information Fusion}, vol.~12, no.~2, pp. 85--92, 2011.

\bibitem{fuzzy_SD} S.~Das and M.~K. Kundu, ``A neuro-fuzzy approach for medical image fusion,''
\emph{IEEE Transactions on Biomedical Engineering}, vol.~60, no.~12, pp.
3347--3353, 2013.

\bibitem{fuzzy_PB} P.~Balasubramaniam and V.~Ananthi, ``Image fusion using intuitionistic fuzzy
sets,'' \emph{Information Fusion}, vol.~20, pp. 21--30, 2014.

\bibitem{hmf} J.~Lapuyade-Lahorgue, J.-H. Xue, and S.~Ruan, ``Segmenting multi-source images
using hidden markov fields with copula-based multivariate statistical
distributions,'' \emph{IEEE Transactions on Image Processing}, vol.~26,
no.~7, pp. 3187--3195, 2017.

\bibitem{believe_m} P.~Smets, ``The combination of evidence in the transferable belief model,''
\emph{IEEE Transactions on Pattern Analysis and Machine Intelligence},
vol.~12, no.~5, pp. 447--458, 1990.

\bibitem{believe_f} C.~Lian, S.~Ruan, T.~Den{\oe}ux, H.~Li, and P.~Vera, ``Joint tumor segmentation
in pet-ct images using co-clustering and fusion based on belief functions,''
\emph{IEEE Transactions on Image Processing}, vol.~28, no.~2, pp. 755--766,
2018.

\bibitem{ml_NS} N.~Srivastava and R.~R. Salakhutdinov, ``Multimodal learning with deep
boltzmann machines,'' \emph{Advances in Neural Information Processing
	Systems}, vol.~25, 2012.

\bibitem{ml_HC} H.~Cai, R.~Verma, Y.~Ou, S.-k. Lee, E.~R. Melhem, and C.~Davatzikos,
``Probabilistic segmentation of brain tumors based on multi-modality magnetic
resonance images,'' in \emph{2007 4th IEEE International Symposium on
	Biomedical Imaging: From Nano to Macro}.\hskip 1em plus 0.5em minus
0.4em\relax IEEE, 2007, pp. 600--603.

\bibitem{ml_AV} A.~Vazquez-Reina, M.~Gelbart, D.~Huang, J.~Lichtman, E.~Miller, and H.~Pfister,
  ``Segmentation fusion for connectomics,'' in \emph{2011 International
	  Conference on Computer Vision}.\hskip 1em plus 0.5em minus 0.4em\relax IEEE,
  2011, pp. 177--184.

\bibitem{seg_bt} F.~Fang, Y.~Yao, T.~Zhou, G.~Xie, and J.~Lu, ``Self-supervised multi-modal
hybrid fusion network for brain tumor segmentation,'' \emph{IEEE Journal of
	Biomedical and Health Informatics}, vol.~26, no.~11, pp. 5310--5320, 2021.

\bibitem{seg_ga} T.~Spaide, J.~Jiang, J.~Patil, N.~Anegondi, V.~Steffen, M.~G. Kawczynski, E.~M.
Newton, C.~Rabe, S.~S. Gao, A.~Y. Lee \emph{et~al.}, ``Geographic atrophy
segmentation using multimodal deep learning,'' \emph{Translational Vision
	Science \& Technology}, vol.~12, no.~7, pp. 10--10, 2023.

\bibitem{seg_non_paired_organ} H.~Liu, Y.~Zhuang, E.~Song, X.~Xu, G.~Ma, C.~Cetinkaya, and C.-C. Hung, ``A
modality-collaborative convolution and transformer hybrid network for
unpaired multi-modal medical image segmentation with limited annotations,''
\emph{Medical Physics}, vol.~50, no.~9, pp. 5460--5478, 2023.

\bibitem{seg_sl} Y.~Gheibi, K.~Shirini, S.~N. Razavi, M.~Farhoudi, and T.~Samad-Soltani,
``Cnn-res: deep learning framework for segmentation of acute ischemic stroke
lesions on multimodal mri images,'' \emph{BMC Medical Informatics and
	Decision Making}, vol.~23, no.~1, p. 192, 2023.

\bibitem{class_az} A.~Massalimova and H.~A. Varol, ``Input agnostic deep learning for
alzheimer¡¯s disease classification using multimodal mri images,'' in
\emph{2021 43rd Annual International Conference of the IEEE Engineering in
	Medicine \& Biology Society (EMBC)}.\hskip 1em plus 0.5em minus 0.4em\relax
IEEE, 2021, pp. 2875--2878.

\bibitem{class_bc} K.~Takahashi, T.~Fujioka, J.~Oyama, M.~Mori, E.~Yamaga, Y.~Yashima, T.~Imokawa,
A.~Hayashi, Y.~Kujiraoka, J.~Tsuchiya \emph{et~al.}, ``Deep learning using
multiple degrees of maximum-intensity projection for pet/ct image
classification in breast cancer,'' \emph{Tomography}, vol.~8, no.~1, pp.
131--141, 2022.

\bibitem{class_md} Q.~Chen, T.~D. Keenan, A.~Allot, Y.~Peng, E.~Agr¨®n, A.~Domalpally, C.~C.~W.
Klaver, D.~T. Luttikhuizen, M.~H. Colyer, C.~A. Cukras, H.~E. Wiley,
M.~Teresa~Magone, C.~Cousineau-Krieger, W.~T. Wong, Y.~Zhu, E.~Y. Chew,
Z.~Lu, and for~the AREDS2 Deep Learning Research~Group, ``{Multimodal,
	multitask, multiattention (M3) deep learning detection of reticular
	pseudodrusen: Toward automated and accessible classification of age-related
	macular degeneration},'' \emph{Journal of the American Medical Informatics
	Association}, vol.~28, no.~6, pp. 1135--1148, 04 2021. [Online]. Available:
\url{https://doi.org/10.1093/jamia/ocaa302}

\bibitem{gen_mri} R.~Touati and S.~Kadoury, ``Bidirectional feature matching based on deep
pairwise contrastive learning for multiparametric mri image synthesis,''
\emph{Physics in Medicine \& Biology}, vol.~68, no.~12, p. 125010, jun 2023.
[Online]. Available: \url{https://dx.doi.org/10.1088/1361-6560/acda78}

\bibitem{gen_video} J.~Fan, Z.~Liu, D.~Yang, J.~Qiao, J.~Zhao, J.~Wang, and W.~Hu, ``Multimodal
image translation via deep learning inference model trained in video
domain,'' \emph{BMC Medical Imaging}, vol.~22, no.~1, p. 124, 2022.

\bibitem{diag_az} M.~Golovanevsky, C.~Eickhoff, and R.~Singh, ``Multimodal attention-based deep
learning for alzheimer¡¯s disease diagnosis,'' \emph{Journal of the American
	Medical Informatics Association}, vol.~29, no.~12, pp. 2014--2022, 2022.

\bibitem{diag_koa} Y.~Hu, J.~Tang, S.~Zhao, and Y.~Li, ``Deep learning-based multimodal 3 t mri
for the diagnosis of knee osteoarthritis,'' \emph{Computational and
	Mathematical Methods in Medicine}, vol. 2022, no.~1, p. 7643487, 2022.

\bibitem{diag_ras} X.~Wang, S.~Cai, H.~Wang, J.~Li, and Y.~Yang, ``Deep-learning-based renal
artery stenosis diagnosis via multimodal fusion,'' \emph{Journal of Applied
	Clinical Medical Physics}, vol.~25, no.~3, p. e14298, 2024.

\bibitem{pred_cir} X.~Wang, Y.~Jiang, H.~Chen, T.~Zhang, Z.~Han, C.~Chen, Q.~Yuan, W.~Xiong,
W.~Wang, G.~Li \emph{et~al.}, ``Cancer immunotherapy response prediction from
multi-modal clinical and image data using semi-supervised deep learning,''
\emph{Radiotherapy and Oncology}, vol. 186, p. 109793, 2023.

\bibitem{pred_mvi} F.~Wang, Q.~Chen, Y.~Chen, Y.~Zhu, Y.~Zhang, D.~Cao, W.~Zhou, X.~Liang,
Y.~Yang, L.~Lin \emph{et~al.}, ``A novel multimodal deep learning model for
preoperative prediction of microvascular invasion and outcome in
hepatocellular carcinoma,'' \emph{European Journal of Surgical Oncology},
vol.~49, no.~1, pp. 156--164, 2023.

%

\bibitem{pred_nac} S.~Joo, E.~S. Ko, S.~Kwon, E.~Jeon, H.~Jung, J.-Y. Kim, M.~J. Chung, and Y.-H.
Im, ``Multimodal deep learning models for the prediction of pathologic
response to neoadjuvant chemotherapy in breast cancer,'' \emph{Scientific
	Reports}, vol.~11, no.~1, p. 18800, 2021.

\bibitem{eval_doc} K.~Kasa, D.~Burns, M.~G. Goldenberg, O.~Selim, C.~Whyne, and M.~Hardisty,
``Multi-modal deep learning for assessing surgeon technical skill,''
\emph{Sensors}, vol.~22, no.~19, 2022. [Online]. Available:
\url{https://www.mdpi.com/1424-8220/22/19/7328}

\bibitem{if_bt} S.~Pereira, A.~Pinto, V.~Alves, and C.~A. Silva, ``Brain tumor segmentation
using convolutional neural networks in mri images,'' \emph{IEEE Transactions
	on Medical Imaging}, vol.~35, no.~5, pp. 1240--1251, 2016.

\bibitem{if_bg} S.~Cui, L.~Mao, J.~Jiang, C.~Liu, and S.~Xiong, ``Automatic semantic
segmentation of brain gliomas from mri images using a deep cascaded neural
network,'' \emph{Journal of Healthcare Engineering}, vol. 2018, no.~1, p.
4940593, 2018.

\bibitem{if_bt_brats_2017} F.~Isensee, P.~Kickingereder, W.~Wick, M.~Bendszus, and K.~H. Maier-Hein,
``Brain tumor segmentation and radiomics survival prediction: Contribution to
the brats 2017 challenge,'' in \emph{Brainlesion: Glioma, Multiple Sclerosis,
	Stroke and Traumatic Brain Injuries: Third International Workshop, BrainLes
	2017, Held in Conjunction with MICCAI 2017, Quebec City, QC, Canada,
	September 14, 2017, Revised Selected Papers 3}.\hskip 1em plus 0.5em minus
0.4em\relax Springer, 2018, pp. 287--297.

\bibitem{if_bt_miccai} A.~Myronenko, ``3d mri brain tumor segmentation using autoencoder
regularization,'' in \emph{Brainlesion: Glioma, Multiple Sclerosis, Stroke
	and Traumatic Brain Injuries: 4th International Workshop, BrainLes 2018, Held
	in Conjunction with MICCAI 2018, Granada, Spain, September 16, 2018, Revised
	Selected Papers, Part II 4}.\hskip 1em plus 0.5em minus 0.4em\relax Springer,
2019, pp. 311--320.

\bibitem{matr} W.~Tang, F.~He, Y.~Liu, and Y.~Duan, ``Matr: Multimodal medical image fusion
via multiscale adaptive transformer,'' \emph{IEEE Transactions on Image
	Processing}, vol.~31, pp. 5134--5149, 2022.


\bibitem{m4oe} Y.~Jiang and Y.~Shen, ``M4oE: A foundation model for medical multimodal image
segmentation with mixture of experts,'' \emph{arXiv preprint
	arXiv:2405.09446}, 2024.

\bibitem{mm_fusion_seg} X.~Fan, L.~Liu, and H.~Zhang, ``Multimodal information interaction for medical
image segmentation,'' \emph{arXiv preprint arXiv:2404.16371}, 2024.

\bibitem{of_bi} D.~Nie, L.~Wang, Y.~Gao, and D.~Shen, ``Fully convolutional networks for
multi-modality isointense infant brain image segmentation,'' in \emph{2016
	IEEE 13Th International Symposium on Biomedical Imaging (ISBI)}.\hskip 1em
plus 0.5em minus 0.4em\relax IEEE, 2016, pp. 1342--1345.

\bibitem{of_eb} L.~Rokach, ``Ensemble-based classifiers,'' \emph{Artificial Intelligence
	Review}, vol.~33, pp. 1--39, 2010.

\bibitem{of_rbts} K.~Kamnitsas, W.~Bai, E.~Ferrante, S.~McDonagh, M.~Sinclair, N.~Pawlowski,
M.~Rajchl, M.~Lee, B.~Kainz, D.~Rueckert \emph{et~al.}, ``Ensembles of
multiple models and architectures for robust brain tumour segmentation,'' in
\emph{Brainlesion: Glioma, Multiple Sclerosis, Stroke and Traumatic Brain
	Injuries: Third International Workshop, BrainLes 2017, Held in Conjunction
	with MICCAI 2017, Quebec City, QC, Canada, September 14, 2017, Revised
	Selected Papers 3}.\hskip 1em plus 0.5em minus 0.4em\relax Springer, 2018,
pp. 450--462.

\bibitem{scale_law}
J.~Kaplan, S.~McCandlish, T.~Henighan, T.~B. Brown, B.~Chess, R.~Child,
  S.~Gray, A.~Radford, J.~Wu, and D.~Amodei, ``Scaling laws for neural language
  models,'' \emph{arXiv preprint arXiv:2001.08361}, 2020.

\bibitem{MedNLI} A.~Romanov and C.~Shivade, ``Lessons from natural language inference in the
clinical domain,'' \emph{arXiv preprint arXiv:1808.06752}, 2018.

\bibitem{SEER} S.~Dubey, G.~Tiwari, S.~Singh, S.~Goldberg, and E.~Pinsky, ``Using machine
learning for healthcare treatment planning,'' \emph{Frontiers in Artificial
	Intelligence}, vol.~6, p. 1124182, 2023.

\bibitem{MIMIC3} A.~E. Johnson, T.~J. Pollard, L.~Shen, L.-w.~H. Lehman, M.~Feng, M.~Ghassemi,
B.~Moody, P.~Szolovits, L.~Anthony~Celi, and R.~G. Mark, ``Mimic-iii, a
freely accessible critical care database,'' \emph{Scientific Data}, vol.~3,
no.~1, pp. 1--9, 2016.

\bibitem{MeQSum} A.~B. Abacha and D.~Demner-Fushman, ``On the summarization of consumer health
questions,'' in \emph{Proceedings of the 57th Annual Meeting of the
	Association for Computational Linguistics}, 2019, pp. 2228--2234.

\bibitem{hcm} G.~Zeng, W.~Yang, Z.~Ju, Y.~Yang, S.~Wang, R.~Zhang, M.~Zhou, J.~Zeng, X.~Dong,
R.~Zhang \emph{et~al.}, ``Meddialog: Large-scale medical dialogue datasets,''
in \emph{Proceedings of the 2020 Conference on Empirical Methods in Natural
	Language Processing (EMNLP)}, 2020, pp. 9241--9250.

\bibitem{cxr} S.~Jaeger, S.~Candemir, S.~Antani, Y.-X.~J. W{\'a}ng, P.-X. Lu, and G.~Thoma,
``Two public chest x-ray datasets for computer-aided screening of pulmonary
diseases,'' \emph{Quantitative Imaging in Medicine and Surgery}, vol.~4,
no.~6, p. 475, 2014.

\bibitem{cdcm} R.~S. Lee, F.~Gimenez, A.~Hoogi, K.~K. Miyake, M.~Gorovoy, and D.~L. Rubin, ``A
curated mammography data set for use in computer-aided detection and
diagnosis research,'' \emph{Scientific Data}, vol.~4, no.~1, pp. 1--9, 2017.

\bibitem{mmr} J.~Yang, R.~Shi, D.~Wei, Z.~Liu, L.~Zhao, B.~Ke, H.~Pfister, and B.~Ni,
``Medmnist v2-a large-scale lightweight benchmark for 2d and 3d biomedical
image classification,'' \emph{Scientific Data}, vol.~10, no.~1, p.~41, 2023.

\bibitem{cdum} J.~Liu, Y.~Zhang, J.-N. Chen, J.~Xiao, Y.~Lu, B.~A~Landman, Y.~Yuan, A.~Yuille,
Y.~Tang, and Z.~Zhou, ``Clip-driven universal model for organ segmentation
and tumor detection,'' in \emph{Proceedings of the IEEE/CVF International
	Conference on Computer Vision}, 2023, pp. 21\,152--21\,164.

\bibitem{aba8k} C.~Qu, T.~Zhang, H.~Qiao, Y.~Tang, A.~L. Yuille, Z.~Zhou \emph{et~al.},
``Abdomenatlas-8k: Annotating 8,000 ct volumes for multi-organ segmentation
in three weeks,'' \emph{Advances in Neural Information  Processing Systems},
vol.~36, 2024.

\bibitem{medsam} J.~Ma, Y.~He, F.~Li, L.~Han, C.~You, and B.~Wang, ``Segment anything in medical
images,'' \emph{Nature Communications}, vol.~15, no.~1, p. 654, 2024.

\bibitem{sammed3d} H.~Wang, P.~K.~A. Vasu, F.~Faghri, R.~Vemulapalli, M.~Farajtabar, S.~Mehta,
M.~Rastegari, O.~Tuzel, and H.~Pouransari, ``Sam-clip: Merging vision
foundation models towards semantic and spatial understanding,'' in
\emph{Proceedings of the IEEE/CVF Conference on Computer Vision and Pattern
	Recognition}, 2024, pp. 3635--3647.

\bibitem{EyeFound} D.~Shi, W.~Zhang, X.~Chen, Y.~Liu, J.~Yang, S.~Huang, Y.~C. Tham, Y.~Zheng, and
M.~He, ``Eyefound: A multimodal generalist foundation model for ophthalmic
imaging,'' \emph{arXiv preprint arXiv:2405.11338}, 2024.

\bibitem{RETFound} Y.~Zhou, M.~A. Chia, S.~K. Wagner, M.~S. Ayhan, D.~J. Williamson, R.~R.
Struyven, T.~Liu, M.~Xu, M.~G. Lozano, P.~Woodward-Court \emph{et~al.}, ``A
foundation model for generalizable disease detection from retinal images,''
\emph{Nature}, vol. 622, no. 7981, pp. 156--163, 2023.

\bibitem{swinunetr} A.~Hatamizadeh, V.~Nath, Y.~Tang, D.~Yang, H.~R. Roth, and D.~Xu, ``Swin unetr:
Swin transformers for semantic segmentation of brain tumors in mri images,''
in \emph{International MICCAI Brainlesion Workshop}.\hskip 1em plus 0.5em
minus 0.4em\relax Springer, 2021, pp. 272--284.

\bibitem{da} J.~M.~J. Valanarasu, Y.~Tang, D.~Yang, Z.~Xu, C.~Zhao, W.~Li, V.~M. Patel,
B.~Landman, D.~Xu, Y.~He \emph{et~al.}, ``Disruptive autoencoders: Leveraging
low-level features for 3d medical image pretraining,'' \emph{arXiv preprint
	arXiv:2307.16896}, 2023.

\bibitem{roco} O.~Pelka, S.~Koitka, J.~R{\"u}ckert, F.~Nensa, and C.~M. Friedrich, ``Radiology
objects in context (roco): a multimodal image dataset,'' in
\emph{Intravascular Imaging and Computer Assisted Stenting and Large-Scale
	Annotation of Biomedical Data and Expert Label Synthesis: 7th Joint
	International Workshop, CVII-STENT 2018 and Third International Workshop,
	LABELS 2018, Held in Conjunction with MICCAI 2018, Granada, Spain, September
	16, 2018, Proceedings 3}.\hskip 1em plus 0.5em minus 0.4em\relax Springer,
2018, pp. 180--189.

\bibitem{medicat} S.~M. B. B. M. v. Z. S. P. S. S. M.~G. Sanjay~Subramanian, Lucy Lu~Wang and
H.~Hajishirzi, ``{MedICaT: A Dataset of Medical Images, Captions, and Textual
	References},'' in \emph{Findings of EMNLP}, 2020.

\bibitem{pmcoa} W.~Lin, Z.~Zhao, X.~Zhang, C.~Wu, Y.~Zhang, Y.~Wang, and W.~Xie, ``Pmc-clip:
Contrastive language-image pretraining using biomedical documents,'' in
\emph{International Conference on Medical Image Computing and
	Computer-Assisted Intervention}.\hskip 1em plus 0.5em minus 0.4em\relax
Springer, 2023, pp. 525--536.

\bibitem{chimedvl} J.~Liu, Z.~Wang, Q.~Ye, D.~Chong, P.~Zhou, and Y.~Hua, ``Qilin-med-vl: Towards
chinese large vision-language model for general healthcare,'' \emph{arXiv
	preprint arXiv:2310.17956}, 2023.

\bibitem{ffair} M.~Li, W.~Cai, R.~Liu, Y.~Weng, X.~Zhao, C.~Wang, X.~Chen, Z.~Liu, C.~Pan,
M.~Li \emph{et~al.}, ``Ffa-ir: Towards an explainable and reliable medical
report generation benchmark,'' in \emph{Thirty-fifth Conference on Neural Information Processing Systems Datasets and Benchmarks Track (Round 2)},
2021.

\bibitem{FNL_Bustos} A.~Bustos, A.~Pertusa, J.-M. Salinas, and M.~De~La Iglesia-Vaya, ``Padchest: A
large chest x-ray image dataset with multi-label annotated reports,''
\emph{Medical Image Analysis}, vol.~66, p. 101797, 2020.

\bibitem{MIMICCXR} A.~E. Johnson, T.~J. Pollard, S.~J. Berkowitz, N.~R. Greenbaum, M.~P. Lungren,
C.-y. Deng, R.~G. Mark, and S.~Horng, ``Mimic-cxr, a de-identified publicly
available database of chest radiographs with free-text reports,''
\emph{Scientific Data}, vol.~6, no.~1, p. 317, 2019.

\bibitem{CT-RATE} I.~E. Hamamci, S.~Er, F.~Almas, A.~G. Simsek, S.~N. Esirgun, I.~Dogan, M.~F.
Dasdelen, B.~Wittmann, E.~Simsar, M.~Simsar \emph{et~al.}, ``A foundation
model utilizing chest ct volumes and radiology reports for supervised-level
zero-shot detection of abnormalities,'' \emph{arXiv preprint
	arXiv:2403.17834}, 2024.

\bibitem{OpenPath} Z.~Huang, F.~Bianchi, M.~Yuksekgonul, T.~J. Montine, and J.~Zou, ``A
visual--language foundation model for pathology image analysis using medical
twitter,'' \emph{Nature Medicine}, vol.~29, no.~9, pp. 2307--2316, 2023.

\bibitem{Quilt1M} W.~Ikezogwo, S.~Seyfioglu, F.~Ghezloo, D.~Geva, F.~Sheikh~Mohammed, P.~K.
Anand, R.~Krishna, and L.~Shapiro, ``Quilt-1m: One million image-text pairs
for histopathology,'' \emph{Advances in Neural Information Processing
	Systems}, vol.~36, 2024.

\bibitem{pgiu} J.~Pavlopoulos, V.~Kougia, and I.~Androutsopoulos, ``A survey on biomedical
image captioning,'' in \emph{Proceedings of The Second Workshop on
	Shortcomings in Vision and Language}, 2019, pp. 26--36.

\bibitem{SLAKE} B.~Liu, L.-M. Zhan, L.~Xu, L.~Ma, Y.~Yang, and X.-M. Wu, ``Slake: A
semantically-labeled knowledge-enhanced dataset for medical visual question
answering,'' in \emph{2021 IEEE 18th International Symposium on Biomedical
	Imaging (ISBI)}.\hskip 1em plus 0.5em minus 0.4em\relax IEEE, 2021, pp.
1650--1654.

\bibitem{pathvqa} X.~He, Y.~Zhang, L.~Mou, E.~Xing, and P.~Xie, ``Pathvqa: 30000+ questions for
medical visual question answering,'' \emph{arXiv preprint arXiv:2003.10286},
2020.

\bibitem{VQARAD} J.~J. Lau, S.~Gayen, A.~Ben~Abacha, and D.~Demner-Fushman, ``A dataset of
clinically generated visual questions and answers about radiology images,''
\emph{Scientific Data}, vol.~5, no.~1, pp. 1--10, 2018.

\bibitem{retclip} J.~Du, J.~Guo, W.~Zhang, S.~Yang, H.~Liu, H.~Li, and N.~Wang, ``Ret-clip: A
retinal image foundation model pre-trained with clinical diagnostic
reports,'' \emph{arXiv preprint arXiv:2405.14137}, 2024.

\bibitem{kirillov2023segment} A.~Kirillov, E.~Mintun, N.~Ravi, H.~Mao, C.~Rolland, L.~Gustafson, T.~Xiao,
S.~Whitehead, A.~C. Berg, W.-Y. Lo \emph{et~al.}, ``Segment anything,'' in
\emph{Proceedings of the IEEE/CVF International Conference on Computer
	Vision}, 2023, pp. 4015--4026.

\bibitem{cheng2023sam} J.~Cheng, J.~Ye, Z.~Deng, J.~Chen, T.~Li, H.~Wang, Y.~Su, Z.~Huang, J.~Chen,
L.~Jiang \emph{et~al.}, ``Sam-med2d,'' \emph{arXiv preprint
	arXiv:2308.16184}, 2023.

\bibitem{dong2024segment} H.~Dong, H.~Gu, Y.~Chen, J.~Yang, and M.~A. Mazurowski, ``Segment anything
model 2: an application to 2d and 3d medical images,'' \emph{arXiv preprint
	arXiv:2408.00756}, 2024.

\bibitem{zhu2024medical} J.~Zhu, Y.~Qi, and J.~Wu, ``Medical sam 2: Segment medical images as video via
segment anything model 2,'' \emph{arXiv preprint arXiv:2408.00874}, 2024.

\bibitem{gong20243dsam} S.~Gong, Y.~Zhong, W.~Ma, J.~Li, Z.~Wang, J.~Zhang, P.-A. Heng, and Q.~Dou,
``3dsam-adapter: Holistic adaptation of sam from 2d to 3d for promptable
tumor segmentation,'' \emph{Medical Image Analysis}, vol.~98, p. 103324,
2024.

\bibitem{chen2024ma} C.~Chen, J.~Miao, D.~Wu, A.~Zhong, Z.~Yan, S.~Kim, J.~Hu, Z.~Liu, L.~Sun, X.~Li
\emph{et~al.}, ``Ma-sam: Modality-agnostic sam adaptation for 3d medical
image segmentation,'' \emph{Medical Image Analysis}, vol.~98, p. 103310,
2024.

\bibitem{wang2023sammed3d} H.~Wang, S.~Guo, J.~Ye, Z.~Deng, J.~Cheng, T.~Li, J.~Chen, Y.~Su, Z.~Huang,
Y.~Shen, B.~Fu, S.~Zhang, J.~He, and Y.~Qiao, ``Sam-med3d,'' \emph{arXiv preprint
	arXiv:2310.15161}, 2023.

\bibitem{du2023segvol} Y.~Du, F.~Bai, T.~Huang, and B.~Zhao, ``Segvol: Universal and interactive
volumetric medical image segmentation,'' \emph{arXiv preprint
	arXiv:2311.13385}, 2023.

\bibitem{wang2024sam} G.~Wang, J.~Ye, J.~Cheng, T.~Li, Z.~Chen, J.~Cai, J.~He, and B.~Zhuang,
``Sam-med3d-moe: Towards a non-forgetting segment anything model via mixture
of experts for 3d medical image segmentation,'' in \emph{International Conference on Medical Image Computing and
	Computer-Assisted Intervention}.\hskip 1em plus 0.5em minus 0.4em\relax Springer, 2024, pp.
552--561.

\bibitem{huang2023stu} Z.~Huang, H.~Wang, Z.~Deng, J.~Ye, Y.~Su, H.~Sun, J.~He, Y.~Gu, L.~Gu, S.~Zhang
\emph{et~al.}, ``Stu-net: Scalable and transferable medical image
segmentation models empowered by large-scale supervised pretraining,''
\emph{arXiv preprint arXiv:2304.06716}, 2023.

\bibitem{he2022masked} K.~He, X.~Chen, S.~Xie, Y.~Li, P.~Doll{\'a}r, and R.~Girshick, ``Masked
autoencoders are scalable vision learners,'' in \emph{Proceedings of the
	IEEE/CVF Conference on Computer Vision and Pattern Recognition}, 2022, pp.
16\,000--16\,009.

\bibitem{xie2022simmim} Z.~Xie, Z.~Zhang, Y.~Cao, Y.~Lin, J.~Bao, Z.~Yao, Q.~Dai, and H.~Hu, ``Simmim:
A simple framework for masked image modeling,'' in \emph{Proceedings of the
	IEEE/CVF Conference on Computer Vision and Pattern Recognition}, 2022, pp.
9653--9663.

\bibitem{wang2023autosmim} Z.~Wang, J.~Lyu, and X.~Tang, ``Autosmim: Automatic superpixel-based masked
image modeling for skin lesion segmentation,'' \emph{IEEE Transactions on
	Medical Imaging}, 2023.

\bibitem{li2024anatomask} Y.~Li, T.~Luan, Y.~Wu, S.~Pan, Y.~Chen, and X.~Yang, ``Anatomask: Enhancing
medical image segmentation with reconstruction-guided self-masking,''
\emph{arXiv preprint arXiv:2407.06468}, 2024.

\bibitem{wang2023masked} H.~Wang, Y.~Tang, Y.~Wang, J.~Guo, Z.-H. Deng, and K.~Han, ``Masked image
modeling with local multi-scale reconstruction,'' in \emph{Proceedings of the
	IEEE/CVF Conference on Computer Vision and Pattern Recognition}, 2023, pp.
2122--2131.

\bibitem{cai2024uni4eye++} Z.~Cai, L.~Lin, H.~He, P.~Cheng, and X.~Tang, ``Uni4eye++: A general masked
image modeling multi-modal pretraining framework for ophthalmic image
classification and segmentation,'' \emph{IEEE Transactions on Medical
	Imaging}, 2024.

\bibitem{xie2023medim} Y.~Xie, L.~Gu, T.~Harada, J.~Zhang, Y.~Xia, and Q.~Wu, ``Medim: Boost medical
image representation via radiology report-guided masking,'' in
\emph{International Conference on Medical Image Computing and
	Computer-Assisted Intervention}.\hskip 1em plus 0.5em minus 0.4em\relax
Springer, 2023, pp. 13--23.

\bibitem{yang2023mrm} Q.~Yang, W.~Li, B.~Li, and Y.~Yuan, ``Mrm: Masked relation modeling for medical
image pretraining with genetics,'' in \emph{Proceedings of the IEEE/CVF
	International Conference on Computer Vision}, 2023, pp. 21\,452--21\,462.

\bibitem{zhou2023advancing} H.-Y. Zhou, C.~Lian, L.~Wang, and Y.~Yu, ``Advancing radiograph representation
learning with masked record modeling,'' \emph{arXiv preprint
	arXiv:2301.13155}, 2023.

\bibitem{luo2022self} Y.~Luo, Z.~Chen, S.~Zhou, and X.~Gao, ``Self-distillation augmented masked
autoencoders for histopathological image classification,'' \emph{arXiv
	preprint arXiv:2203.16983}, 2022.

\bibitem{zhuang2023advancing} J.-X. Zhuang, L.~Luo, and H.~Chen, ``Advancing volumetric medical image
segmentation via global-local masked autoencoder,'' \emph{arXiv preprint
	arXiv:2306.08913}, 2023.

\bibitem{liu2023m3ae} H.~Liu, D.~Wei, D.~Lu, J.~Sun, L.~Wang, and Y.~Zheng, ``M3ae: multimodal
representation learning for brain tumor segmentation with missing
modalities,'' in \emph{Proceedings of the AAAI Conference on Artificial
	Intelligence}, vol.~37, no.~2, 2023, pp. 1657--1665.

\bibitem{wu2024voco} L.~Wu, J.~Zhuang, and H.~Chen, ``Voco: A simple-yet-effective volume
contrastive learning framework for 3d medical image analysis,'' in
\emph{Proceedings of the IEEE/CVF Conference on Computer Vision and Pattern
	Recognition}, 2024, pp. 22\,873--22\,882.

\bibitem{zeng2021positional} D.~Zeng, Y.~Wu, X.~Hu, X.~Xu, H.~Yuan, M.~Huang, J.~Zhuang, J.~Hu, and Y.~Shi,
``Positional contrastive learning for volumetric medical image
segmentation,'' in \emph{International Conference on Medical Image Computing and
	Computer-Assisted Intervention}.\hskip 1em plus 0.5em
minus 0.4em\relax Springer, 2021, pp. 221--230.

\bibitem{wang2024contrast} Y.~Wang, Y.~Han, H.~Wang, and X.~Zhang, ``Contrast everything: A hierarchical
contrastive framework for medical time-series,'' \emph{Advances in Neural
	Information Processing Systems}, vol.~36, 2024.

\bibitem{azizi2021big} S.~Azizi, B.~Mustafa, F.~Ryan, Z.~Beaver, J.~Freyberg, J.~Deaton, A.~Loh,
A.~Karthikesalingam, S.~Kornblith, T.~Chen \emph{et~al.}, ``Big
self-supervised models advance medical image classification,'' in
\emph{Proceedings of the IEEE/CVF International Conference on Computer
	Vision}, 2021, pp. 3478--3488.

\bibitem{haghighi2024self} F.~Haghighi, M.~R.~H. Taher, M.~B. Gotway, and J.~Liang, ``Self-supervised
learning for medical image analysis: Discriminative, restorative, or
adversarial?'' \emph{Medical Image Analysis}, vol.~94, p. 103086, 2024.

\bibitem{goodfellow2020generative} I. Goodfellow, J. Pouget-Abadie, M.Mirza, B. Xu, D. Warde-Farley, S. Ozair, A. Courville, and Y. Bengio, "Generative adversarial networks," Communications of the ACM,vol.63, no.11, pp.139--144, 2020.

\bibitem{zhou2021preservational} H.-Y. Zhou, C.~Lu, S.~Yang, X.~Han, and Y.~Yu, ``Preservational learning
improves self-supervised medical image models by reconstructing diverse
contexts,'' in \emph{Proceedings of the IEEE/CVF International Conference on
	Computer Vision}, 2021, pp. 3499--3509.

\bibitem{zhou2023unified} H.-Y. Zhou, C.~Lu, C.~Chen, S.~Yang, and Y.~Yu, ``A unified visual information
preservation framework for self-supervised pretraining in medical image
analysis,'' \emph{IEEE Transactions on Pattern Analysis and Machine
	Intelligence}, vol.~45, no.~7, pp. 8020--8035, 2023.

\bibitem{xie2022unimiss} Y.~Xie, J.~Zhang, Y.~Xia, and Q.~Wu, ``Unimiss: Universal medical
self-supervised learning via breaking dimensionality barrier,'' in
\emph{Proceedings of the European Conference on Computer Vision}, 2022, pp. 558--575.

\bibitem{wang2023swinmm} Y.~Wang, Z.~Li, J.~Mei, Z.~Wei, L.~Liu, C.~Wang, S.~Sang, A.~L. Yuille, C.~Xie,
and Y.~Zhou, ``Swinmm: masked multi-view with swin transformers for 3d
medical image segmentation,'' in \emph{International Conference on Medical
	Image Computing and Computer-Assisted Intervention}.\hskip 1em plus 0.5em
minus 0.4em\relax Springer, 2023, pp. 486--496.

\bibitem{clip_global_1} Y.~Zhang, H.~Jiang, Y.~Miura, C.~D. Manning, and C.~P. Langlotz, ``Contrastive
learning of medical visual representations from paired images and text,'' in
\emph{Machine Learning for Healthcare Conference}.\hskip 1em plus 0.5em minus
0.4em\relax PMLR, 2022, pp. 2--25.

\bibitem{clip_global_2} H.-Y. Zhou, X.~Chen, Y.~Zhang, R.~Luo, L.~Wang, and Y.~Yu, ``Generalized
radiograph representation learning via cross-supervision between images and
free-text radiology reports,'' \emph{Nature Machine Intelligence}, vol.~4,
no.~1, pp. 32--40, 2022.

\bibitem{gloria} S.-C. Huang, L.~Shen, M.~P. Lungren, and S.~Yeung, ``Gloria: A multimodal
global-local representation learning framework for label-efficient medical
image recognition,'' in \emph{Proceedings of the IEEE/CVF International
	Conference on Computer Vision}, 2021, pp. 3942--3951.

\bibitem{LoVT} P.~M{\"u}ller, G.~Kaissis, C.~Zou, and D.~Rueckert, ``Joint learning of
localized representations from medical images and reports,'' in
\emph{Proceedings of the European Conference on Computer Vision}, 2022, pp. 685--701.

\bibitem{local_pm} P.~M{\"u}ller, G.~Kaissis, and D.~Rueckert, ``The role of local alignment and
uniformity in image-text contrastive learning on medical images,''
\emph{arXiv preprint arXiv:2211.07254}, 2022.

\bibitem{localmi} R.~Liao, D.~Moyer, M.~Cha, K.~Quigley, S.~Berkowitz, S.~Horng, P.~Golland, and
W.~M. Wells, ``Multimodal representation learning via maximization of local
mutual information,'' in \emph{International Conference on Medical Image Computing and
	Computer-Assisted Intervention}.\hskip 1em plus 0.5em
minus 0.4em\relax Springer, 2021, pp. 273--283.

\bibitem{local_seibold} C.~Seibold, S.~Rei{\ss}, M.~S. Sarfraz, R.~Stiefelhagen, and J.~Kleesiek,
``Breaking with fixed set pathology recognition through report-guided
contrastive training,'' in \emph{International Conference on Medical Image
	Computing and Computer-Assisted Intervention}.\hskip 1em plus 0.5em minus
0.4em\relax Springer, 2022, pp. 690--700.

\bibitem{local_tier} A.~Palepu and A.~Beam, ``Tier: Text-image entropy regularization for medical
clip-style models,'' in \emph{Machine Learning for Healthcare
	Conference}.\hskip 1em plus 0.5em minus 0.4em\relax PMLR, 2023, pp. 548--564.

\bibitem{PRIOR} P.~Cheng, L.~Lin, J.~Lyu, Y.~Huang, W.~Luo, and X.~Tang, ``Prior: Prototype
representation joint learning from medical images and reports,'' in
\emph{Proceedings of the IEEE/CVF International Conference on Computer
	Vision}, 2023, pp. 21\,361--21\,371.

\bibitem{medclip} Z.~Wang, Z.~Wu, D.~Agarwal, and J.~Sun, ``Medclip: Contrastive learning from
unpaired medical images and text,'' \emph{arXiv preprint arXiv:2210.10163},
2022.

\bibitem{FNL_LiuBo} B.~Liu, D.~Lu, D.~Wei, X.~Wu, Y.~Wang, Y.~Zhang, and Y.~Zheng, ``Improving
medical vision-language contrastive pretraining with semantics-aware
triage,'' \emph{IEEE Transactions on Medical Imaging}, 2023.

\bibitem{umcl} Y.~Wang and G.~Wang, ``Umcl: Unified medical image-text-label contrastive
learning with continuous prompt,'' in \emph{2023 IEEE International
	Conference on Bioinformatics and Biomedicine (BIBM)}.\hskip 1em plus 0.5em
minus 0.4em\relax IEEE, 2023, pp. 2285--2289.

\bibitem{biovil} S.~Bannur, S.~Hyland, Q.~Liu, F.~Perez-Garcia, M.~Ilse, D.~C. Castro,
B.~Boecking, H.~Sharma, K.~Bouzid, A.~Thieme \emph{et~al.}, ``Learning to
exploit temporal structure for biomedical vision-language processing,'' in
\emph{Proceedings of the IEEE/CVF Conference on Computer Vision and Pattern
	Recognition}, 2023, pp. 15\,016--15\,027.

\bibitem{ka_1} J.~Lei, L.~Dai, H.~Jiang, C.~Wu, X.~Zhang, Y.~Zhang, J.~Yao, W.~Xie, Y.~Zhang,
Y.~Li \emph{et~al.}, ``Unibrain: Universal brain mri diagnosis with
hierarchical knowledge-enhanced pretraining,'' \emph{arXiv preprint
	arXiv:2309.06828}, 2023.

\bibitem{medkebert} X.~Zhang, C.~Wu, Y.~Zhang, W.~Xie, and Y.~Wang, ``Knowledge-enhanced
visual-language pretraining on chest radiology images,'' \emph{Nature
	Communications}, vol.~14, no.~1, p. 4542, 2023.

\bibitem{medklip} C.~Wu, X.~Zhang, Y.~Zhang, Y.~Wang, and W.~Xie, ``Medklip: Medical knowledge
enhanced language-image pretraining for x-ray diagnosis,'' in
\emph{Proceedings of the IEEE/CVF International Conference on Computer
	Vision}, 2023, pp. 21\,372--21\,383.

\bibitem{med_flamingo} M.~Moor, Q.~Huang, S.~Wu, M.~Yasunaga, C.~Zakka, Y.~Dalmia, E.~P. Reis,
P.~Rajpurkar, and J.~Leskovec, ``Med-flamingo: a multimodal medical few-shot
learner,'' 2023. [Online]. Available: \url{https://arxiv.org/abs/2307.15189}

\bibitem{pathchat} M.~Y. Lu, B.~Chen, D.~F. Williamson, R.~J. Chen, M.~Zhao, A.~K. Chow,
K.~Ikemura, A.~Kim, D.~Pouli, A.~Patel \emph{et~al.}, ``A multimodal
generative ai copilot for human pathology,'' \emph{Nature}, pp. 1--3, 2024.

\bibitem{Vinyals}
O.~Vinyals, A.~Toshev, S.~Bengio, and D.~Erhan, ``Show and tell: A neural image
  caption generator,'' in \emph{Proceedings of the IEEE conference on computer
  vision and pattern recognition}, 2015, pp. 3156--3164.

\bibitem{tienet}
X.~Wang, Y.~Peng, L.~Lu, Z.~Lu, and R.~M. Summers, ``Tienet: Text-image
  embedding network for common thorax disease classification and reporting in
  chest x-rays,'' in \emph{Proceedings of the IEEE conference on computer
  vision and pattern recognition}, 2018, pp. 9049--9058.

\bibitem{Wehbe}
R.~M. Wehbe, A.~K. Katsaggelos, K.~J. Hammond, H.~Hong, F.~S. Ahmad, D.~Ouyang,
  S.~J. Shah, P.~M. McCarthy, and J.~D. Thomas, ``Deep learning for
  cardiovascular imaging: A review,'' \emph{JAMA cardiology}, vol.~8, no.~11,

\bibitem{liknowledge}
C.~Y. Li, X.~Liang, Z.~Hu, and E.~P. Xing, ``Knowledge-driven encode, retrieve,
  paraphrase for medical image report generation,'' in \emph{Proceedings of the
  AAAI conference on artificial intelligence}, vol.~33, no.~01, 2019, pp.
  6666--6673.

\bibitem{chen2020generating}
Z.~Chen, Y.~Song, T.-H. Chang, and X.~Wan, ``Generating radiology reports via
  memory-driven transformer,'' \emph{arXiv preprint arXiv:2010.16056}, 2020.

\bibitem{liu2021exploring}
F.~Liu, X.~Wu, S.~Ge, W.~Fan, and Y.~Zou, ``Exploring and distilling posterior
  and prior knowledge for radiology report generation,'' in \emph{Proceedings
  of the IEEE/CVF conference on computer vision and pattern recognition}, 2021,
  pp. 13\,753--13\,762.

\bibitem{qin2022medical}
Z.~Qin, H.~Yi, Q.~Lao, and K.~Li, ``Medical image understanding with pretrained
  vision language models: A comprehensive study,'' \emph{arXiv preprint
  arXiv:2209.15517}, 2022.

\bibitem{chen2023medblip}
Q.~Chen, X.~Hu, Z.~Wang, and Y.~Hong, ``Medblip: Bootstrapping language-image
  pretraining from 3d medical images and texts,'' \emph{arXiv preprint
  arXiv:2305.10799}, 2023.

\bibitem{xu2024foundation}
L.~Xu, Z.~Ni, H.~Sun, H.~Li, and S.~Zhang, ``A foundation model for
  generalizable disease diagnosis in chest x-ray images,'' \emph{arXiv preprint
  arXiv:2410.08861}, 2024.

\bibitem{miller2024ai}
R.~J. Miller, A.~Shanbhag, A.~Killekar, M.~Lemley, B.~Bednarski, S.~D.
  Van~Kriekinge, P.~B. Kavanagh, A.~Feher, E.~J. Miller, A.~J. Einstein
  \emph{et~al.}, ``Ai-derived epicardial fat measurements improve
  cardiovascular risk prediction from myocardial perfusion imaging,'' \emph{NPJ
  Digital Medicine}, vol.~7, no.~1, p.~24, 2024.

\bibitem{vaid2023foundational}
A.~Vaid, J.~Jiang, A.~Sawant, S.~Lerakis, E.~Argulian, Y.~Ahuja, J.~Lampert,
  A.~Charney, H.~Greenspan, J.~Narula \emph{et~al.}, ``A foundational vision
  transformer improves diagnostic performance for electrocardiograms,''
  \emph{NPJ Digital Medicine}, vol.~6, no.~1, p. 108, 2023.

\bibitem{quer2024potential}
G.~Quer and E.~J. Topol, ``The potential for large language models to transform
  cardiovascular medicine,'' \emph{The Lancet Digital Health}, vol.~6

\bibitem{silva2025foundation}
J.~Silva-Rodriguez, H.~Chakor, R.~Kobbi, J.~Dolz, and I.~B. Ayed, ``A
  foundation language-image model of the retina (flair): Encoding expert
  knowledge in text supervision,'' \emph{Medical Image Analysis}, vol.~99, p.
  103357, 2025.

\bibitem{vorontsov2024foundation}
E.~Vorontsov, A.~Bozkurt, A.~Casson, G.~Shaikovski, M.~Zelechowski,
  K.~Severson, E.~Zimmermann, J.~Hall, N.~Tenenholtz, N.~Fusi \emph{et~al.},
  ``A foundation model for clinical-grade computational pathology and rare
  cancers detection,'' \emph{Nature medicine}, pp. 1--12, 2024.

\bibitem{wang2024pathology}
X.~Wang, J.~Zhao, E.~Marostica, W.~Yuan, J.~Jin, J.~Zhang, R.~Li, H.~Tang,
  K.~Wang, Y.~Li \emph{et~al.}, ``A pathology foundation model for cancer
  diagnosis and prognosis prediction,'' \emph{Nature}, pp. 1--9, 2024.

\bibitem{chen2024mammo}
X.~Chen, Y.~Li, M.~Hu, E.~Salari, X.~Chen, R.~L. Qiu, B.~Zheng, and X.~Yang,
  ``Mammo-clip: Leveraging contrastive language-image pretraining (clip) for
  enhanced breast cancer diagnosis with multi-view mammography,'' \emph{arXiv
  preprint arXiv:2404.15946}, 2024.

\bibitem{wu2023towards}
C.~Wu, X.~Zhang, Y.~Zhang, Y.~Wang, and W.~Xie, ``Towards generalist foundation
  model for radiology,'' \emph{arXiv preprint arXiv:2308.02463}, 2023.
  

\bibitem{deepbid} M.~Yasunaga, A.~Bosselut, H.~Ren, X.~Zhang, C.~D. Manning, P.~S. Liang, and
J.~Leskovec, ``Deep bidirectional language-knowledge graph pretraining,''
\emph{Advances in Neural Information Processing Systems}, vol.~35, pp.
37\,309--37\,323, 2022.

\bibitem{qagnn} M.~Yasunaga, H.~Ren, A.~Bosselut, P.~Liang, and J.~Leskovec, ``Qa-gnn:
Reasoning with language models and knowledge graphs for question answering,''
\emph{arXiv preprint arXiv:2104.06378}, 2021.

\bibitem{surgicalgpt} L.~Seenivasan, M.~Islam, G.~Kannan, and H.~Ren, ``Surgicalgpt: end-to-end
language-vision gpt for visual question answering in surgery,'' in
\emph{International Conference on Medical Image Computing and
	Computer-Assisted Intervention}.\hskip 1em plus 0.5em minus 0.4em\relax
Springer, 2023, pp. 281--290.

\bibitem{longtail} A.~G. Roy, J.~Ren, S.~Azizi, A.~Loh, V.~Natarajan, B.~Mustafa, N.~Pawlowski,
J.~Freyberg, Y.~Liu, Z.~Beaver \emph{et~al.}, ``Does your dermatology
classifier know what it doesn't know? detecting the long-tail of unseen
conditions,'' \emph{Medical Image Analysis}, vol.~75, p. 102274, 2022.

\bibitem{Endo_FM} Z.~Wang, C.~Liu, S.~Zhang, and Q.~Dou, ``Foundation model for endoscopy video
analysis via large-scale self-supervised pre-train,'' in \emph{International Conference on Medical Image Computing and
	Computer-Assisted Intervention}.\hskip 1em plus 0.5em minus 0.4em\relax Springer, 2023, pp.
101--111.

\bibitem{palme} D.~Driess, F.~Xia, M.~S. Sajjadi, C.~Lynch, A.~Chowdhery, B.~Ichter, A.~Wahid,
J.~Tompson, Q.~Vuong, T.~Yu \emph{et~al.}, ``Palm-e: An embodied multimodal
language model,'' \emph{arXiv preprint arXiv:2303.03378}, 2023.

\bibitem{vima} Y.~Jiang, A.~Gupta, Z.~Zhang, G.~Wang, Y.~Dou, Y.~Chen, L.~Fei-Fei,
A.~Anandkumar, Y.~Zhu, and L.~Fan, ``Vima: General robot manipulation with
multimodal prompts,'' \emph{arXiv preprint arXiv:2210.03094}, vol.~2, no.~3,
p.~6, 2022.



\bibitem{schlesinger2020deep}
D.~E. Schlesinger and C.~M. Stultz, ``Deep learning for cardiovascular risk
  stratification,'' \emph{Current Treatment Options in Cardiovascular
  Medicine}, vol.~22, pp. 1--14, 2020.

\bibitem{zhou2024comprehensive}
C.~Zhou, P.~Dai, A.~Hou, Z.~Zhang, L.~Liu, A.~Li, and F.~Wang, ``A
  comprehensive review of deep learning-based models for heart disease
  prediction,'' \emph{Artificial Intelligence Review}, vol.~57, no.~10, p. 263,
  2024.

\bibitem{williams2024artificial}
M.~C. Williams, J.~R. Weir-McCall, L.~A. Baldassarre, C.~N. De~Cecco, A.~D.
  Choi, D.~Dey, M.~R. Dweck, I.~Isgum, M.~Kolossvary, J.~Leipsic \emph{et~al.},
  ``Artificial intelligence and machine learning for cardiovascular computed
  tomography (cct): a white paper of the society of cardiovascular computed
  tomography (scct),'' \emph{Journal of Cardiovascular Computed Tomography},
  2024.

\bibitem{moshawrab2023reviewing}
M.~Moshawrab, M.~Adda, A.~Bouzouane, H.~Ibrahim, and A.~Raad, ``Reviewing
  multimodal machine learning and its use in cardiovascular diseases
  detection,'' \emph{Electronics}, vol.~12, no.~7, p. 1558, 2023.

\bibitem{alaa2019demystifying}
A.~M. Alaa and M.~van~der Schaar, ``Demystifying black-box models with symbolic
  metamodels,'' \emph{Advances in neural information processing systems},
  vol.~32, 2019.

\bibitem{esteva2017dermatologist}
A.~Esteva, B.~Kuprel, R.~A. Novoa, J.~Ko, S.~M. Swetter, H.~M. Blau, and
  S.~Thrun, ``Dermatologist-level classification of skin cancer with deep
  neural networks,'' \emph{nature}, vol. 542, no. 7639, pp. 115--118, 2017.

\bibitem{zeng2024realistic}
T.~Zeng, G.~Loza~Galindo, J.~Hu, P.~Valdastri, and D.~Jones, ``Realistic
  surgical image dataset generation based on 3d gaussian splatting,'' in
  \emph{International Conference on Medical Image Computing and
  Computer-Assisted Intervention}.\hskip 1em plus 0.5em minus 0.4em\relax
  Springer, 2024, pp. 510--519.

\bibitem{ahmed2024deep}
F.~A. Ahmed, M.~Yousef, M.~A. Ahmed, H.~O. Ali, A.~Mahboob, H.~Ali, Z.~Shah,
  O.~Aboumarzouk, A.~Al~Ansari, and S.~Balakrishnan, ``Deep learning for
  surgical instrument recognition and segmentation in robotic-assisted
  surgeries: a systematic review,'' \emph{Artificial Intelligence Review},
  vol.~58, no.~1, p.~1, 2024.

\bibitem{ronneberger2015u}
O.~Ronneberger, P.~Fischer, and T.~Brox, ``U-net: Convolutional networks for
  biomedical image segmentation,'' in \emph{Medical image computing and
  computer-assisted intervention--MICCAI 2015: 18th international conference,
  Munich, Germany, October 5-9, 2015, proceedings, part III 18}.\hskip 1em plus
  0.5em minus 0.4em\relax Springer, 2015, pp. 234--241.

\bibitem{he2017mask}
K.~He, G.~Gkioxari, P.~Doll{\'a}r, and R.~Girshick, ``Mask r-cnn,'' in
  \emph{Proceedings of the IEEE international conference on computer vision},
  2017, pp. 2961--2969.

\bibitem{hussain2022deep}
S.~M. Hussain, A.~Brunetti, G.~Lucarelli, R.~Memeo, V.~Bevilacqua, and
  D.~Buongiorno, ``Deep learning based image processing for robot assisted
  surgery: a systematic literature survey,'' \emph{IEEE Access}, vol.~10, pp.
  122\,627--122\,657, 2022.

\bibitem{dipietro2016recognizing}
R.~DiPietro, C.~Lea, A.~Malpani, N.~Ahmidi, S.~S. Vedula, G.~I. Lee, M.~R. Lee,
  and G.~D. Hager, ``Recognizing surgical activities with recurrent neural
  networks,'' in \emph{Medical Image Computing and Computer-Assisted
  Intervention--MICCAI 2016: 19th International Conference, Athens, Greece,
  October 17-21, 2016, Proceedings, Part I 19}.\hskip 1em plus 0.5em minus
  0.4em\relax Springer, 2016, pp. 551--558.

\bibitem{ahmidi2017dataset}
N.~Ahmidi, L.~Tao, S.~Sefati, Y.~Gao, C.~Lea, B.~B. Haro, L.~Zappella,
  S.~Khudanpur, R.~Vidal, and G.~D. Hager, ``A dataset and benchmarks for
  segmentation and recognition of gestures in robotic surgery,'' \emph{IEEE
  Transactions on Biomedical Engineering}, vol.~64, no.~9, pp. 2025--2041,
  2017.




\end{thebibliography}
\end{document}